\documentclass[journal,twoside,web]{IEEEtran}

\usepackage{amsthm}
\usepackage{amsmath}
\usepackage{amsfonts}
\usepackage{amssymb}
\usepackage[export]{adjustbox}
\usepackage{bm}
\usepackage{subcaption}
\usepackage{accents}
\newcommand{\ubar}[1]{\underaccent{\bar}{#1}}
\makeatletter
\newcommand{\pushright}[1]{\ifmeasuring@#1\else\omit\hfill$\displaystyle#1$\fi\ignorespaces}
\newcommand{\pushleft}[1]{\ifmeasuring@#1\else\omit$\displaystyle#1$\hfill\fi\ignorespaces}
\makeatother

\theoremstyle{definition}
\newtheorem{theorem}{Theorem}[section]
\newtheorem{corollary}[theorem]{Corollary}
\newtheorem{assumption}[theorem]{\textbf{Assumption}}
\newtheorem{lemma}[theorem]{Lemma}
\newtheorem{proposition}[theorem]{Proposition}
\newtheorem{definition}[theorem]{Definition}
\newtheorem{remark}[theorem]{Remark}
\theoremstyle{remark}
\newtheorem{claim}{Claim}[theorem]

\newcommand{\proj}{\textrm{proj}}
\newcommand{\cov}{\textrm{cov}}
\newcommand{\var}{\textrm{var}}

\usepackage{algorithm}

\usepackage{algpseudocode}
\algrenewcommand\textproc{}
\usepackage[english]{babel}
\usepackage[utf8]{inputenc}
\usepackage[dvipsnames]{xcolor}
 
\algnewcommand{\parState}[1]{\State%
	\parbox[t]{\dimexpr\linewidth-\algmargin}{\strut\hangindent= \algorithmicindent \hangafter=1 #1\strut}}
\usepackage{array}
\allowdisplaybreaks

\title{
Lightweight Distributed Gaussian Process Regression for Online Machine Learning
}

\author{Zhenyuan Yuan \qquad Minghui Zhu
\thanks{Zhenyuan Yuan and Minghui Zhu are with School of Electrical Engineering and Computer Science, Pennsylvania State University, University Park, PA 16802,USA (email:
        {\tt\small \{zqy5086,muz16\}@psu.edu}). This work was partially supported by NSF grants ECCS-1710859 and CNS-1830390.}%
}

\begin{document}

\maketitle

\IEEEpeerreviewmaketitle

\begin{abstract}
In this paper, we study the problem where a group of agents aim to collaboratively learn a common  static latent function through streaming data. We propose a lightweight distributed Gaussian process regression (GPR) algorithm that is cognizant of agents' limited capabilities in communication, computation and memory. Each agent independently runs agent-based GPR using local streaming data to predict test points of interest; then the agents collaboratively execute distributed GPR to obtain global predictions over a common sparse set of test points; finally, each agent fuses results from distributed GPR with  agent-based GPR to refine its predictions.  By quantifying the transient and steady-state performances in predictive variance and error, we show that limited inter-agent communication improves learning performances in the sense of Pareto.  Monte Carlo simulation is conducted to evaluate the developed algorithm. 
\end{abstract}

\section{INTRODUCTION}
Networks of agents can access large amount of streaming data online in many applications, e.g.,  autonomous driving \cite{levinson2011towards}, and precision agriculture \cite{milioto2018real}. Machine learning has been increasingly adopted to extract reliable and actionable information from big data and enable agents to adapt and react in uncertain and dynamically changing environments. Nowadays machine learning algorithms have achieved remarkable performances in terms of consistency \cite{vapnik2013nature}, generalization \cite{valiant1984theory} and robustness \cite{biggio2018wild}. However, limited resources challenge implementation of the algorithms on physical agents, e.g., mobile robots.

Gaussian process regression (GPR) \cite{williams2006gaussian} is an efficient nonparametric statistical learning model.  GPR models a target function as a sample from a Gaussian process prior specified by a pair of mean and covariance functions only depending on inputs. With proper choice of prior covariance function, also referred as kernel, and mild assumptions on the target function, GPR is able to consistently approximate any continuous function \cite{choi2007posterior}. With optimal sampling in the input space and covariance functions obeying Sacks-Ylvisaker conditions of order $r$, the generalization error of GPR diminishes at the rate of $\mathcal{O}(n_s^{-(2r+1)/(2r+2)})$ (Section V.2, \cite{ritter2007average}), where $n_s$ is the number of samples. 
GPR is able to quantify learning uncertainties since it predicts the target function in the form of a posterior probability distribution. GPR has demonstrated powerful capabilities in various applications, e.g., optimization \cite{deisenroth2009gaussian}, motion planning \cite{mukadam2016gaussian}, and trajectory estimation \cite{barfoot2014batch}. 

GPR scales as $\mathcal{O}(n_s^3)$ in computational complexity and $\mathcal{O}(n_s^2)$ in memory (page 171, \cite{williams2006gaussian}), which prohibits applications with large datasets. There are multiple sparse approximation methods for large datasets.  One major class of approximation methods, which is also referred as global approximation, tackles the computational complexity by achieving the sparsity of the Gram matrix. Methods include using a subset of data to approximate the whole training dataset, designing a sparse kernel, and sparsifying the Gram matrix. The best possible result can be achieved by global approximation algorithms is $\mathcal{O}(m_s^3)$ in computational complexity and $\mathcal{O}(m_s^2)$ in memory, where 
  $m_s\ll n_s$ is the number of inducing points or the size of a subset of training data. 
More details about global approximation can be found in the recent survey paper \cite{liu2020gaussian}.
In the community of Geostatistics, Nearest-neighbor GPR \cite{vecchia1988estimation} is applied \cite{datta2016hierarchical}\cite{finley2019efficient}, where the predictions are made only using the training data of the nearest input. It requires only $\mathcal{O}(n_s)$ in both memory and (worst-case) computation. 

Centralized implementation of GPR is not suitable for networks of agents due to poor scalability in data size, high cost in communication and memory, and fragility to single-point failures. 
There have been studies on distributed GPR over server-client architecture, which is also referred to as divide-and-conquer approach or local approximations \cite{liu2020gaussian}. In the server-client architecture, a server acts as the centralized entity that partitions a dataset and assigns each subset of the data to computing units (clients). The clients perform training independently and send their learning results to the server for post-processing. 
These methods speed up the training process and are able to scale to arbitrarily large datasets. Communication budget constraint is considered in \cite{tavassolipour2019learning} by reducing the dimensionality of transmitted data to approximate the whole dataset. Sparse approximation of full GPR is used in \cite{tavassolipour2019learning} to further relieve the communication overhead. Notice that the server-client architecture requires each client being well-connected with the server, and is not robust to the failure of the server.
Paper \cite{chen2015gaussian} decentralizes sparse approximations of full GPR for fixed datasets over complete communication graphs. A distributed algorithm is also proposed to deal with fixed and sparse graphs. However, this paper considers offline learning with static datasets on the agents and does not provide theoretic guarantee on the distributed algorithm. 

Our work is related to multi-agent regression using kernel methods and basis functions. 
Papers \cite{predd2009collaborative}\cite{pillonetto2018distributed}\cite{varagnolo2012distributed} study offline learning, where all training data is provided before learning,  using kernel methods. Papers \cite{martinez2010distributed}\cite{xu2013efficient}\cite{choi2009distributed} study online learning, where training data is collected successively by mobile robots, using basis functions. In particular, they approximate the unknown functions with a linear combination of a finite number of known basis functions. This reduces the problem into a parameter estimation problem. 
From the perspective of regression, the problem investigated in \cite{martinez2010distributed} is equivalent to selecting the centers of a finite number of basis functions defined by Voronoi partition. In contrast, this paper considers online learning of abstract agents using Gaussian processes, where the unknown function is modelled as a sample from a distribution of functions.

\textbf{Contribution statement.} We consider the problem where a group of agents aim to collaboratively learn a common static latent   function  through streaming data. We propose Lightweight Distributed Gaussian Process Regression (LiDGPR) algorithm for the agents to solve the problem. More specifically, each agent independently runs agent-based GPR using local streaming data to predict the test points of interest; then the agents collaboratively execute distributed GPR to obtain global predictions over a common sparse set of test points; finally, each agent fuses the results from distributed GPR with those from  agent-based GPR to refine its predictions.   Our analysis of the transient and steady-state performances in predictive variance and error reveals that through communication agents whose data samples have lower dispersion  (or observation noise has lower variance) help improve the performance of the agents whose data samples have higher dispersion (or observation noise has higher variance). The improvements  in learning performances are in the sense of Pareto, i.e., some agents’ performances improve without
sacrificing other agents’ performances. In summary, our major contributions are two-fold:
\begin{itemize}
    \item 
    We develop LiDGPR that is cognizant of agents' limited capabilities in communication, computation and memory. 
    \item 
    We analyze the predictive mean and variance of LiDGPR and quantify the  improvements of the agents' learning performances resulted from inter-agent communication. 

\end{itemize}
Monte Carlo simulation is conducted to evaluate the developed algorithm. 
In addition to the preliminary version of this paper \cite{YUAN2020Comm}, this version includes a new set of theoretical results on the steady-state performance of predictive errors, identifies the factors that affect the improvement of learning performances, provides the proofs of the theoretical results, and discusses how the algorithm is cognizant of limited resources.


{\em Notations: }\label{sec:notation}
We use lower-case letters, e.g., $a$, to denote scalars, bold letters, e.g., $\bm{a}$, to denote vectors; we use upper-case letters, e.g., $A$, to denote matrices, calligraphic letters, e.g., $\mathcal{A}$, to denote sets, and bold calligraphic letters, e.g., $\bm{\mathcal{A}}$, to denote spaces. For any vector $\bm{a}$, we use $a_i$ to denote the $i$-th entry of $\bm{a}$. For any matrix $A$, we denote $a_{ij}$ as the entry at $i$-th row $j$-th column. Denote $I_n\in\mathbb{R}^{n\times n}$ the $n$-by-$n$-dimensional identity matrix, $\bm{1}_n\in\mathbb{R}^n$ the $n$-dimensional column vector with all $1$'s, i.e., $[1,\cdots,1]^T$, and $\bm{0}_n$ analogously.

We use superscript $(\cdot)^{[i]}$ to distinguish the local values of agent $i$, and $(\cdot)^{\max}$ ($(\cdot)^{min}$) denote the maximum (minimum) of the local values, e.g., $a^{\max}\triangleq\max_{i\in\mathcal{V}}a^{[i]}$. We denote superscript $(\cdot)^T$ the transpose of a vector or matrix, bracket $[\cdot]_\mathcal{E}$ the column vector with elements satisfying event $\mathcal{E}$.  Denote $\mathbb{E}_{a}[\cdot]$ the expectation taken over the distribution of random variable $a$, and $P\{\cdot\}$ a distribution.
We use $\mathcal{O}(\cdot)$ to denote the conventional Big O notation, i.e.,  $\mathcal{O}(g(t))$ represents the limiting behavior  of some function $f(t)$ if $\underset{t\to\infty}{\lim}\frac{f(t)}{g(t)}=a$ for some constant $a>0$.

We use $\succeq(\preceq)$ to denote element-wise comparison between two vectors, i.e., for any $\bm{a},\bm{b}\in\mathbb{R}^n$, $\bm{a}\succeq(\preceq)\bm{b}$ if and only if $a_i\geqslant(\leqslant) b_i$ for all $i=1,\cdots, n$. 
Operation $|\bm{a}|$ takes the absolute values element-wise on vector $\bm{a}$, $|\mathcal{A}|$ returns the cardinality of set $\mathcal{A}$, 
$\|\bm{a}\|_\infty\triangleq \max_j |a_j|$ for any vector $\bm{a}$.
Define the distance metric $\rho(\bm{z},\bm{z}')\triangleq\|\bm{z}-\bm{z}'\|$, the point to set distance as $\rho(\bm{z},\mathcal{Z})\triangleq\underset{\bm{z}'\in\mathcal{Z}}{\inf}\rho(\bm{z},\bm{z}')$.  Define $\proj(\bm{z},\mathcal{Z})\triangleq\{\bm{z}'\in\bm{\mathcal{Z}}|\rho(\bm{z},\bm{z}')=\rho(\bm{z},\mathcal{Z})\}$ the projection set of point $\bm{z}$ onto set $\mathcal{Z}$.   Denote the supremum of a function $\eta$ as $\|\eta\|_{\bm{\mathcal{Z}}}\triangleq\sup_{\bm{z}\in\bm{\mathcal{Z}}}|\eta(\bm{z})|$.

\section{Problem Statement}\label{sec:problem statement}
\emph{Network model.} Consider a network of agents represented by a directed time-varying communication graph $\mathcal{G}(t)\triangleq(\mathcal{V},\mathcal{E}(t))$, where $\mathcal{V}\triangleq\{1,\cdots,n\}$ represents the agent set, and $\mathcal{E}(t)\subseteq \mathcal{V}\times \mathcal{V}$ denotes the edge set at time $t$. Notice that $(i,j)\in \mathcal{E}(t)$ if and only if agent $i$ can receive messages from agent $j$ at time $t$. 
Define the set of the neighbors of agent $i$ at time $t$ as $\mathcal{N}^{[i]}(t)\triangleq\{j\in \mathcal{V}:(i,j)\in \mathcal{E}(t) \textrm{, and } j\neq i\}$.  The matrix $A(t)\in\mathbb{R}^{n\times n}$ represents the adjacency matrix of $\mathcal{G}(t)$ where $a_{ij}(t)\neq 0$ if $(i,j)\in \mathcal{E}(t)$. We make the following standard assumptions \cite{zhu2015distributed} about the network topology:
\begin{assumption}\label{assmp:Periodical Strong Connectivity}
(Periodical Strong Connectivity). There exists positive integer $b\geqslant 1$ such that, for all time instant $t\geqslant 0$, the directed graph $(\mathcal{V},\mathcal{E}(t)\cup\mathcal{E}(t+1)\cup\cdots\cup\mathcal{E}(t+(b-1)))$ is strongly connected. $\hfill\blacksquare$
\end{assumption}
This guarantees the information of each agent can reach any other agents in the network within finite time.

\begin{assumption}\label{assmp:Balanced Communication}
(Balanced Communication). It holds that $\bm{1}_{n}^TA(t)=\bm{1}_{n}^T$ and $A(t)\bm{1}_{n}=\bm{1}_{n}$, for all $t\geqslant 0$.$\hfill\blacksquare$
\end{assumption}
In the consensus literature, the first part of Assumption \ref{assmp:Balanced Communication} is called column stochasticity and is a standard sufficient condition to reach consensus. The second part is called row stochasticity and is needed to guarantee average consensus.

\begin{assumption}\label{assmp:Non-degeneracy} (Non-degeneracy).
There exists a constant $\alpha>0$ such that $a_{ii}(t)\geqslant\alpha$ and  $a_{ij}(t)\in\{0\}\cup[\alpha,1]$, $i\neq j$,  for all $t\geqslant0$.$\hfill\blacksquare$
\end{assumption}
That is, each agent assigns nontrivial weights on information from itself and its neighbors.

\emph{Observation model.} At each time instant $t$, each agent independently observes the outputs of a continuous common  static latent  function  $\eta:\bm{\mathcal{Z}}\rightarrow \bm{\mathcal{Y}}$ with zero-mean Gaussian noise, where $\bm{\mathcal{Z}}\subseteq\mathbb{R}^{n_z}$ is the compact input space for $\eta$. The observation model is given by 
\begin{equation}\label{eq: observation model}
    y^{[i]}(t)=\eta(\bm{z}^{[i]}(t))+e^{[i]}(t),\quad e^{[i]}(t)\sim\mathcal{N}(0,(\sigma_e^{[i]})^2),
\end{equation}
where $\bm{z}^{[i]}(t)\in\bm{\mathcal{Z}}$ is the input of $\eta$ from agent $i$ at time $t$, $y^{[i]}(t)\in\bm{\mathcal{Y}}$ is the observation of agent $i$, and $e^{[i]}(t)$ is independent Gaussian  noise.   Note that we do not assume that input $\bm{z}^{[i]}(t)$ follows any distribution, which is a standard assumption in statistical learning \cite{vapnik2013nature}. We let $\bm{\eta}(\mathcal{Z})$ return a column vector $[\eta(\bm{z})]_{\bm{z}\in\mathcal{Z}}$, and similarly for other functions. For notational simplicity, it is assumed that the output space $\bm{\mathcal{Y}}\subseteq\mathbb{R}$ because multi-dimensional observations can always be decomposed as aggregation of one-dimensional observations.



\emph{Problem Statement.} The objective of this paper is to design a distributed algorithm for the agents to learn the common static latent  function $\eta$ via streaming data $\{y^{[i]}(t), \bm{z}^{[i]}(t)\}_{t\geqslant 1}$. The challenges of the problem stem from the fact that the training dataset is monotonically expanding due to incremental sampling while the agents have limited resources in communication, computation and memory.

The followings are examples of potential applications of this formulation. One example can be a group of mobile robots deployed in a vast open area to collaboratively monitor a  static signal, such as temperature or wind field (see the case study in Section \ref{sec:simulation}). Other examples includes the learning of the dynamics of a moving target using a network of static sensors \cite{XZ-MZ:IROS20}. In addition to robotic applications, this formulation also applies to profit predictions in marketing and wheat crop prediction \cite{gyorfi2002distribution}.

\section{Preliminaries}\label{sec:preliminaries}
 In this section, we provide necessary background on GPR. 
Let $\eta:\bm{\mathcal{Z}}\to\mathbb{R}$ be the target function, where $\bm{\mathcal{Z}}\subseteq\mathbb{R}^{n_z}$. Given input $\bm{z}(t)\in\bm{\mathcal{Z}}$ at time $t$, the corresponding output is: 
$
y(t) = \eta(\bm{z}(t))+e(t)$, $ e(t)\sim \mathcal{N}(0, \sigma_e^2)
$, 
where $e(t)$ is the Gaussian measurement noise.
Let training data be in the form $\mathcal{D}\triangleq(\mathcal{Z}, \bm{y})$, where $\mathcal{Z}\triangleq\{\bm{z}(1),\cdots, \bm{z}(n_s)\}$ is the set of input data and $\bm{y}\triangleq[y(1),\cdots, y(n_s)]^T$ is the column vector aggregating the outputs. GPR aims to estimate the function over a set of test data points $\mathcal{Z}_*\subset\bm{\mathcal{Z}}$ using $\mathcal{D}$ by modelling $\eta$ as a sample from a Gaussian process prior.

\begin{algorithm}[t]
\caption{LiDGPR}\label{alg:distributed_GPR}
\begin{algorithmic}[1]
\Procedure{}{}
\State \textbf{Input: } network of agents: $\mathcal{V}$; test inputs: $\mathcal{Z}_*$; common inputs: $\mathcal{Z}_{agg}$; adjacency matrix: $A(t)$; prior mean function: $\mu$; kernel function: $k$; noise variance: $(\sigma^{[i]}_e)^2$ for $i\in\mathcal{V}$.

\State \textbf{Init: } 
$\mathcal{D}(0)=\emptyset$, $\bm{\xi}^{[i]}(0)=\bm{r}^{[i]}_{\bm{\xi}}(-1)=\bm{0}_{|\mathcal{Z}_{agg}|}$,  $\bm{r}^{[i]}_{\bm{\xi}}(0)=\frac{1}{\sigma_f^2}\bm{1}_{|\mathcal{Z}_{agg}|}$, $\bm{\theta}^{[i]}(0)=\bm{r}^{[i]}_{\bm{\theta}}(0)$, $\bm{\lambda}^{[i]}(0)=\bm{r}^{[i]}_{\bm{\lambda}}(0)$, $\sigma_f^2$ satisfying \eqref{eq: sigma_f condition}. \label{ln: dGPR init}
\For{$t=1,2,\cdots$}
\For{$i\in V$}
\State $\mathcal{D}^{[i]}(t)=\mathcal{D}^{[i]}(t-1)\cup(\bm{z}^{[i]}(t),y^{[i]}(t))$ \label{ln:rGPR_data}

\Statex \{Agent-based GPR\}
\State $\check{\bm{\mu}}_{\mathcal{Z}_{*}|\mathcal{D}^{[i]}(t)}, \check{\bm{\sigma}}^2_{\mathcal{Z}_{*}|\mathcal{D}^{[i]}(t)}=\textrm{aGPR}(\mathcal{D}^{[i]}(t))$\label{ln: agent-based GPR}
\Statex \{Distributed GPR\}
\State $\hat{\bm{\mu}}^{[i]}_{\mathcal{Z}_{agg}|\mathcal{D}(t)},(\hat{\bm{\sigma}}^{[i]}_{\mathcal{Z}_{agg}|\mathcal{D}(t)})^2, (\hat{\bm{\sigma}}_{\mathcal{Z}_{agg}|\mathcal{D}(t)}^{ave,[i]})^2=\textrm{dGPR}(\check{\bm{\mu}}_{\mathcal{Z}_{agg}|\mathcal{D}^{[i]}(t)}, \check{\bm{\sigma}}^2_{\mathcal{Z}_{agg}|\mathcal{D}^{[i]}(t)})$ \label{ln:dGPR}
\Statex \{Fused GPR\}
\State $\tilde{\bm{\mu}}^{[i]}_{\mathcal{Z}_{*}|\mathcal{D}(t)}, (\tilde{\bm{\sigma}}^{[i]}_{\mathcal{Z}_{*}|\mathcal{D}(t)})^2 =\textrm{fGPR}(\check{\bm{\mu}}_{\mathcal{Z}_{*}|\mathcal{D}^{[i]}(t)}$, $\check{\bm{\sigma}}^2_{\mathcal{Z}_{*}|\mathcal{D}^{[i]}(t)}$, $\hat{\bm{\mu}}^{[i]}_{\mathcal{Z}_{agg}|\mathcal{D}(t)}$, $(\hat{\bm{\sigma}}^{[i]}_{\mathcal{Z}_{agg}|\mathcal{D}(t)})^2$,   $(\hat{\bm{\sigma}}_{\mathcal{Z}_{agg}|\mathcal{D}(t)}^{ave,[i]})^2$)\label{ln:fGPR}
\EndFor
\EndFor

\EndProcedure
\end{algorithmic}
\end{algorithm}

\begin{definition}
(page 13, \cite{williams2006gaussian}) A Gaussian process is a collection of random variables, any finite number of which have a joint Gaussian distribution.$\hfill\blacksquare$
\end{definition}

Define kernel function $k:\mathbb{R}^{n_z}\times\mathbb{R}^{n_z}\rightarrow \mathbb{R}$ that is symmetric and positive semi-definite; i.e., $\int k(\bm{z},\bm{z}')f(\bm{z})f(\bm{z}')d\nu(\bm{z})d\nu(\bm{z}')\geqslant 0$ for all $f\in L_2(\bm{\mathcal{Z}},\nu)$, where $\nu$ denotes a measure (page 80, \cite{williams2006gaussian}). By modeling $\eta(\cdot)$ as a sample from the Gaussian process prior specified by mean function $\mu(\cdot)$ and kernel $k(\cdot,\cdot)$,
 the training outputs $\bm{y}$ and the test outputs $\bm{\eta}(\mathcal{Z}_*)$ are jointly distributed as:
\begin{equation*}\label{eq:joint_Gaussian}
    \begin{bmatrix}\bm{y}\\\bm{\eta}(\mathcal{Z}_*)
    \end{bmatrix}\sim \mathcal{N}(\begin{bmatrix}\bm{\mu}(\mathcal{Z})\\\bm{\mu}(\mathcal{Z}_*)
    \end{bmatrix},\begin{bmatrix}K(\mathcal{Z},\mathcal{Z})+\sigma_e^2I_{n_s}&K(\mathcal{Z},\mathcal{Z}_*)\\K(\mathcal{Z}_*,\mathcal{Z})&K(\mathcal{Z}_*,\mathcal{Z}_*)
    \end{bmatrix}),
\end{equation*}
where $K(\mathcal{Z},\mathcal{Z}_*)$ returns a matrix such that the entry at the $i^{th}$ row and the $j^{th}$ column is $k(\bm{z}(i),\bm{z}_*(j))$, $\bm{z}_*(j)\in\mathcal{Z}_*$, and analogously for $K(\mathcal{Z},\mathcal{Z})$ and $K(\mathcal{Z}_*,\mathcal{Z}_*)$.

Utilizing identities of joint Gaussian distribution (page 200, \cite{williams2006gaussian}), GPR makes predictions of $\eta$ on $\mathcal{Z}_*$ based on dataset $\mathcal{D}$ as
$
   \bm{\eta}(\mathcal{Z}_*) \sim\mathcal{N}(\bm{\mu}_{\mathcal{Z}_*|\mathcal{D}},\Sigma_{\mathcal{Z}_*|\mathcal{D}})
$,
where
\begin{align}
\bm{\mu}_{\mathcal{Z}_*|\mathcal{D}}&\triangleq\bm{\mu}(\mathcal{Z}_*)+K(\mathcal{Z}_*,\mathcal{Z})\ring{K}(\mathcal{Z},\mathcal{Z})^{-1}(\bm{y}-\bm{\mu}(\mathcal{Z})),\nonumber\\
\Sigma_{\mathcal{Z}_*|\mathcal{D}}&\triangleq K(\mathcal{Z}_*,\mathcal{Z}_*)-K(\mathcal{Z}_*,\mathcal{Z})\ring{K}(\mathcal{Z},\mathcal{Z})^{-1}K(\mathcal{Z},\mathcal{Z}_*),\label{eq:full GPR}
\end{align} 
where $\ring{K}(\mathcal{Z},\mathcal{Z})\triangleq K(\mathcal{Z},\mathcal{Z})+\sigma_e^2I_{n_s}$.
We refer \eqref{eq:full GPR} as full GPR.
If $y$ is multi-dimensional, GPR is performed for each element.


\begin{figure}
    \centering
    \includegraphics[width=0.5\textwidth]{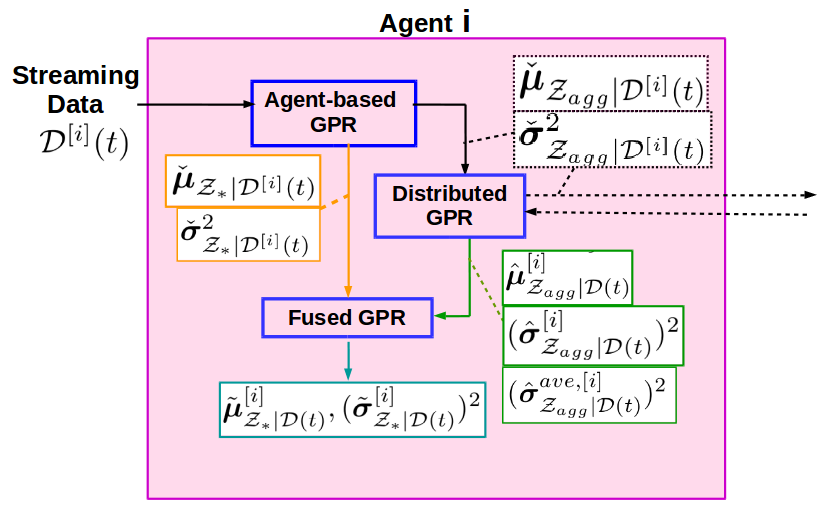}
    \caption{Flow diagram of LiDGPR in one iteration}
    \label{fig:diagram}
\end{figure}

\section{Lightweight distributed GPR}\label{sec:algorithm statement}
In this section, we propose the Lightweight distributed GPR (LiDGPR) algorithm which allows the agents to collaboratively learn the  static latent  function subject to limited resources.   As shown in Figure \ref{fig:diagram},  LiDGPR is composed of three parts: (i) agent-based GPR (Algorithm \ref{alg:GPR}), where the agents make their own predictions of $\eta$ over a given set of points of interest $\mathcal{Z}_*\subseteq\bm{\mathcal{Z}}$ using local streaming data $\mathcal{D}^{[i]}(t)\triangleq(\mathcal{Z}^{[i]}(t),\bm{y}^{[i]}(t))$, where $\mathcal{Z}^{[i]}(t)\triangleq\{\bm{z}^{[i]}(1),\cdots, \bm{z}^{[i]}(t)\}$ aggregates local input data and $\bm{y}^{[i]}(t)\triangleq[y^{[i]}(1),\cdots,y^{[i]}(t)]^T$ aggregates the outputs; (ii) distributed GPR (Algorithm \ref{alg:dGPR}), where the agents integrate their predictions with those of their neighbors on a pre-defined common set $\mathcal{Z}_{agg}\subset\mathcal{Z}_*$ and estimate the predictions on this set given the global training datatset $\mathcal{D}(t)\triangleq \cup_{i\in\mathcal{V}}\mathcal{D}^{[i]}(t)$; (iii) fused GPR (Algorithm \ref{alg:fGPR}), where the agents refine the predictions on $\mathcal{Z}_*$ by fusing the results from distributed GPR with those from agent-based GPR. The formal statement of  LiDGPR is presented in Algorithm \ref{alg:distributed_GPR}.   For each iteration $t$, each agent $i$ collects data online and updates local dataset $\mathcal{D}^{[i]}(t)=\mathcal{D}^{[i]}(t-1)\cup(\bm{z}^{[i]}(t),y^{[i]}(t))$, and then sequentially executes  agent-based GPR, distributed GPR, and fused GPR. 

\begin{algorithm}[t]
\caption{Agent-based GPR}\label{alg:GPR}
\begin{algorithmic}[1]
\Procedure{\lowercase{a}GPR}{$\mathcal{D}^{[i]}(t)$}
\For{$\bm{z}_*\in\mathcal{Z}_*$}
\State choose $\bm{z}^{[i]}_*(t)\in \proj(\bm{z}_*, \mathcal{Z}^{[i]}(t))$\label{eq: z_*(t)}
\State $\check{\mu}_{\bm{z}_*|\mathcal{D}^{[i]}(t)}=\mu(\bm{z}_*)+k(\bm{z}_*,\bm{z}^{[i]}_*(t))\ring{k}(\bm{z}^{[i]}_*(t),\bm{z}^{[i]}_*(t))^{-1}(y_{\bm{z}^{[i]}_*(t)}^{[i]}-\mu(\bm{z}^{[i]}_*(t)))$ \label{ln: agent-based GPR mu}
\State $\check{\sigma}^2_{\bm{z}_*|\mathcal{D}^{[i]}(t)}= k(\bm{z}_*,\bm{z}_*)-k(\bm{z}_*,\bm{z}^{[i]}_*(t))\ring{k}(\bm{z}^{[i]}_*(t),\bm{z}^{[i]}_*(t))^{-1}k(\bm{z}^{[i]}_*(t),\bm{z}_*)$\label{ln: agent-based GPR var}
\EndFor
\State \textbf{Return } $\check{\bm{\mu}}_{\mathcal{Z}_{*}|\mathcal{D}^{[i]}(t)}, \check{\bm{\sigma}}^2_{\mathcal{Z}_{*}|\mathcal{D}^{[i]}(t)}$
\EndProcedure
\end{algorithmic}
\end{algorithm}

\subsection{Agent-based GPR}
To reduce computational complexity, we implement Nearest-neighbor GPR  as  agent-based GPR. Instead of feeding the whole training dataset to full GPR in \eqref{eq:full GPR}, agent-based GPR only feeds the nearest input $\bm{z}^{[i]}_*(t)\in\proj(\bm{z}_*, \mathcal{Z}^{[i]}(t))$, and the corresponding output $y_{\bm{z}^{[i]}_*(t)}^{[i]}$, i.e., $(\bm{z}^{[i]}_*(t), y_{\bm{z}^{[i]}_*(t)}^{[i]})$, to \eqref{eq:full GPR} for each $\bm{z}_*\in\mathcal{Z}_*$. If there are repeated observations over $\bm{z}^{[i]}_*(t)$, $y_{\bm{z}^{[i]}_*(t)}^{[i]}$ can be the average of the observations.
The predictive mean and variance for each $\bm{z}_*$ are given in  Line \ref{ln: agent-based GPR mu} and \ref{ln: agent-based GPR var} of agent-based GPR.  Agent-based GPR returns $\check{\bm{\mu}}_{\mathcal{Z}_{*}|\mathcal{D}^{[i]}(t)}\triangleq [\check{\mu}_{\bm{z}_{*}|\mathcal{D}^{[i]}(t)}]_{\bm{z}_*\in\mathcal{Z}_*}$ and $ \check{\bm{\sigma}}^2_{\mathcal{Z}_{*}|\mathcal{D}^{[i]}(t)}\triangleq [\check{\sigma}^2_{\bm{z}_{*}|\mathcal{D}^{[i]}(t)}]_{\bm{z}_*\in\mathcal{Z}_*}$. 

\subsection{Distributed GPR}\label{sec:dGPR}
Note that each agent only maintains $\mathcal{D}^{[i]}(t)$, a portion of the global training dataset $\mathcal{D}(t)$. Besides collecting more data, information exchanges between the agents could enhance the learning performance upon agent-based GPR. 
However, limited communication budget prevents the agents from sharing $\mathcal{D}^{[i]}(t)$, whose size monotonically increases.
Hence, we develop distributed GPR where the agents communicate with the predictive means and variances over a common set $\mathcal{Z}_{agg}$.

\begin{algorithm}[t]
\caption{Distributed GPR}\label{alg:dGPR}
\begin{algorithmic}[1]
\Procedure{\lowercase{d}GPR}{$\check{\bm{\mu}}_{\mathcal{Z}_{agg}|\mathcal{D}^{[i]}(t)}, \check{\bm{\sigma}}^2_{\mathcal{Z}_{agg}|\mathcal{D}^{[i]}(t)}$}

\For{$\bm{z}_*\in\mathcal{Z}_{agg}$}
\Statex \{Dynamic average consensus\}
\State $r^{[i]}_{\bm{\theta},\bm{z}_*}(t)=\check{\sigma}^{-2}_{\bm{z}_*|\mathcal{D}^{[i]}(t)} \check{\mu}_{\bm{z}_*|\mathcal{D}^{[i]}(t)}$ \label{ln:dynamics consensus_r}
\State $\theta_{\bm{z}_*}^{[i]}(t)=\theta_{\bm{z}_*}^{[i]}(t-1)+\sum_{j\neq i}a_{ij}(t-1)(\theta_{\bm{z}_*}^{[j]}(t-1)-\theta_{\bm{z}_*}^{[i]}(t-1))+\Delta r^{[i]}_{\bm{\theta},\bm{z}_*}(t-1)$\label{ln:dynamic consensus_x}
\State $r^{[i]}_{\bm{\xi},\bm{z}_*}(t)=\check{\sigma}^{-2}_{\bm{z}_*|\mathcal{D}^{[i]}(t)}$ \label{ln:r xi}
\State $\xi_{\bm{z}_*}^{[i]}(t)= \xi_{\bm{z}_*}^{[i]}(t-1)+\sum_{j\neq i}a_{ij}(t-1)(\xi_{\bm{z}_*}^{[j]}(t-1)-\xi_{\bm{z}_*}^{[i]}(t-1))+\Delta r^{[i]}_{\bm{\xi},\bm{z}_*}(t-1)$\label{ln:dynamic consensus_xi}
\State $r^{[i]}_{\bm{\lambda},\bm{z}_*}(t)=\check{\sigma}^{2}_{\bm{z}_*|\mathcal{D}^{[i]}(t)}$
\State  $\lambda_{\bm{z}_*}^{[i]}(t)= \lambda_{\bm{z}_*}^{[i]}(t-1)+\sum_{j\neq i}a_{ij}(t-1)(\lambda_{\bm{z}_*}^{[j]}(t-1)-\lambda_{\bm{z}_*}^{[i]}(t-1))+\Delta r^{[i]}_{\bm{\lambda},\bm{z}_*}(t-1)$\label{ln:dynamic consensus_iota}
\Statex \{Prediction on $\mathcal{Z}_{agg}$\}
\State $({\hat{\sigma}}^{[i]}_{\bm{z}_*|\mathcal{D}(t)})^2
=(\xi_{\bm{z}_*}^{[i]}(t))^{-1}$\label{ln:estimation_sigma_M}
\State $\hat{\mu}^{[i]}_{\bm{z}_*|\mathcal{D}(t)}=(\hat{\sigma}^{[i]}_{\bm{z}_*|\mathcal{D}(t)})^2 \theta_{\bm{z}_*}^{[i]}(t)$ \label{ln:estimation_mu_M}
\State $(\hat{\sigma}_{\bm{z}_{*}|\mathcal{D}(t)}^{ave,[i]})^2=\lambda_{\bm{z}_*}^{[i]}(t)$\label{ln: hat sigma ave}

\EndFor
\State $\textrm{Send }(\bm{\theta}^{[i]}(t),\bm{\xi}^{[i]}(t), \bm{\lambda}^{[i]}(t))$ to $\mathcal{N}^{[i]}(t)$\label{ln:broadcast}
\State \textbf{Return }  $\hat{\bm{\mu}}^{[i]}_{\mathcal{Z}_{agg}|\mathcal{D}(t)},(\hat{\bm{\sigma}}^{[i]}_{\mathcal{Z}_{agg}|\mathcal{D}(t)})^2, (\hat{\bm{\sigma}}_{\mathcal{Z}_{agg}|\mathcal{D}(t)}^{ave,[i]})^2$
\EndProcedure
\end{algorithmic}
\end{algorithm}

In order to deal with large dataset using GPR, local approximation methods such as Product of Expert (PoE) \cite{hinton2002training} and Bayesian Committee Machine \cite{tresp2000bayesian} are proposed to factorize the training process.
We consider the following PoE aggregation model for predicting each $\bm{z}_*\in \mathcal{Z}_{agg}$:
\begin{align}
   &\check{\mu}^{(agg)}_{\bm{z}_*|\mathcal{D}(t)}=\frac{(\check{\sigma}^{(agg)}_{\bm{z}_*|\mathcal{D}(t)})^2}{n}  \sum_{i=1}^n \check{\sigma}^{-2}_{\bm{z}_*|\mathcal{D}^{[i]}(t)}\check{\mu}_{\bm{z}_*|\mathcal{D}^{[i]}(t)},\label{eq:agg_mu}\\
      &(\check{\sigma}^{(agg)}_{\bm{z}_*|\mathcal{D}(t)})^{-2}=\frac{1}{n}\sum_{i=1}^n\check{\sigma}^{-2} _{\bm{z}_*|\mathcal{D}^{[i]}(t)}\label{eq:agg_var},
\end{align}
which is consistent with full GPR (Proposition 2, \cite{liu2018generalized}).

The two summations in \eqref{eq:agg_mu} and \eqref{eq:agg_var} involve the global training dataset. To decentralize the computation, we consider the computation of the two summations as a dynamic average consensus problem and use FODAC in \cite{zhu2010discrete} to track the time-varying sums in a distributed manner.
Denote $\big(\bm{\theta}^{[i]}(t),\bm{\xi}^{[i]}(t), \bm{\lambda}^{[i]}(t)\big)$ the consensus states of agent $i$. 
Each entry of the consensus states, $\big(\theta_{\bm{z}_*}^{[i]}(t),\xi_{\bm{z}_*}^{[i]}(t), \lambda_{\bm{z}_*}^{[i]}(t)\big)$, estimates $\big(\frac{1}{n}\sum_{i=1}^n\check{\sigma}^{-2}_{\bm{z}_*|\mathcal{D}^{[i]}(t)} \check{\mu}_{\bm{z}_*|\mathcal{D}^{[i]}(t)}$, $\frac{1}{n}\sum_{i=1}^n \check{\sigma}^{-2}_{\bm{z}_*|\mathcal{D}^{[i]}(t)}$, $\frac{1}{n}\sum_{i=1}^n\check{\sigma}^{2}_{\bm{z}_*|\mathcal{D}^{[i]}(t)}\big)$ for each $\bm{z}_*\in\mathcal{Z}_{agg}$. 
State $\bm{\lambda}^{[i]}(t)$ is used as one of the criteria for applying fusion between  agent-based GPR and distributed GPR in fused GPR. 
The dynamics of FODAC  is shown in Line \ref{ln:dynamic consensus_x}, Line \ref{ln:dynamic consensus_xi} and Line  \ref{ln:dynamic consensus_iota} of  distributed GPR respectively for each consensus state,
where $\Delta r(t)\triangleq r(t)-r(t{-}1)$ denotes the temporal change of the signal $r$. Specifically, $\theta_{\bm{z}_*}^{[i]}(t)$ tracks the average of the signal $r^{[i]}_{\bm{\theta},\bm{z}_*}(t)$ defined in Line \ref{ln:dynamics consensus_r} among the agents. In particular, agent $i$ 
computes a convex combination of $\theta_{\bm{z}_*}^{[j]}(t)$ for $j\in \{i\}\cup \mathcal{N}^{[i]}(t-1)$, and then adds the combination into the temporal change of $r^{[i]}_{\bm{\theta},\bm{z}_*}(t)$.
The update laws for $\xi_{\bm{z}_*}^{[i]}(t)$  and $\lambda_{\bm{z}_*}^{[i]}(t)$ are similar. 
The updated states are sent to each agent in $\mathcal{N}^{[i]}(t)$ as in Line \ref{ln:broadcast}.  Notice that consensus is not necessarily reached at each time $t$. We will show that consensus is reached in an asymptotic way in Section \ref{sec:analysis}.


\begin{algorithm}[t]
\caption{Fused GPR}\label{alg:fGPR}
\begin{algorithmic}[1]
\Procedure{\lowercase{f}GPR}{$\check{\bm{\mu}}_{\mathcal{Z}_{*}|\mathcal{D}^{[i]}(t)}, \check{\bm{\sigma}}^2_{\mathcal{Z}_{*}|\mathcal{D}^{[i]}(t)}$, $\hat{\bm{\mu}}^{[i]}_{\mathcal{Z}_{agg}|\mathcal{D}(t)},(\hat{\bm{\sigma}}^{[i]}_{\mathcal{Z}_{agg}|\mathcal{D}(t)})^2$,  $(\hat{\bm{\sigma}}_{\mathcal{Z}_{agg}|\mathcal{D}(t)}^{ave,[i]})^2$}
\State $\mathcal{Z}^{[i]}_{agg}(t)=\{\bm{z}_{agg}\in\mathcal{Z}_{agg}|(\hat{\sigma}^{[i]}_{\bm{z}_{agg}|\mathcal{D}(t)})^2<\check{\sigma}^2_{\bm{z}_{agg}|\mathcal{D}^{[i]}(t)}\textrm{ and } (\hat{\sigma}_{\bm{z}_{agg}|\mathcal{D}(t)}^{ave,[i]})^2<\check{\sigma}^{2}_{\bm{z}_{agg}|\mathcal{D}^{[i]}(t)}\}$\label{ln:Z_agg(t)}
\If{$\mathcal{Z}^{[i]}_{agg}(t)==\emptyset$ } \label{ln: empty Z_agg}
\State \textbf{Return } $\tilde{\bm{\mu}}^{[i]}_{\mathcal{Z}_{*}|\mathcal{D}(t)}=\check{\bm{\mu}}_{\mathcal{Z}_{*}|\mathcal{D}^{[i]}(t)},(\tilde{\bm{\sigma}}^{[i]}_{\mathcal{Z}_{*}|\mathcal{D}(t)})^2=\check{\bm{\sigma}}^2_{\mathcal{Z}_{*}|\mathcal{D}^{[i]}(t)}$\label{ln: return local}
\EndIf
\For{$\bm{z}_*\in\mathcal{Z}_{*}$}
\State choose $\bm{z}^{[i]}_{agg*}(t)\in\textrm{proj}(\bm{z}_*,\mathcal{Z}^{[i]}_{agg}(t))$
\State $g^{[i]}_{\bm{z}_*}(t)=g(\bm{z}_*,t)k(\bm{z}_*,\bm{z}^{[i]}_{agg*}(t))
$ \label{ln: g}
\label{ln:fusion Q}
\State $v^{[i]}_{\bm{z}_*}(t)=g^{[i]}_{\bm{z}_*}(t)\check{\sigma}^{-2}_{\bm{z}^{[i]}_{agg*}(t)|\mathcal{D}^{[i]}(t)}$\label{ln: fusion v}
\State $\mu'^{[i]}_{\bm{z}^{[i]}_{agg*}(t)|\mathcal{D}(t)}=\hat{\mu}^{[i]}_{\bm{z}^{[i]}_{agg*}(t)|\mathcal{D}(t)}-\check{\mu}_{\bm{z}^{[i]}_{agg*}(t)|\mathcal{D}^{[i]}(t)}$\label{ln: mu'}
\State $\tilde{\mu}^{[i]}_{\bm{z}_*|\mathcal{D}(t)}= v^{[i]}_{\bm{z}_*}(t) \mu'^{[i]}_{\bm{z}^{[i]}_{agg*}(t)|\mathcal{D}(t)}+\check{\mu}_{\bm{z}_*|\mathcal{D}^{[i]}(t)}$\label{ln:fusion mu}
\State $(\tilde{\sigma}^{[i]}_{\bm{z}_*|\mathcal{D}(t)})^2= \check{\sigma}^2_{\bm{z}_*|\mathcal{D}^{[i]}(t)}+(v_{\bm{z}_*}^{[i]}(t))^2\big((\hat{\sigma}^{[i]}_{\bm{z}^{[i]}_{agg*}(t)|\mathcal{D}(t)})^2-\check{\sigma}^2_{\bm{z}^{[i]}_{agg*}(t)|\mathcal{D}^{[i]}(t)}\big)$ \label{ln:fusion Sigma}
\EndFor
\State \textbf{Return } $\tilde{\bm{\mu}}^{[i]}_{\mathcal{Z}_{*}|\mathcal{D}(t)}, (\tilde{\bm{\sigma}}^{[i]}_{\mathcal{Z}_{*}|\mathcal{D}(t)})^2$
\EndProcedure
\end{algorithmic}
\end{algorithm}

\subsection{Fused GPR}\label{sec:fGPR}
Fused GPR aims to refine predictions of $\bm{\eta}(\mathcal{Z}_*)$ by integrating agent-based GPR with distributed GPR.
The goal is to obtain an estimate of the  predictive distribution $P\{\eta(\bm{z}_*)|\mathcal{D}(t)\}$
for each $\bm{z}_*\in\mathcal{Z}_*$. 
Note that distributed GPR obtains new estimates of $\bm{\eta}(\mathcal{Z}_{agg})$ by combining results from each agent through convex combination.  It can return results with more uncertain predictions, and
these predictions should be ignored.
The set of inputs predicted by distributed GPR with lower uncertainty is defined as $\mathcal{Z}^{[i]}_{agg}(t)$
(Line \ref{ln:Z_agg(t)} of fused GPR). 
Set $\mathcal{Z}^{[i]}_{agg}(t)$ is the set of inputs  $\bm{z}_{agg}\in\mathcal{Z}_{agg}$, where the two variance estimates from  distributed GPR,  the estimates of $(\check{\sigma}^{(agg)}_{\bm{z}_{agg}|\mathcal{D}(t)})^{-2}$ in \eqref{eq:agg_var} and the estimates of $\frac{1}{n}\sum_{i=1}^n\check{\sigma}^2_{\bm{z}_{agg}|\mathcal{D}^{[i]}(t)}$, are lower than $\check{\sigma}^2_{\bm{z}_{agg}|\mathcal{D}^{[i]}(t)}$ from  agent-based GPR. If this set is empty (Line \ref{ln: empty Z_agg}-\ref{ln: return local} in fused GPR), the results from distributed GPR are ignored and those from  agent-based GPR are used.
Otherwise, $P\{\eta(\bm{z}_*)|\mathcal{D}(t)\}$ is estimated as follows.

Notice that for all $\bm{z}_{agg}\in\mathcal{Z}^{[i]}_{agg}(t)$, we have
\begin{align*}
   &P\{\eta(\bm{z}_*)|\mathcal{D}(t)\}=\int P\{\eta(\bm{z}_{agg}), \eta(\bm{z}_*)|\mathcal{D}(t)\} d\eta(\bm{z}_{agg}), \\
   &P\{\eta(\bm{z}_*), \eta(\bm{z}_{agg})|\mathcal{D}(t)\}=P\{\eta(\bm{z}_{agg})|\mathcal{D}(t)\}\\
   &\qquad\qquad\qquad\qquad\quad\qquad\cdot  P\{\eta(\bm{z}_*)|\eta(\bm{z}_{agg}),\mathcal{D}(t)\}.
   \end{align*}
   However, $P\{\eta(\bm{z}_{agg})|\mathcal{D}(t)\}$ and $P\{\eta(\bm{z}_*)|\eta(\bm{z}_{agg}),\mathcal{D}(t)\}$ are unknown but can be estimated using the results from distributed GPR and  agent-based GPR respectively. 
   In particular, the results from distributed GPR are used to estimate $P\{\eta(\bm{z}_{agg})|\mathcal{D}(t)\}$ since the estimate has lower variance (uncertainty). The results from  agent-based GPR are used to estimate $P\{\eta(\bm{z}_*)|\eta(\bm{z}_{agg}),\mathcal{D}(t)\}$ because distributed GPR is limited to $\mathcal{Z}_{agg}$ and $\bm{z}_*$ may not be in $\mathcal{Z}_{agg}$. 
The product of $P\{\eta(\bm{z}_*)|\eta(\bm{z}_{agg}),\mathcal{D}(t)\}$ and $P\{\eta(\bm{z}_{agg})|\mathcal{D}(t)\}$, which yields $P\{\eta(\bm{z}_*), \eta(\bm{z}_{agg})|\mathcal{D}(t)\}$, then contains information from the local agent and those from the other agents in the network. 
The overall process can be interpreted as a fusion of global information with local information, where improvement is expected because more information is provided. After integrating over $\eta(\bm{z}_{agg})$, we obtain the estimate of
 $P\{\eta(\bm{z}_*)|\mathcal{D}(t)\}$. The detailed procedure is broken down into the following steps.
   
{\em Step 1: Estimation of $P\{\eta(\bm{z}_{agg})|\mathcal{D}(t)\}$}.
Consider any  $\bm{z}_{agg}\in\mathcal{Z}_{agg}$.
Agent $i$'s estimate of  $P\{\eta(\bm{z}_{agg})|\mathcal{D}(t)\}$, denoted by $\tilde{P}^{[i]}\{\eta(\bm{z}_{agg})|\mathcal{D}(t)\}\triangleq\mathcal{N}(\hat{\mu}^{[i]}_{\bm{z}_{agg}|\mathcal{D}(t)},(\hat{\sigma}^{[i]}_{\bm{z}_{agg}|\mathcal{D}(t)})^2)$, is given by distributed GPR.  

{\em Step 2: Estimation of $P\{\eta(\bm{z}_*)|\eta(\bm{z}_{agg}),\mathcal{D}(t))\}$}.
Note that agent-based GPR does not return covariance  $\cov(\eta(\bm{z}_*),\eta(\bm{z}_{agg})|\mathcal{D}^{[i]}(t))$ that reflects the correlation between $\eta(\bm{z}_{agg})$ and $\eta(\bm{z}_*)$ similar to $k(\bm{z}_*,\bm{z}_{agg})$. We set  $\cov(\eta(\bm{z}_*),\eta(\bm{z}_{agg})|\mathcal{D}^{[i]}(t))=g(\bm{z}_*,t)k(\bm{z}_*,\bm{z}_{agg})$ and define $$
g(\bm{z}_*,t)\triangleq\frac{\min\{\check{\sigma}^2_{\bm{z}_*|\mathcal{D}^{[i]}(t)}, \check{\sigma}^2_{\bm{z}_{agg}|\mathcal{D}^{[i]}(t)}\}\cdot\max\{0,c-\psi^{[i]}\}}{k(\bm{z}_*,\bm{z}_*)},$$
where $c\triangleq \mu^{-1}_{\chi}\big(\frac{1}{n}\sum_{j=1}^n\chi^{[j]}\psi^{[j]}\big)$, $\psi^{[i]}\triangleq\frac{\sigma_f^2}{\sigma_f^2+(\sigma^{[i]}_e)^2}$, $\chi^{[i]}\triangleq \frac{1}{(\sigma^{[i]}_e)^2}+\frac{1}{\sigma_f^2}$ and $\mu_{\chi}\triangleq\frac{1}{n}\sum_{i=1}^n\chi^{[i]}$. A distributed method for agent $i$ to obtain $(\sigma_e^{[j]})^2$, $j\neq i$, is given in Section \ref{sec: hyperparameter}.
This ensures covariance matrix $$\tilde{\Sigma}_{\bm{z}_*,\bm{z}_{agg}|\mathcal{D}^{[i]}(t)}\triangleq\begin{bmatrix}\check{\sigma}^2_{\bm{z}_*|\mathcal{D}^{[i]}(t)}& g(\bm{z}_*,t)k(\bm{z}_*,\bm{z}_{agg})\\
    g(\bm{z}_*,t)k(\bm{z}_*,\bm{z}_{agg})& \check{\sigma}^2_{\bm{z}_{agg}|\mathcal{D}^{[i]}(t)}
    \end{bmatrix}$$ is positive definite.   We further verify this choice is valid by showing $(\tilde{\sigma}^{[i]}_{\bm{z}_*|\mathcal{D}(t)})^2> 0$ for all $t\geqslant1$ and $\bm{z}_*\in\mathcal{Z}_*$ in Section \ref{analysis: var fGPR}.
We can write 
\begin{align*}
   \tilde{P}\{\eta(\bm{z}_*), \eta(\bm{z}_{agg})&|\mathcal{D}^{[i]}(t)\}\triangleq\\
   & \mathcal{N}(\begin{bmatrix}\check{\mu}_{\bm{z}_*|\mathcal{D}^{[i]}(t)}\\\check{\mu}_{\bm{z}_{agg}|\mathcal{D}^{[i]}(t)} \end{bmatrix},\tilde{\Sigma}_{\bm{z}_*,\bm{z}_{agg}|\mathcal{D}^{[i]}(t)}).
\end{align*}
Then agent $i$'s estimate of $P\{\eta(\bm{z}_*)|\eta(\bm{z}_{agg}),\mathcal{D}(t))\}$, denoted by $\tilde{P}^{[i]}\{\eta(\bm{z}_*)|\eta(\bm{z}_{agg}),\mathcal{D}(t)\}$, is given by $\tilde{P}\{\eta(\bm{z}_*)|\eta(\bm{z}_{agg}),\mathcal{D}^{[i]}(t)\}$ applying identities of joint Gaussian distribution (page 200, \cite{williams2006gaussian}) on $\tilde{P}\{\eta(\bm{z}_*), \eta(\bm{z}_{agg})|\mathcal{D}^{[i]}(t)\}$.

{\em Step 3: Estimation of $P\{\eta(\bm{z}_*)|\mathcal{D}(t)\}$}.
Combining the previous two steps, agent $i$ estimates $P\{\eta(\bm{z}_*),\eta(\bm{z}_{agg})|\mathcal{D}(t)\}$ as 
\begin{align*}
     \tilde{P}^{[i]}\{\eta(\bm{z}_*),\eta(\bm{z}_{agg})|\mathcal{D}(t)\}&=\tilde{P}^{[i]}\{\eta(\bm{z}_{agg})|\mathcal{D}(t)\}\\
     &\quad\cdot  \tilde{P}^{[i]}\{\eta(\bm{z}_*)|\eta(\bm{z}_{agg}),\mathcal{D}(t))\}.
\end{align*}
Applying the same trick of nearest-neighbor prediction as in agent-based GPR, we choose $\bm{z}_{agg}=\bm{z}^{[i]}_{agg*}(t)\in\proj(\bm{z}_*,\mathcal{Z}^{[i]}_{agg}(t))$ for each $\bm{z}_{*}\in\mathcal{Z}_*$. Then we have agent $i$'s estimate of $P\{\eta(\bm{z}_*)|\mathcal{D}(t)\}$ given by
\begin{align*}
   \tilde{P}^{[i]}\{\eta(\bm{z}_*)&|\mathcal{D}(t)\}=\\
   &\int \tilde{P}^{[i]}\{\eta(\bm{z}^{[i]}_{agg*}(t)), \eta(\bm{z}_*)|\mathcal{D}(t)\} d\eta(\bm{z}^{[i]}_{agg*}(t)),
\end{align*}
which has mean and variance in Line \ref{ln:fusion mu}-\ref{ln:fusion Sigma} of fused GPR (see Section \ref{sec: derive fused GPR} for derivation). 

\subsection{Choice of the kernel}\label{sec: hyperparameter}
In this paper, we assume the following properties of the kernel $k$ used in LiDGPR algorithm.
\begin{assumption}\label{assum:kernel}
\begin{enumerate}
\item (Decomposition). The kernel function $k(\cdot,\cdot)$ can be decomposed in such a way that $k(\cdot,\cdot)=\kappa(\rho(\cdot,\cdot))$, where $\kappa:\mathbb{R}_{\geqslant0}\rightarrow \mathbb{R}_{>0}$ is continuous.
    \item(Boundedness). It holds that $0<\kappa(r)\leqslant \sigma_f^2$ for all $r\geqslant0$ and some $\sigma_f>0$.
    \item (Monotonicity). It holds that $\kappa(r)$ is monotonically decreasing as $r$ increases and $\kappa(0)=\sigma_f^2$.
    $\hfill\blacksquare$
\end{enumerate}
\end{assumption}
\begin{remark}\label{remark: GPR kernel}
In GPR, kernel can be interpreted as the prior correlation between function evaluations. For a continuous function, it is reasonable to assume bounded correlation and the correlation is negatively related to the distance between two inputs. One example that satisfies Assumption \ref{assum:kernel} above is the class of squared exponential kernels having the form $k(\bm{z},\bm{z}')=\sigma_f^2\exp(-||\bm{z}-\bm{z}'||^2/\ell^2)$ (page 83, \cite{williams2006gaussian}). $\hfill\blacksquare$
\end{remark}

To obtain the theoretic guarantees in Section \ref{sec: performance},  $\sigma_f^2$ is chosen for initialization as follows. Let
$\sigma^2_{\chi}\triangleq\sum_{i=1}^n(\chi^{[i]}+\mu_{\chi})^2$, 
$\mathcal{V}^+\triangleq\{i\in\mathcal{V}|c\sigma_f^2-\psi^{[i]}>0\}$,  $\epsilon_+\triangleq\min_{i\in\mathcal{V}^+}\{c\sigma_f^2-\psi^{[i]}\}$. We choose $\sigma_f^2\geqslant 1$ satisfying
\begin{align}\label{eq: sigma_f condition}
   \sigma^2_{\chi}/(\mu^2_{\chi}\epsilon_+)\leqslant (\sigma^{\min}_{e})^2/(\sigma^{\max}_{e})^2. 
\end{align}
When $\sigma_f^2$ increases,
$\chi^{[i]}$, $\mu_\chi$ and $\sigma^2_\chi$ converge to positive constants,  $\epsilon_+$ has growth rate $\mathcal{O}(\sigma_f^2)$,  which gives the left hand side of \eqref{eq: sigma_f condition} diminishing at $\mathcal{O}(\frac{1}{\sigma_f^2})$. Hence inequality \eqref{eq: sigma_f condition} is satisfied when $\sigma_f^2$ is sufficiently large. 

A distributed way to choose a single  $\sigma_f^2$  is as follows.  
By using the Floodset algorithm (page 103, \cite{lynch1996distributed}), each agent $i$ sends $(\sigma^{[i]}_e)^2$ to its neighbors. By Assumption \ref{assmp:Periodical Strong Connectivity}, within $n(b-1)$ iterations, each agent obtains a copy of $(\sigma^{[i]}_e)^2$ from all $i\in\mathcal{V}$. Then all the values in \eqref{eq: sigma_f condition} can be calculated. 
To further consider data fitting, each agent can incorporate   \eqref{eq: sigma_f condition} with existing hyperparameter optimization
methods, such as \cite{romeres2016online} which uses a given amount of data points collected during initialization,  or \cite{huber2014recursive} which recursively updates whenever new data arrive. The resulting local hyperparameter of agent $i$ is denoted as $\sigma_f^{[i]}$, then
  all the agents employ maximum consensus  \cite{olfati2004consensus} to compute $\sigma_f=\max_{i\in\mathcal{V}}\{\sigma_f^{[i]}\}$, which terminates in $n(b-1)$ iterations.

\subsection{Performance guarantee}\label{sec: performance}
In this section, we present the performance of predictive mean and variance returned by LiDGPR. The main results are summarized in Theorem \ref{thm: variance convergence} and Theorem \ref{thm: mean convergence}, and their proofs are presented in Section \ref{sec:analysis} and  Section \ref{sec:analysis mean}.

Part of the performance is quantified in terms of the dispersion of local data defined as
$
    d^{[i]}(t)\triangleq \underset{\bm{z}\in\bm{\mathcal{Z}}}{\sup}\rho(\bm{z},\mathcal{Z}^{[i]}(t)).
$
We can interpret dispersion as a measurement of how dense the sampled data is distributed within a compact space. 
For notational simplicity, we introduce shorthand $\rho^{\mathcal{Z}}_{\bm{z}}\triangleq\rho(\bm{z},\mathcal{Z})$. 

Theorem \ref{thm: variance convergence} shows that LiDGPR makes predictions with lower uncertainty than agent-based GPR. 
\begin{theorem}\label{thm: variance convergence} {\em (Uncertainty reduction).}
{\bf Part I}: Suppose Assumption \ref{assum:kernel} holds. For all $\bm{z}_*\in\bm{\mathcal{Z}}$ and $i\in\mathcal{V}$, the predictive variance by agent-based GPR is bounded as
$$
 \frac{\sigma_f^2(\sigma^{[i]}_e)^2}{\sigma_f^2+(\sigma^{[i]}_e)^2}\leqslant\check{\sigma}^2_{\bm{z}_*|\mathcal{D}^{[i]}(t)}\leqslant 
   \sigma_f^2-\frac{\kappa(d^{[i]}(t))^2}{\sigma_f^2+(\sigma^{[i]}_e)^2}.
$$

{\bf Part II}: Suppose Assumptions \ref{assmp:Periodical Strong Connectivity}, \ref{assmp:Balanced Communication}, \ref{assmp:Non-degeneracy} and \ref{assum:kernel} hold.
For all $\bm{z}_*\in\bm{\mathcal{Z}}$ and $i\in\mathcal{V}$, there exists a non-negative sequence  $\gamma^{[i]}_{\sigma,\bm{z}_*}(t)$ such that the predictive variance by  LiDGPR is 
$$
     0<
     (\tilde{\sigma}^{[i]}_{\bm{z}_*|\mathcal{D}(t)})^2= \check{\sigma}^2_{\bm{z}_*|\mathcal{D}^{[i]}(t)}-\gamma^{[i]}_{\sigma,\bm{z}_*}(t). 
$$
In particular, if $\mathcal{Z}^{[i]}_{agg}(t)=\emptyset$, $\gamma^{[i]}_{\sigma,\bm{z}_*}(t)=0$; otherwise :
\begin{align*}
    \gamma^{[i]}_{\sigma,\bm{z}_*}(t)\geqslant &\mathcal{O}\Big(\kappa(\rho_{\bm{z}_*}^{\bm{z}^{[i]}_{agg*}(t)})^2\big(\frac{1}{n}\sum_{j=1}^n\kappa(\rho_{\bm{z}^{[i]}_{agg*}(t)}^{\mathcal{Z}^{[j]}(t)})^2\\
    & -\frac{\sigma_f^2+(\sigma^{\max}_e)^2}{\sigma_f^2+(\sigma^{[i]}_e)^2}\kappa(\rho_{\bm{z}^{[i]}_{agg*}(t)}^{\mathcal{Z}^{[i]}(t)})^2
    \big)\Big). \qquad\qquad\hfill\blacksquare
\end{align*}
\end{theorem}

We provide the steady-state results assuming that the dispersion is diminishing. Lemma 6 in \cite{mueller2013anytime_archive} shows that dispersion does go to zero under uniform sampling.


\begin{corollary}\label{corollary: liminf gamma sigma}
If $\underset{t\to\infty}{\lim}d^{[j]}(t)=0$ for all $j\in\mathcal{V}$ and all the conditions in Theorem \ref{thm: variance convergence} are satisfied, then
$$
    \liminf_{t\to\infty}\gamma^{[i]}_{\sigma,\bm{z}_*}(t)\geqslant \mathcal{O}\Big(\underset{t\to\infty}{\lim} \check{\sigma}^2_{\bm{z}_*|\mathcal{D}^{[i]}(t)}-\frac{1}{n}\sum_{j=1}^n\underset{t\to\infty}{\lim} \check{\sigma}^2_{\bm{z}_*|\mathcal{D}^{[j]}(t)}\Big),
$$
where $\underset{t\to\infty}{\lim} \check{\sigma}^2_{\bm{z}_*|\mathcal{D}^{[j]}(t)}=\frac{\sigma_f^2(\sigma^{[j]}_e)^2}{\sigma_f^2+(\sigma^{[j]}_e)^2}$.$\hfill\blacksquare$
\end{corollary}

To ensure the improvement on prediction accuracy, we need to assume that the prior covariance function of $\eta$ is correctly specified. Note that any non-zero mean Gaussian process can be decomposed into a deterministic process plus a zero-mean stochastic process such that GPR can be performed over the zero-mean stochastic process (page 27, \cite{williams2006gaussian}). Therefore, without loss of generality, we assume $\eta$ follows a zero-mean Gaussian process for notational simplicity.
\begin{assumption}\label{assmp: covariance}

It satisfies that $\eta\sim\mathcal{GP}(0,k)$. $\hfill\blacksquare$
\end{assumption}
That is, the target function $\eta$ is completely specified by a zero-mean Gaussian process with kernel $k$. This assumption is common in the analysis of GPR (Theorem 1, \cite{srinivas2012information}). 

Furthermore, we need to assume that the state transition matrix induced by $A(t)$ is constant.
\begin{assumption}\label{assmp: constant A}
It holds that $\prod_{\tau=1}^tA(\tau)=\prod_{\tau=1}^{t'}A(\tau)$ for any $t,t'>1$. $\hfill\blacksquare$
\end{assumption}
One example that satisfies this assumption is each entry of $A(t)$ being constant $\frac{1}{n}$, which is a complete graph.

Furthermore, we assume $\eta$ is Lipschitz continuous.
\begin{assumption}\label{assmp:lipschitz}
There exists some positive constant $\ell_\eta\in\mathbb{R}$ such that $\sup_{\bm{z},\bm{z}'\in\bm{\mathcal{Z}}}|\eta(\bm{z})-\eta(\bm{z}')|\leqslant\ell_\eta\rho(\bm{z},\bm{z}')$. $\hfill\blacksquare$
\end{assumption}

Theorem \ref{thm: mean convergence} below compares the predictive errors of  agent-based GPR with those of LiDGPR.

\begin{theorem}\label{thm: mean convergence}
{\em (Accuracy improvement).}
{\bf Part I}: Suppose Assumptions \ref{assum:kernel}, \ref{assmp: covariance} and \ref{assmp:lipschitz}  hold. For all $\bm{z}_*\in\bm{\mathcal{Z}}$ and $i\in\mathcal{V}$, with probability at least $1-\frac{(\sigma^{\max}_e)^2}{\epsilon^2}$, $\epsilon>\sigma^{\max}_e$, the predictive error resulted from agent-based GPR is bounded as
$$
|\check{\mu}_{\bm{z}_*|\mathcal{D}^{[i]}(t)}-\eta(\bm{z}_*)|\leqslant (1-\frac{\kappa(d^{[i]}(t))}{\sigma_f^2+(\sigma^{[i]}_e)^2})\|\eta\|_{\bm{\mathcal{Z}}}+\ell_\eta d^{[i]}(t)+\epsilon.
$$

{\bf Part II}: Suppose $\underset{t\to\infty}{\lim}d^{[i]}(t)=0$, $\forall i\in\mathcal{V}$, and Assumptions \ref{assmp:Periodical Strong Connectivity}, \ref{assmp:Balanced Communication}, \ref{assmp:Non-degeneracy}, \ref{assum:kernel}, \ref{assmp: covariance}, \ref{assmp: constant A} and \ref{assmp:lipschitz} hold. For all $\bm{z}_*\in\bm{\mathcal{Z}}$ and $i\in\mathcal{V}$: if $\underset{t\to\infty}{\lim}\mathcal{Z}^{[i]}_{agg}(t)\neq\emptyset$, then  
\begin{align*}
    \underset{t\to\infty}{\lim}\mathbb{E}[(\tilde{\mu}^{[i]}_{\bm{z}_*|\mathcal{D}(t)}-\eta(\bm{z}_*))^2&-(\check{\mu}_{\bm{z}_*|\mathcal{D}^{[i]}(t)}-\eta(\bm{z}_*))^2]\\
    &\leqslant-\mathcal{O}\Big(\kappa(\rho_{\bm{z}_*}^{\mathcal{Z}_{agg}})\Big)<0;
\end{align*}otherwise, $$\displaystyle\underset{t\to\infty}{\lim}(\tilde{\mu}^{[i]}_{\bm{z}_*|\mathcal{D}(t)}-\eta(\bm{z}_*))^2=\underset{t\to\infty}{\lim}(\check{\mu}_{\bm{z}_*|\mathcal{D}^{[i]}(t)}-\eta(\bm{z}_*))^2.$$
Further, if $\displaystyle(\sigma^{[i]}_e)^2>\frac{1}{n}\sum_{j=1}^n(\sigma^{[j]}_e)^2$,  $\underset{t\to\infty}{\lim}\mathcal{Z}^{[i]}_{agg}(t)=\mathcal{Z}_{agg}$.
 $\hfill\blacksquare$

\end{theorem}

 The two theorems indicate that LiDGPR leverages inter-agent communication to improve transient and steady-state learning performance; meanwhile,  no agent suffers from degraded learning performance.
This improvement of learning performance is achieved by the fact that the agents whose data samples have higher dispersion (or observation noise has higher variance) benefit from those with data samples having lower dispersion (or observation noise having lower variance) via communication.  Next we elaborate on the fact.

{\em Transient improvement}. Term $\frac{1}{n}\sum_{j=1}^n\kappa(\rho_{\bm{z}^{[i]}_{agg*}(t)}^{\mathcal{Z}^{[j]}(t)})^2 -\frac{\sigma_f^2+(\sigma^{\max}_e)^2}{\sigma_f^2+(\sigma^{[i]}_e)^2}\kappa(\rho_{\bm{z}^{[i]}_{agg*}(t)}^{\mathcal{Z}^{[i]}(t)})^2$ in the lower bound of $\gamma^{[i]}_{\sigma,\bm{z}_*}(t)$ in Theorem \ref{thm: variance convergence} indicates that agent $i$ benefits in variance prediction when $\frac{\sigma_f^2+(\sigma^{\max}_e)^2}{\sigma_f^2+(\sigma^{[i]}_e)^2}\kappa(\rho_{\bm{z}^{[i]}_{agg*}(t)}^{\mathcal{Z}^{[i]}(t)})^2$ is below  $\frac{1}{n}\sum_{j=1}^n\kappa(\rho_{\bm{z}^{[i]}_{agg*}(t)}^{\mathcal{Z}^{[j]}(t)})^2$. 
Note that $\rho_{\bm{z}^{[j]}_{agg*}(t)}^{\mathcal{Z}^{[i]}(t)}$ is closely related to dispersion $d^{[j]}(t)$ and recall the monotonicity property of $\kappa$.
Hence, LiDGPR enables agents whose data samples have higher dispersion (data sparsely sampled) and observation noise has higher variance to benefit from those with data samples having lower dispersion (data densely sampled) and observation noise having lower variance.

{\em Steady-state improvement}. If $\underset{t\to\infty}{\liminf}\gamma^{[i]}_{\sigma,\bm{z}_*}(t)>0$, it indicates that agent $i$ obtains improvement in steady-state learning performance in predictive variance.
From Corollary \ref{corollary: liminf gamma sigma}, we can see that  $\underset{t\to\infty}{\liminf}\gamma^{[i]}_{\sigma,\bm{z}_*}(t)>0$, if its steady-state local predictive variance, $\underset{t\to\infty}{\lim} \check{\sigma}^2_{\bm{z}_*|\mathcal{D}^{[i]}(t)}$, is above the average over the agents in $\mathcal{V}$. By Corollary \ref{corollary: liminf gamma sigma}, $\underset{t\to\infty}{\lim} \check{\sigma}^2_{\bm{z}_*|\mathcal{D}^{[i]}(t)}$ is positively related to $(\sigma_e^{[i]})^2$. Hence agents with observation noise of higher variance might obtain steady-state improvement in predictive variance from those with lower variance.

Steady-state improvement in prediction accuracy is reflected by the case $\underset{t\to\infty}{\lim}\mathcal{Z}^{[i]}_{agg}(t)\neq\emptyset$ in Theorem \ref{thm: mean convergence}. The sufficient condition  $(\sigma^{[i]}_e)^2>\frac{1}{n}\sum_{j=1}^n(\sigma^{[j]}_e)^2$ indicates that  agent $i$ obtains steady-state improvement when $(\sigma_e^{[i]})^2$ is above the average.
That is,  agents with observation noise of higher variance benefits from those with smaller variance.

The improvements  $\gamma^{[i]}_{\sigma,\bm{z}_*}(t)$ and $\underset{t\to\infty}{\lim}\mathbb{E}[(\tilde{\mu}^{[i]}_{\bm{z}_*|\mathcal{D}(t)}-\eta(\bm{z}_*))^2-(\check{\mu}_{\bm{z}_*|\mathcal{D}^{[i]}(t)}-\eta(\bm{z}_*))^2]$ are positively related to $\kappa(\rho_{\bm{z}_*}^{\bm{z}^{[i]}_{agg*}(t)})^2$ and $\kappa(\rho_{\bm{z}_*}^{\mathcal{Z}_{agg}})$  respectively. By monotonicity of $\kappa$ in Assumption \ref{assum:kernel}, these terms indicate that the benefit brought by communication decays as $\bm{z}_*$ is moving away from $\bm{z}^{[i]}_{agg*}(t)$ and $\mathcal{Z}_{agg}$ respectively. That is, a denser set $\mathcal{Z}_{agg}$ could induce larger improvements.

\subsection{Discussion}\label{sec: discussion}

\textit{Relevance:} The two theorems indicate that both prediction uncertainties and prediction errors  reduce as local dispersion $d^{[i]}(t)$ reduces. This provides insights on data sampling such that the agents should sample in a way that minimizes $d^{[i]}(t)$. The terms $\kappa(\rho_{\bm{z}_*}^{\bm{z}^{[i]}_{agg*}(t)})^2$ and $\kappa(\rho_{\bm{z}_*}^{\mathcal{Z}_{agg}})$ in Theorem \ref{thm: variance convergence} and Theorem \ref{thm: mean convergence} show that the improvement of learning performances obtained from communication decreases as the test point $\bm{z}_*$  is moving away from $\mathcal{Z}_{agg}$.  This can guide the design process  of $\mathcal{Z}_{agg}$ such that if the test points in $\mathcal{Z}_*$ are known a priori, $\mathcal{Z}_{agg}$ should be allocated such that $\sup_{\bm{z}_*\in\mathcal{Z}_*}\min_{\bm{z}_{agg}\in\mathcal{Z}_{agg}}\rho(\bm{z}_*,\bm{z}_{agg})$ is minimized; otherwise $\mathcal{Z}_{agg}$ should be designed such that $\sup_{\bm{z}_*\in\bm{\mathcal{Z}}}\min_{\bm{z}_{agg}\in\mathcal{Z}_{agg}}\rho(\bm{z}_*,\bm{z}_{agg})$ is minimized.

{\em Complexities related to $\mathcal{Z}_*$ and $\mathcal{Z}_{agg}$.}
The communication overhead scales as $\mathcal{O}(|\mathcal{Z}_{agg}||\mathcal{N}^{[i]}(t)|)$.
Due to the use of Nearest-neighbor GPR, agent-based GPR only requires $\mathcal{O}(t)$ in memory. The memory requirements for both distributed GPR and fused GPR are $\mathcal{O}(|\mathcal{Z}_{agg}|)$.  The computational complexities scale as $\mathcal{O}(t|\mathcal{Z}_*|)$ for agent-based GPR, $\mathcal{O}(|\mathcal{Z}_{agg}|)$ for distributed GPR, and $\mathcal{O}(|\mathcal{Z}_*||\mathcal{Z}_{agg}|)$ for fused GPR. 

{\em Nearest-neighbor GPR vs. full GPR.}
Part I of Theorem \ref{thm: variance convergence} and Theorem \ref{thm: mean convergence} characterize the  steady-state errors of agent-based GPR.
Paper \cite{choi2007posterior} shows that $\sigma^2_{\bm{z}_*\mid \mathcal{D}^{[i]}(t)}\to (\sigma^{[i]}_e)^2$ and $\mu_{\bm{z}_*\mid \mathcal{D}^{[i]}(t)}\to\eta(\bm{z}_*)$ almost surely as $t\to\infty$ for full GPR.  Part I of Theorem \ref{thm: variance convergence} indicates $\check{\sigma}^2_{\bm{z}_*\mid \mathcal{D}^{[i]}(t)} \to \frac{\sigma_f^2(\sigma^{[i]}_e)^2}{\sigma_f^2+(\sigma^{[i]}_e)^2}$, hence the variance for noisy prediction (page 19 \cite{williams2006gaussian}) equals $\frac{\sigma_f^2(\sigma^{[i]}_e)^2}{\sigma_f^2+(\sigma^{[i]}_e)^2}+(\sigma^{[i]}_e)^2$, and Theorem \ref{thm: mean convergence} indicates 
$$
\underset{t\to\infty}{\limsup}|\check{\mu}_{\bm{z}_*|\mathcal{D}^{[i]}(t)}-\eta(\bm{z}_*)|\leqslant \frac{(\sigma^{[i]}_e)^2}{\sigma_f^2+(\sigma^{[i]}_e)^2}\|\eta\|_{\bm{\mathcal{Z}}}+\epsilon,
$$
assuming $d^{[i]}(t)\to 0$. The discrepancy can be caused by the fact that full GPR in \cite{choi2007posterior} makes prediction using all the data in the dataset while Nearest-neighbor GPR only uses the data of the nearest input. Full GPR has computational complexity $\mathcal{O}(t^3)$ while Nearest-neighbor GPR has the same computational complexity as nearest neighbor search, which is $\mathcal{O}(t)$ for the worst case \cite{abbasifard2014survey}. This is the trade-off between learning accuracy and computational complexity. Note that both full GPR and Nearest-neighbor GPR have the same steady-state errors under noise-free condition, i.e., $(\sigma^{[i]}_e)^2=0$.



\begin{table*}[t]
\begin{center}
\begin{tabular}{ | m{8em} | m{25em}|m{25em}|} 
\hline
Symbol & Meaning/definition & Equivalence\\
\hline
$\check{\mu}_{\mathbf{z}_*|\mathcal{D}^{[i]}(t)}$ & Predictive mean  from agent-based GPR of agent $i$ & $
    \check{\mu}_{\bm{z}_*|\mathcal{D}^{[i]}(t)}=\check{r}^{[i]}_{\bm{z}_*}(t)+\check{e}^{[i]}_{\bm{z}_*}(t)$\\
\hline
$\hat{\mu}^{[i]}_{\mathbf{z}_*|\mathcal{D}(t)}$ & Predictive mean  from distributed GPR of agent $i$ & $\hat{\mu}_{\bm{z}_*|\mathcal{D}(t)}^{[i]}
=\tilde{r}^{[i]}_{\bm{z}_*}(t)+\tilde{e}^{[i]}_{\bm{z}_*}(t)$\\
\hline
$\tilde{\mu}^{[i]}_{\mathbf{z}_*|\mathcal{D}(t)}$&Predictive mean  from fused GPR of agent $i$ &\\

\hline
$\check{\sigma}^2_{\bm{z}_*|\mathcal{D}^{[i]}(t)}$ &Predictive variance  from agent-based GPR of agent $i$ & $\check{\sigma}^2_{\bm{z}_*|\mathcal{D}^{[i]}(t)}= \sigma_f^2-\frac{\kappa(\rho_{\bm{z}_*}^{\mathcal{Z}^{[i]}(t)})^2}{\sigma_f^2+(\sigma^{[i]}_e)^2}$\\
\hline
$(\hat{\sigma}^{[i]}_{\bm{z}_*|\mathcal{D}(t)})^2$ &Predictive variance  from distributed GPR of agent $i$ & $({\hat{\sigma}}^{[i]}_{\bm{z}_*|\mathcal{D}(t)})^2
=(\xi_{\bm{z}_*}^{[i]}(t))^{-1}$ \\
\hline
$(\tilde{\sigma}^{[i]}_{\mathbf{z}_*|\mathcal{D}(t)})^2$&Predictive variance  from fused GPR of agent $i$ &\\
\hline
$r^{[i]}_{\bm{\theta},\bm{z}_*}(t)$& Reference signal for consensus state $\bm{\theta}$ in distributed GPR& $r^{[i]}_{\bm{\theta},\bm{z}_*}(t)=\hat{r}^{[i]}_{\bm{z}_*}(t)+\hat{e}^{[i]}_{\bm{z}_*}(t)$\\

\hline
$r^{[i]}_{\bm{\xi},\bm{z}_*}(t)$& Reference signal for consensus state $\bm{\xi}$ in distributed GPR&\\

\hline
$r^{[i]}_{\bm{\lambda},\bm{z}_*}(t)$& Reference signal for consensus state $\bm{\lambda}$ in distributed GPR&\\
\hline
$\check{r}^{[i]}_{\bm{z}_*}(t)$& Real-valued component of $\check{\mu}_{\bm{z}_*|\mathcal{D}^{[i]}(t)}$ & $\check{r}^{[i]}_{\bm{z}_*}(t)=\frac{\kappa(\rho_{\bm{z}_*}^{\mathcal{Z}^{[i]}(t)})\eta(\bm{z}^{[i]}_*(t))}{\sigma_f^2+(\sigma^{[i]}_e)^2}$ \\

\hline
$\check{r}^{[i]}_{\bm{z}_*}$& $\check{r}^{[i]}_{\bm{z}_*}\triangleq\lim_{t\to\infty}\check{r}^{[i]}_{\bm{z}_*}(t)$ &
$\check{r}^{[i]}_{\bm{z}_*} =\frac{\sigma_f^2\eta(\bm{z}_*)}{\sigma_f^2+(\sigma^{[i]}_e)^2}=\psi^{[i]}\eta(\bm{z}_*)$\\
\hline
$\hat{r}^{[i]}_{\bm{z}_*}(t)$& Real-valued component of $r^{[i]}_{\bm{\theta},\bm{z}_*}(t)$&
$\hat{r}^{[i]}_{\bm{z}_*}(t)=\check{\sigma}^{-2}_{\bm{z}_*|\mathcal{D}^{[i]}(t)}\check{r}^{[i]}_{\bm{z}_*}(t)$ \\
\hline
$\hat{r}^{[i]}_{\bm{z}_*}$& $\hat{r}^{[i]}_{\bm{z}_*}\triangleq\lim_{t\to\infty}\hat{r}^{[i]}_{\bm{z}_*}(t)$& $\hat{r}^{[i]}_{\bm{z}_*}=(\frac{\sigma_f^2(\sigma^{[i]}_e)^2}{\sigma_f^2+(\sigma^{[i]}_e)^2})^{-1}\check{r}^{[i]}_{\bm{z}_*}$\\
\hline  
$\tilde{r}^{[i]}_{\bm{z}_*}(t)$&Real-valued component of $\hat{\mu}^{[i]}_{\mathbf{z}_*|\mathcal{D}(t)}$& $\tilde{r}^{[i]}_{\bm{z}_*}(t)=(\hat{\sigma}^{[i]}_{\bm{z}_*|\mathcal{D}(t)})^2{\theta}^{[i]}_{\bm{z}_*,\bm{r}}(t)$ \\
\hline
$e_{\bm{z}_*}^{[i]}$&Observation error at $\bm{z}_*$&$e_{\bm{z}_*}^{[i]}= y^{[i]}_{\bm{z}_*}-\eta(\bm{z}_*)$\\ 
\hline
$\check{e}^{[i]}_{\bm{z}_*}(t)$& Stochastic component of  $\check{\mu}_{\bm{z}_*|\mathcal{D}^{[i]}(t)}$ &
$\check{e}^{[i]}_{\bm{z}_*}(t)= \frac{\kappa(\rho_{\bm{z}_*}^{\mathcal{Z}^{[i]}(t)})}{\sigma_f^2+(\sigma^{[i]}_e)^2} \big(y^{[i]}_{\bm{z}^{[i]}_*(t)}-\eta(\bm{z}^{[i]}_*(t))\big)$\\
\hline
$\check{e}^{[i]}_{\bm{z}_*}$& $\check{e}^{[i]}_{\bm{z}_*}\triangleq\lim_{t\to\infty}\check{e}^{[i]}_{\bm{z}_*}(t)$&$\check{e}^{[i]}_{\bm{z}_*}= \frac{\sigma_f^2}{\sigma_f^2+(\sigma^{[i]}_e)^2}e_{\bm{z}_*}^{[i]}=\psi^{[i]}e_{\bm{z}_*}^{[i]}$\\
\hline
$\hat{e}^{[i]}_{\bm{z}_*}(t)$&Stochastic component of $r^{[i]}_{\bm{\theta},\bm{z}_*}(t)$&
$\hat{e}^{[i]}_{\bm{z}_*}(t)=\check{\sigma}^{-2}_{\bm{z}_*|\mathcal{D}^{[i]}(t)}\check{e}^{[i]}_{\bm{z}_*}(t)$\\ 
\hline

$\tilde{e}^{[i]}_{\bm{z}_*}(t)$&Stochastic component of $\hat{\mu}^{[i]}_{\mathbf{z}_*|\mathcal{D}(t)}$&
$\tilde{e}^{[i]}_{\bm{z}_*}(t)= (\hat{\sigma}^{[i]}_{\bm{z}_*|\mathcal{D}(t)})^2{\theta}^{[i]}_{\bm{z}_*,\bm{e}}(t)$\\
\hline
$\chi^{[i]}$& $\chi^{[i]}\triangleq \frac{1}{(\sigma^{[i]}_e)^2}+\frac{1}{\sigma_f^2}$ &$\chi^{[i]}=\underset{t\to\infty}{\lim}\check{\sigma}^{-2}_{\bm{z}_*|\mathcal{D}^{[i]}(t)}$\\
\hline
$\psi^{[i]}$&$\psi^{[i]}\triangleq\frac{\sigma_f^2}{\sigma_f^2+(\sigma^{[i]}_e)^2}$&\\

\hline
$\mu_{\chi}$& $\mu_{\chi}\triangleq\frac{1}{n}\sum_{i=1}^n\chi^{[i]}$ &$\mu^{-1}_{\chi}=\underset{t\to\infty}{\lim}(\hat{\sigma}^{[i]}_{\bm{z}_*|\mathcal{D}(t)})^2=\underset{t\to\infty}{\lim}(\check{\sigma}^{(agg)}_{\bm{z}_*|\mathcal{D}(t)})^2$\\
\hline
$c$&  $c\triangleq \mu^{-1}_{\chi}\big(\frac{1}{n}\sum_{j=1}^n\chi^{[j]}\psi^{[j]}\big)$& \\
\hline
$\sigma^2_{\chi}$ &$\sigma^2_{\chi}\triangleq\sum_{i=1}^n(\chi^{[i]}+\mu_{\chi})^2$&\\ 
\hline
$\epsilon_+$&$\epsilon_+\triangleq\min_{i\in\mathcal{V}^+}\{c\sigma_f^2-\psi^{[i]}\}$&\\
\hline
\end{tabular}
\end{center}
\captionsetup{skip=0pt}
\caption{Table of symbols}
\label{table: symbol}
\vspace{-7mm}
\end{table*}

\section{Proofs}
In this section, we  present the derivation of  Line \ref{ln:fusion mu}-\ref{ln:fusion Sigma} in fused GPR and the proofs of Theorem \ref{thm: variance convergence} and Theorem \ref{thm: mean convergence}.
 Table \ref{table: symbol} shows the symbols that are used in multiple important results and the relation among them.

\subsection{Derivation of  Line \ref{ln:fusion mu}-\ref{ln:fusion Sigma} in fused GPR}\label{sec: derive fused GPR}
     Recall that $\tilde{P}^{[i]}\{\eta(\bm{z}_*)|\eta(\bm{z}_{agg}),\mathcal{D}(t))\}$ is given by applying identities of joint Gaussian distribution (page 200, \cite{williams2006gaussian}) to 
     $$\tilde{P}\{\begin{bmatrix}\eta(\bm{z}_*)\\ \eta(\bm{z}_{agg})\end{bmatrix}|\mathcal{D}^{[i]}(t)\}=\mathcal{N}(\begin{bmatrix}\check{\mu}_{\bm{z}_*|\mathcal{D}^{[i]}(t)}\\\check{\mu}_{\bm{z}_{agg}|\mathcal{D}^{[i]}(t)} \end{bmatrix},\tilde{\Sigma}_{\bm{z}_*,\bm{z}_{agg}|\mathcal{D}^{[i]}(t)}).$$ This gives $\tilde{P}^{[i]}\{\eta(\bm{z}_*)|\eta(\bm{z}_{agg}),\mathcal{D}(t))\}$ as a Gaussian distribution with mean and variance 
     \begin{align*}
       &\tilde{\mu}^{[i]}_{\bm{z}_*|\bm{z}_{agg},\mathcal{D}(t)}=\check{\mu}_{\bm{z}_*|\mathcal{D}^{[i]}(t)}\\
       &+g(\bm{z}_*,t)k(\bm{z}_*,\bm{z}_{agg}) \check{\sigma}^{-2}_{\bm{z}_{agg}|\mathcal{D}^{[i]}(t)}(\eta(\bm{z}_{agg})-\check{\mu}_{\bm{z}_{agg}|\mathcal{D}^{[i]}(t)}), \\
       &(\tilde{\sigma}^{[i]}_{\bm{z}_*|\bm{z}_{agg},\mathcal{D}(t)})^2=\check{\sigma}^2_{\bm{z}_*|\mathcal{D}^{[i]}(t)}\\
       &-g(\bm{z}_*,t)k(\bm{z}_*,\bm{z}_{agg}) \check{\sigma}^{-2}_{\bm{z}_{agg}|\mathcal{D}^{[i]}(t)}k(\bm{z}_*,\bm{z}_{agg})g(\bm{z}_*,t).    
       \end{align*}
     
     Notice that the mean and variance of $\tilde{P}^{[i]}\{\eta(\bm{z}_{agg})|\mathcal{D}(t)\}$ is given in distributed GPR. 
Then we have the product 
\begin{align*}
     &\tilde{P}^{[i]}\{\eta(\bm{z}_*),\eta(\bm{z}_{agg})|\mathcal{D}(t)\}\\
     &=\tilde{P}^{[i]}\{\eta(\bm{z}_{agg})|\mathcal{D}(t)\}\cdot  \tilde{P}^{[i]}\{\eta(\bm{z}_*)|\eta(\bm{z}_{agg}),\mathcal{D}(t))\}\\
     &=\mathcal{N}(\hat{\mu}^{[i]}_{\bm{z}_{agg}|\mathcal{D}(t)},(\hat{\sigma}^{[i]}_{\bm{z}_{agg}|\mathcal{D}(t)})^2)\\
     &\quad \cdot\mathcal{N}(\tilde{\mu}^{[i]}_{\bm{z}_*|\bm{z}_{agg},\mathcal{D}(t)},(\tilde{\sigma}^{[i]}_{\bm{z}_*|\bm{z}_{agg},\mathcal{D}(t)})^2).
     \end{align*} 
     After some basic algebraic manipulations (finding the corresponding terms in (A.6) \cite{williams2006gaussian}) or directly plugging the terms in equation (9) of \cite{huber2014recursive}, we have
     \begin{align*}
        &\tilde{P}^{[i]}\{\eta(\bm{z}_*),\eta(\bm{z}_{agg})|\mathcal{D}(t)\}=\\
        &\mathcal{N}(\begin{bmatrix}\tilde{\mu}^{[i]}_{\bm{z}_*|\mathcal{D}(t)}\\\tilde{\mu}^{[i]}_{\bm{z}_{agg}|\mathcal{D}(t)} \end{bmatrix}, \begin{bmatrix}(\tilde{\sigma}^{[i]}_{\bm{z}_*|\mathcal{D}(t)})^2&\tilde{\sigma}^{[i]}_{\bm{z}_*, \bm{z}_{agg}|\mathcal{D}(t)} \\\tilde{\sigma}^{[i]}_{\bm{z}_{agg}, \bm{z}_*|\mathcal{D}(t)}& (\tilde{\sigma}^{[i]}_{\bm{z}_{agg}|\mathcal{D}(t)})^2 \end{bmatrix}).
     \end{align*}
    Replacing $\bm{z}_{agg}$ with $\bm{z}^{[i]}_{agg*}(t)$, $\tilde{\mu}^{[i]}_{\bm{z}_*|\mathcal{D}(t)}$ and $(\tilde{\sigma}^{[i]}_{\bm{z}_*|\mathcal{D}(t)})^2$ have the forms in Line \ref{ln:fusion mu}-\ref{ln:fusion Sigma} in fused GPR.
       Hence we have the marginal distribution $\tilde{P}^{[i]}\{\eta(\bm{z}_*)|\mathcal{D}(t)\}$.

\subsection{Proof of Theorem \ref{thm: variance convergence}}\label{sec:analysis}
In this section, we first derive the lower bound and the upper bound of the predictive variance of agent-based GPR and prove Part I of Theorem \ref{thm: variance convergence} in Section \ref{sec:var agent-based GPR}. Then we derive the bounds of distributed GPR in Proposition \ref{prop:rate of convergence of FODAC}  in Section \ref{sec: var distributed GPR}. Lastly, we derive the bounds of fused GPR and prove Part II of Theorem \ref{thm: variance convergence} in Section \ref{analysis: var fGPR}. 

First of all, we introduce some properties of functions $f_1:\mathbb{R}_{>0}\to\mathbb{R}_{>0}$ as $f_1(x)=\frac{1}{x}$ and 
$f_2:\mathbb{R}_{>0}\to\mathbb{R}_{>0}$ as $f_2(x)=\sigma_f^2-\frac{\sigma_f^4}{\sigma_f^2+x}$. These will be used in later analysis.
\begin{lemma}\label{lmm: f}
It holds that 
\begin{align*}
    f_1(\frac{1}{n}\sum_{i=1}^nx_i)&\leqslant \frac{1}{n}\sum_{i=1}^nf_1(x_i),\\
    f_2(\frac{1}{n}\sum_{i=1}^nx_i)&\geqslant\frac{1}{n}\sum_{i=1}^nf_2(x_i),\\
    \big(\frac{1}{n}\sum_{i=1}^nf_2(x_i)\big)^{-1}&\leqslant \frac{1}{n}\sum_{i=1}^n\big(f_2(x_i)\big)^{-1}.
\end{align*}

{\bf Proof: }
It is obvious that $f_1$ is convex. Then Jensen's inequality (page 77, \cite{boyd2004convex}) gives $f_1(\frac{1}{n}\sum_{i=1}^nx_i)\leqslant \frac{1}{n}\sum_{i=1}^nf_1(x_i)$.

It is obvious that $f_2$ is concave. By Jensen's inequality and concavity, we have $f_2(\frac{1}{n}\sum_{i=1}^nx_i)\geqslant\frac{1}{n}\sum_{i=1}^nf_2(x_i)$.

Applying Jensen's inequality utilizing the monotonicity and the convexity of inverse function $f_3(x)=\frac{1}{x}$ for $x>0$, we can also obtain $\big(\frac{1}{n}\sum_{i=1}^nf_2(x_i)\big)^{-1}\leqslant \frac{1}{n}\sum_{i=1}^n\big(f_2(x_i)\big)^{-1}$. $\hfill\blacksquare$
\end{lemma}


\subsubsection{Variance analysis of agent-based GPR}\label{sec:var agent-based GPR}

In this section, we present the proof of Theorem  \ref{thm: variance convergence} Part I.

\textbf{Proof of Theorem  \ref{thm: variance convergence} Part I:}
Pick any $\bm{z}_*\in\bm{\mathcal{Z}}$. By monotonicity of $\kappa$ in Assumption \ref{assum:kernel}, Line \ref{ln: agent-based GPR var} in agent-based GPR gives the predictive variance $\check{\sigma}^2_{\bm{z}_*|\mathcal{D}^{[i]}(t)}= \sigma_f^2-\frac{k(\bm{z}_*,\bm{z}_*^{[i]}(t))^2}{\sigma_f^2+(\sigma^{[i]}_e)^2}$. Note that the definition of $\bm{z}_*^{[i]}(t)$ renders $\rho_{\bm{z}_*}^{\bm{z}_*^{[i]}(t)}=\rho_{\bm{z}_*}^{\mathcal{Z}^{[i]}(t)}$. Combining this with the decomposition property of $\kappa$ in Assumption \ref{assum:kernel} gives 
\begin{align}
    \check{\sigma}^2_{\bm{z}_*|\mathcal{D}^{[i]}(t)}= \sigma_f^2-\frac{\kappa(\rho_{\bm{z}_*}^{\mathcal{Z}^{[i]}(t)})^2}{\sigma_f^2+(\sigma^{[i]}_e)^2}\label{eq:  agent-based GPR var rho}.
\end{align}

The definition of local dispersion $d^{[i]}(t)$  renders $d^{[i]}(t)\geqslant\rho(\bm{z}_*,\mathcal{Z}^{[i]}(t))$. Combining this with the monotonicity of $\kappa$ in Assumption \ref{assum:kernel} gives $\kappa(d^{[i]}(t))\leqslant\kappa(\rho_{\bm{z}_*}^{\mathcal{Z}^{[i]}(t)})$, which renders $$
\check{\sigma}^2_{\bm{z}_*|\mathcal{D}^{[i]}(t)}\leqslant \sigma_f^2-\frac{\kappa(d^{[i]}(t))^2}{\sigma_f^2+(\sigma^{[i]}_e)^2}, ~\forall \bm{z}_*\in\bm{\mathcal{Z}}.$$
Applying the boundedness of $\kappa$ in Assumption \ref{assum:kernel} to \eqref{eq:  agent-based GPR var rho}, we have
$
    \check{\sigma}^2_{\bm{z}_*|\mathcal{D}^{[i]}(t)}\geqslant\frac{\sigma_f^2(\sigma^{[i]}_e)^2}{\sigma_f^2+(\sigma^{[i]}_e)^2}.
$
$\hfill\blacksquare$

As $\underset{t\to\infty}{\lim}d^{[j]}(t)=0$, $\forall j\in\mathcal{V}$, the upper bound of $\check{\sigma}^2_{\bm{z}_*|\mathcal{D}^{[i]}(t)}$ converges to its lower bound.

\begin{corollary}\label{corollary: lim check sigma}
Suppose Assumption \ref{assum:kernel} holds. If $\underset{t\to\infty}{\lim}d^{[j]}(t)=0$ for all $j\in\mathcal{V}$, it holds that $\underset{t\to\infty}{\lim}\check{\sigma}^2_{\bm{z}_*|\mathcal{D}^{[i]}(t)}=\frac{\sigma_f^2(\sigma^{[i]}_e)^2}{\sigma_f^2+(\sigma^{[i]}_e)^2}$. $\hfill\blacksquare$
\end{corollary}

\subsubsection{Variance analysis of distributed GPR}\label{sec: var distributed GPR}
First, we define the following notations. We let operators $\Delta$, $\sup$, $\max$ and $\min$ be applied element-wise across the vectors:
\begin{align*}
&\bar{\bm{m}}(t)\triangleq\underset{i\in\mathcal{V}}{\max}~\bm{\xi}^{[i]}(t),\quad \ubar{\bm{m}}(t)\triangleq\underset{i\in\mathcal{V}}{\min}~\bm{\xi}^{[i]}(t),\\
&\bm{\delta}_{\bm{m}}(t)\triangleq\bar{\bm{m}}(t)-\ubar{\bm{m}}(t),\quad \bm{r}^{[i]}_{\bm{\xi}}(t)\triangleq[r^{[i]}_{\bm{\xi},\bm{z}_*}(t)]_{\bm{z}_*\in\mathcal{Z}_{agg}},\\
&\Delta \bm{r}_{\max}(t)\triangleq\underset{i\in\mathcal{V}}{\max}~ \Delta \bm{r}_{\bm{\xi}}^{[i]}(t),\quad \Delta \bm{r}_{\min}(t)\triangleq\underset{i\in\mathcal{V}}{\min} ~\Delta \bm{r}_{\bm{\xi}}^{[i]}(t),\\
&\bar{\Delta} \bm{r}_{\max}(t)\triangleq\max_{i\in\mathcal{V}}\{\sup_{s\geqslant 1}\bm{r}_{\bm{\xi}}^{[i]}(s)-\bm{r}_{\bm{\xi}}^{[i]}(t-1)\},\\
&\bm{\delta}_{\bm{r}_{\bm{\xi}}}(t)\triangleq \Delta \bm{r}_{\max}(t)-\Delta \bm{r}_{\min}(t), \quad \zeta\triangleq\alpha^{\frac{1}{2}{n}({n}+1)b-1}.
\end{align*}

First of all, we introduce several properties in Lemma \ref{lmm:distribution GPR properties}.
\begin{lemma}\label{lmm:distribution GPR properties}
Suppose Assumptions \ref{assmp:Balanced Communication} and \ref{assum:kernel} and $\underset{t\to\infty}{\lim}d^{[j]}(t)=0$ for all $j\in\mathcal{V}$ hold.  For each $\bm{z}_*\in\mathcal{Z}_{agg}$:
\begin{claim}\label{claim: r bound}
It holds that $\frac{1}{\sigma_f^2}\leqslant r^{[i]}_{\bm{\xi},\bm{z}_*}(t)\leqslant (\frac{\sigma_f^2(\sigma^{\min}_e)^2}{\sigma_f^2+(\sigma^{\min}_e)^2})^{-1}$, $\forall i\in\mathcal{V}$, $t\geqslant 0$.
\end{claim}

\begin{claim}\label{claim: r monotonicty}
It holds that $r^{[i]}_{\bm{\xi},\bm{z}_*}(t)\geqslant r^{[i]}_{\bm{\xi},\bm{z}_*}(t-1)$, $\forall t\geqslant0$.
\end{claim} 
\begin{claim}\label{claim: xi bound}
It holds that $ \frac{1}{\sigma_f^2}\leqslant\xi^{[i]}_{\bm{z}_*}(t)\leqslant n(\frac{\sigma_f^2(\sigma^{\min}_e)^2}{\sigma_f^2+(\sigma^{\min}_e)^2})^{-1}$, $ \forall i\in\mathcal{V}$, $t\geqslant 1$.
\end{claim}
\begin{claim}\label{claim: delta r xi <= bar delta r max}
It holds that $ \bm{\delta}_{\bm{r}_{\bm{\xi}}}(t)\preceq\bar{\Delta} \bm{r}_{\max}(t)$, $\forall t\geqslant0$.
\end{claim}
\begin{claim}\label{claim: bar delta r max monotonicity}
It holds that $\bar{\Delta} \bm{r}_{\max}(t)\preceq\bar{\Delta} \bm{r}_{\max}(t-1)$,  $\forall t\geqslant 1$.
\end{claim}
\begin{claim}\label{claim: bar delta r max rate}
It holds that $ \|\bar{\Delta} \bm{r}_{\max}(t)\|_\infty\leqslant \mathcal{O}(\sigma_f^4-\kappa(d^{\max}(t-1))^2)$, $\forall t\geqslant 1$.
\end{claim}

\textbf{Proof:} We prove the claims one-by-one:

{\it Proof of Claim \ref{claim: r bound}}:
Recall the boundedness property in Assumption \ref{assum:kernel} requires that $\kappa(\cdot)>0$. Therefore it follows from Part I of Theorem \ref{thm: variance convergence} that 
$$
\frac{\sigma_f^2(\sigma^{[i]}_e)^2}{\sigma_f^2+(\sigma^{[i]}_e)^2}\leqslant\check{\sigma}^2_{\bm{z}_*|\mathcal{D}^{[i]}(t)}\leqslant\sigma_f^2, ~\forall t\geqslant 1.
$$
Combining this with the definition of $r^{[i]}_{\bm{\xi},\bm{z}_*}(t)$ on Line \ref{ln:r xi} in distributed GPR, gives
$$
\frac{1}{\sigma_f^2}\leqslant
    r^{[i]}_{\bm{\xi},\bm{z}_*}(t)\leqslant (\frac{\sigma_f^2(\sigma^{[i]}_e)^2}{\sigma_f^2+(\sigma^{[i]}_e)^2})^{-1}\leqslant(\frac{\sigma_f^2(\sigma^{\min}_e)^2}{\sigma_f^2+(\sigma^{\min}_e)^2})^{-1}.
$$
Combining with initial condition $r^{[i]}_{\bm{\xi},\bm{z}_*}(0)=\frac{1}{\sigma_f^2}$ gives the above inequalities hold for $t\geqslant0$. $\hfill\qedsymbol$

{\it Proof of Claim \ref{claim: r monotonicty}}: Due to incremental sampling, the local collection of input data $\mathcal{Z}^{[i]}(t)=\mathcal{Z}^{[i]}(t-1)\cup\bm{z}^{[i]}(t)$ monotonically expands, hence $\rho_{\bm{z}_*}^{\mathcal{Z}^{[i]}(t)}$ decreases.
By monotonicity of $\kappa$ in Assumption \ref{assum:kernel}, equation \eqref{eq:  agent-based GPR var rho} indicates $\check{\sigma}^2_{\bm{z}_*|\mathcal{D}^{[i]}(t)}$ decreases as $\rho_{\bm{z}_*}^{\mathcal{Z}^{[i]}(t)}$ decreases. This renders that $r^{[i]}_{\bm{\xi},\bm{z}_*}(t)=\check{\sigma}^{-2}_{\bm{z}_*|\mathcal{D}^{[i]}(t)}$ is non-decreasing, i.e., $r^{[i]}_{\bm{\xi},\bm{z}_*}(t)\geqslant r^{[i]}_{\bm{\xi},\bm{z}_*}(t-1)$, for all $t\geqslant 1$.
    With initial conditions  $r^{[i]}_{\bm{\xi},\bm{z}_*}(-1)=0$ and Claim \ref{claim: r bound}, we have $r^{[i]}_{\bm{\xi},\bm{z}_*}(t)\geqslant r^{[i]}_{\bm{\xi},\bm{z}_*}(t-1)$ for $t\geqslant0$.
    $\hfill\qedsymbol$
    
{\it Proof of Claim \ref{claim: xi bound}}:    
Outline: We first show that $$\frac{1}{\sigma_f^2}\leqslant \xi^{[i]}_{\bm{z}_*}(t)\leqslant \sum_{\tau=0}^{t-1}\max_{j\in\mathcal{V}}\Delta r^{[j]}_{\bm{\xi},\bm{z}_*}(\tau)$$ using induction,  then we find an upper bound for $\sum_{\tau=0}^{t-1}\max_{j\in\mathcal{V}}\Delta r^{[j]}_{\bm{\xi},\bm{z}_*}(\tau)$.
 
First, we show the induction. For $t=1$, by Line \ref{ln:dynamic consensus_xi} in distributed GPR and initial condition $\bm{\xi}^{[i]}(0)=\bm{r}^{[i]}_{\bm{\xi}}(-1)=\bm{0}_{|\mathcal{Z}_{agg}|}$, we have 
 \begin{align*}
    \xi^{[i]}_{\bm{z}_*}(1)=r^{[i]}_{\bm{\xi},\bm{z}_*}(0)=\Delta r^{[i]}_{\bm{\xi},\bm{z}_*}(0)\leqslant\max_{j\in\mathcal{V}}\Delta r^{[j]}_{\bm{\xi},\bm{z}_*}(0). 
 \end{align*}
 Since we have initial condition $\bm{r}^{[i]}_{\bm{\xi}}(0)=\frac{1}{\sigma_f^2}\bm{1}_{|\mathcal{Z}_{agg}|}$, the claim holds for $t=1$. Suppose it holds for $t=m$. Then for $t=m+1$, according to distributed GPR Line \ref{ln:dynamic consensus_xi}, we have 
\begin{align}
    \xi^{[i]}_{\bm{z}_*}(m+1)&=\sum_{j=1}^n a_{ij}(m)\xi^{[j]}_{\bm{z}_*}(m)+\Delta r^{[i]}_{\bm{\xi},\bm{z}_*}(m)\label{eq: xi=a_ij xi+delta r}\\
    &\leqslant \sum_{\tau=0}^{m-1}\max_{j\in\mathcal{V}}\Delta r^{[j]}_{\bm{\xi},\bm{z}_*}(\tau)+\max_{j\in\mathcal{V}}\Delta r^{[j]}_{\bm{\xi},\bm{z}_*}(m)\nonumber\\
    &=\sum_{\tau=0}^{m}\max_{j\in\mathcal{V}}\Delta r^{[j]}_{\bm{\xi},\bm{z}_*}(\tau)\nonumber,
\end{align}
where the inequality follows from the row stochasticity in Assumption \ref{assmp:Balanced Communication}. This proves the upper bound of the induction.

By Claim \ref{claim: r monotonicty}, $\Delta r^{[j]}_{\bm{\xi},\bm{z}_*}(m)\geqslant 0$ for all $m\geqslant0$. Since  $r^{[j]}_{\bm{\xi},\bm{z}_*}(0)=\frac{1}{\sigma_f^2}\leqslant \xi^{[i]}_{\bm{z}_*}(m)$, following from \eqref{eq: xi=a_ij xi+delta r} we have 
$$
    \xi^{[i]}_{\bm{z}_*}(m+1)\geqslant\sum_{j=1}^n a_{ij}(t)r^{[j]}_{\bm{\xi},\bm{z}_*}(0)+\Delta r^{[i]}_{\bm{\xi},\bm{z}_*}(m)
   \geqslant\frac{1}{\sigma_f^2}.
$$
The proof for the induction is completed.

Second, we find the upper bound of $\sum_{\tau=0}^{t-1}\max_{j\in\mathcal{V}}\Delta r^{[j]}_{\bm{\xi},\bm{z}_*}(\tau)$. 
Claim \ref{claim: r monotonicty} implies $\Delta r^{[j]}_{\bm{\xi},\bm{z}_*}(\tau)\geqslant0$ for all $\tau\geqslant0$ and all $j\in\mathcal{V}$. Hence
\begin{align}\label{ineq: max delta <= sum delta}
    \sum_{\tau=0}^{t-1}\max_{j\in\mathcal{V}}\Delta r^{[j]}_{\bm{\xi},\bm{z}_*}(\tau)\leqslant \sum_{\tau=0}^{t-1}\sum_{j=1}^n\Delta r^{[j]}_{\bm{\xi},\bm{z}_*}(\tau).
\end{align}
Given initial condition $r^{[j]}_{\bm{\xi},\bm{z}_*}(-1)=0$ and recall the definition of operator $\Delta$ where $\Delta r(t)\triangleq r(t)-r(t{-}1)$, it follows that $\sum_{\tau=0}^{t-1}\Delta r^{[j]}_{\bm{\xi},\bm{z}_*}(\tau)=r^{[j]}_{\bm{\xi},\bm{z}_*}({t-1})$.  Therefore
$$
    \sum_{\tau=0}^{t-1}\sum_{j=1}^n\Delta r^{[j]}_{\bm{\xi},\bm{z}_*}(\tau)= \sum_{j=1}^n \sum_{\tau=0}^{t-1}\Delta r^{[j]}_{\bm{\xi},\bm{z}_*}(\tau)=\sum_{j=1}^nr^{[j]}_{\bm{\xi},\bm{z}_*}({t-1}).
$$
Applying the upper bound in Claim \ref{claim: r bound}, we have
$$
    \sum_{j=1}^nr^{[j]}_{\bm{\xi},\bm{z}_*}({t-1})\leqslant n(\frac{\sigma_f^2(\sigma^{\min}_e)^2}{\sigma_f^2+(\sigma^{\min}_e)^2})^{-1},
$$
which, by \eqref{ineq: max delta <= sum delta}, is also an upper bound for $\sum_{\tau=0}^{t-1}\max_{j\in\mathcal{V}}\Delta r^{[j]}_{\bm{\xi},\bm{z}_*}(\tau)$.$\hfill\qedsymbol$


{\it Proof of Claim \ref{claim: delta r xi <= bar delta r max}}:
By Claim \ref{claim: r monotonicty}, we have
$\Delta \bm{r}_{\min ,\bm{\xi}}(t)\succeq \bm{0}_{|\mathcal{Z}_{agg}|}$, and hence
\begin{align*}
    \bm{\delta}_{\bm{r}_{\bm{\xi}}}(t)&=\Delta \bm{r}_{\max,\bm{\xi}}(t)-\Delta \bm{r}_{\min ,\bm{\xi}}(t)\\
    &\preceq\Delta \bm{r}_{\max, \bm{\xi}}(t)\\
    &= \max_{i\in\mathcal{V}}\{\bm{r}^{[i]}_{\bm{\xi}}(t)-\bm{r}^{[i]}_{\bm{\xi}}(t-1)\}\\
    &\leqslant\max_{i\in\mathcal{V}}\{\sup_{s\geqslant 1}\bm{r}^{[i]}_{\bm{\xi}}(s)-\bm{r}^{[i]}_{\bm{\xi}}(t-1)\}\\
    &=\bar{\Delta} \bm{r}_{\max, \bm{\xi}}(t), \quad \forall t\geqslant 0. &\qquad\qquad\qquad\quad\qedsymbol
\end{align*}

{\it Proof of Claim \ref{claim: bar delta r max monotonicity}}:
 In Claim \ref{claim: r monotonicty} $r^{[i]}_{\bm{\xi},\bm{z}_*}(t)$ being non-decreasing implies
$0\leqslant\bar{\Delta} \bm{r}_{\max}(t)\preceq\bar{\Delta} \bm{r}_{\max}(t-1)$.
$\hfill\qedsymbol$

{\it Proof of Claim \ref{claim: bar delta r max rate}}:
Notice that
\begin{align*}
  &\sup_{s\geqslant 1}r_{\bm{\xi},\bm{z}_*}^{[i]}(s)-r_{\bm{\xi},\bm{z}_*}^{[i]}(t-1)\\
  &\leqslant\sup_{s\geqslant1}r_{\bm{\xi},\bm{z}_*}^{[i]}(s)-\min_{\bm{z}\in\mathcal{Z}_{agg}}r_{\bm{\xi},\bm{z}}^{[i]}(t-1).  
\end{align*}
The definition of $r_{\bm{\xi},\bm{z}_*}(t)$ on Line \ref{ln:r xi} in distributed GPR gives $$\min_{\bm{z}\in\mathcal{Z}_{agg}}r^{[i]}_{\bm{\xi},\bm{z}}(t-1)=(\max_{\bm{z}\in\mathcal{Z}_{agg}} \check{\sigma}^2_{\bm{z}|\mathcal{D}^{[i]}(t-1)})^{-1}.$$
Applying the upper bound in Theorem \ref{thm: variance convergence} Part I renders $$\min_{\bm{z}\in\mathcal{Z}_{agg}}r^{[i]}_{\bm{\xi},\bm{z}}(t-1)\geqslant (\sigma_f^2-\frac{\kappa( d^{[i]}(t-1))^2}{\sigma_f^2+(\sigma^{[i]}_e)^2})^{-1}.
$$
Combining the definition of $r_{\bm{\xi},\bm{z}_*}(t)$ and Theorem \ref{thm: variance convergence} Part I renders $$\sup_{s\geqslant1}r^{[i]}_{\bm{\xi},\bm{z}_*}(s)\leqslant(\frac{\sigma_f^2(\sigma^{[i]}_e)^2}{\sigma_f^2+(\sigma^{[i]}_e)^2})^{-1}, \forall i\in\mathcal{V}.$$ Therefore, we have
\begin{align*}
    &\sup_{s\geqslant1}r_{\bm{\xi},\bm{z}_*}^{[i]}(s)-r_{\bm{\xi},\bm{z}_*}^{[i]}(t-1)\\
    &\leqslant (\frac{\sigma_f^2(\sigma^{[i]}_e)^2}{\sigma_f^2+(\sigma^{[i]}_e)^2})^{-1}
  -(\sigma_f^2
   -\frac{\kappa( d^{[i]}(t-1))^2}{\sigma_f^2+(\sigma^{[i]}_e)^2})^{-1}.
\end{align*}

Let $p^{[i]}\triangleq \frac{\sigma_f^2(\sigma^{[i]}_e)^2}{\sigma_f^2+(\sigma^{[i]}_e)^2}$ and $q^{[i]}(t)\triangleq \sigma_f^2
   -\frac{\kappa( d^{[i]}(t-1))^2}{\sigma_f^2+(\sigma^{[i]}_e)^2}$.
Based on the monotonicity of $\kappa$ in Assumption \ref{assum:kernel} and since $\underset{t\to\infty}{\lim}d^{[i]}(t)=0$ for all $i\in\mathcal{V}$, we have  $q^{[i]}(t) \searrow p^{[i]}$.
Since $q^{[i]}(t)>p^{[i]}>0$ for all $t\geqslant0$, we can apply  manipulation $$\frac{1}{p^{[i]}}-\frac{1}{q^{[i]}(t)}=\frac{q^{[i]}(t)-p^{[i]}}{p^{[i]}q^{[i]}(t)}\leqslant\frac{q^{[i]}(t)-p^{[i]}}{(p^{[i]})^2}.$$  Then we can further obtain
$$ \sup_{s\geqslant1}r_{\bm{\xi},\bm{z}_*}^{[i]}(s)-r_{\bm{\xi},\bm{z}_*}^{[i]}(t-1)\leqslant\beta^{[i]}_r(\sigma_f^4-\kappa(d^{[i]}(t-1))^2),
$$
where $\beta^{[i]}_r\triangleq\frac{\sigma_f^2+(\sigma^{[i]}_e)^2}{\sigma_f^4(\sigma^{[i]}_e)^4}$.

Plugging the above inequality back into the definition of $\bar{\Delta} \bm{r}_{\max}(t)$ and apply the monotonicity of $\kappa$ in Assumption \ref{assum:kernel}, it follows that 
\begin{align*}
    \|\bar{\Delta} \bm{r}_{\max}(t)\|_\infty&\leqslant\max_{i\in\mathcal{V}}\beta^{[i]}_r(\sigma_f^4-\kappa(d^{[i]}(t-1))^2)\\
    &\leqslant\beta^{\max}_r(\sigma_f^4-\kappa(d^{\max}(t-1))^2). \qquad\qquad\quad\qedsymbol
\end{align*}

The proof of the lemma is completed. $\hfill\blacksquare$
\end{lemma}

We define the subsequence $\{t_j\}$ as follows: $t_{-1}\triangleq1$,
\begin{align*}
  t_0\triangleq \textrm{ceil}\Big(\big(\frac{\log(2({n}b-1)\|\Delta \bm{r}_{\max}(1)\|_\infty)}{\log(1-\zeta)\zeta||\bm{\delta}_{\bm{m}}(1)||_\infty}+1\big)\big(nb-1\big)\Big),  
\end{align*}
where $\textrm{ceil}(x)\triangleq\min\{x'\in\mathbb{Z}\mid x'\geqslant x\}$, and for all $j\geqslant 1$,
\begin{align*}
    t_j\triangleq \textrm{ceil}\Big(\big(\frac{\log(\|\Delta \bm{r}_{\max}(t_{j-1})\|_\infty)}{2\log(1-\zeta)\|\Delta \bm{r}_{\max}(t_{j-2})\|_\infty}+1\big)\big({n}b-1\big)\Big).
\end{align*}
The proposition below characterizes the convergence of predictive variance from distributed GPR.
\begin{proposition}\label{prop:rate of convergence of FODAC} 
{\em (Convergence of distributed GPR).}
Suppose Assumptions \ref{assmp:Periodical Strong Connectivity},  \ref{assmp:Balanced Communication}, \ref{assmp:Non-degeneracy} and \ref{assum:kernel} hold.  
For all $\bm{z}_*\in\mathcal{Z}_{agg}$, for all $i\in\mathcal{V}$ and $t\geqslant1$, the convergence of $(\hat{\sigma}^{[i]}_{\bm{z}_*|\mathcal{D}(t)})^2$ in distributed GPR to $(\check{\sigma}^{(agg)}_{\bm{z}_*|\mathcal{D}(t)})^2$ is characterized by:
For $t<t_0$: $$
|(\hat{\sigma}^{[i]}_{\bm{z}_*|\mathcal{D}(t)})^2-(\check{\sigma}^{(agg)}_{\bm{z}_*|\mathcal{D}(t)})^2|
      \leqslant2\sigma_f^4(1-\zeta)^{\frac{t}{{n}b-1}-1}||\bm{\delta}_{\bm{m}}(1)||_\infty;
$$
For $t\geqslant t_0$: $$
|(\hat{\sigma}^{[i]}_{\bm{z}_*|\mathcal{D}(t)})^2-(\check{\sigma}^{(agg)}_{\bm{z}_*|\mathcal{D}(t)})^2|
 \leqslant\frac{4\sigma_f^4}{\zeta}(nb-1)\|\bar{\Delta} \bm{r}_{\max}(t_{l(t)-1})\|_\infty,
$$
where  $l(t)$ is the largest integer such that $t\geqslant\sum_{j=-1}^{l(t)}t_j$.

\textbf{Proof:}
Line \ref{ln:estimation_sigma_M} in distributed GPR indicates that $\xi_{\bm{z}_*}^{[i]}(t)=({\hat{\sigma}}^{[i]}_{\bm{z}_*|\mathcal{D}(t)})^{-2}
$, and combining Line \ref{ln:r xi} in distributed GPR with \eqref{eq:agg_var} gives $\frac{1}{{n}}\sum_{j=1}^n r^{[j]}_{\bm{\xi},\bm{z}_*}(t)=(\check{\sigma}^{(agg)}_{\bm{z}_*|\mathcal{D}(t)})^{-2}$. Hence, we have 
\begin{align}
(\hat{\sigma}^{[i]}_{\bm{z}_*|\mathcal{D}(t)})^2-(\check{\sigma}^{(agg)}_{\bm{z}_*|\mathcal{D}(t)})^2=\frac{\xi_{\bm{z}_*}^{[i]}(t)-\frac{1}{n}\sum_{j=1}^n r^{[j]}_{\bm{\xi},\bm{z}_*}(t)}{\frac{1}{n}\xi_{\bm{z}_*}^{[i]}(t) \sum_{j=1}^n r^{[j]}_{\bm{\xi},\bm{z}_*}(t)}.\label{eq:sigma BCM-sigma hat}
\end{align}

The upper bound of  $\xi_{\bm{z}_*}^{[i]}(t)-\frac{1}{n}\sum_{j=1}^n r^{[j]}_{\bm{\xi},\bm{z}_*}(t)$ is found by  the following two claims, whose proofs are at the end.

\begin{claim}\label{claim: xi-r<D}
It holds that $$\underset{i\in\mathcal{V}}{\max}||\bm{\xi}^{[i]}(t)-\frac{1}{{n}}\sum_{j=1}^n \bm{r}^{[j]}_{\bm{\xi}}(t)||_\infty\leqslant||\bm{\delta}_{\bm{m}}(t)||_\infty. \qquad\qquad\qquad\qedsymbol$$

\end{claim}

\begin{claim}\label{claim: FODAC convergence}

It holds that
\begin{align*}
     ||\bm{\delta}_{\bm{m}}(t)||_\infty
  &\leqslant 2(1-\zeta)^{\frac{t}{{n}b-1}-1}||\bm{\delta}_{\bm{m}}(1)||_\infty, \quad\forall t<t_0;\\
   ||\bm{\delta}_{\bm{m}}(t)||_\infty &\leqslant4({n}b-1)\|\bar{\Delta} \bm{r}_{\max}(t_{l(t)-1})\|_\infty\frac{1}{\zeta},\quad \forall t\geqslant t_0. \quad\hfill\qedsymbol
\end{align*}
 
\end{claim}

Claim \ref{claim: r bound} and Claim \ref{claim: xi bound} in Lemma \ref{lmm:distribution GPR properties} provide that $\xi_{\bm{z}_*}^{[i]}(t)\geqslant\frac{1}{\sigma_f^2}$ and $r^{[i]}_{\bm{\xi},\bm{z}_*}(t)\geqslant\frac{1}{\sigma_f^2}$ respectively. Combining this and Claim \ref{claim: xi-r<D} and \ref{claim: FODAC convergence}   with \eqref{eq:sigma BCM-sigma hat} finishes the proof. $\hfill\blacksquare$

\end{proposition}

{\em Proof of Claim \ref{claim: xi-r<D}:}
Assumption \ref{assmp:Balanced Communication},  the initial condition $\bm{\xi}^{[i]}(0)=\bm{r}^{[i]}_{\bm{\xi}}(0)$, $\forall i\in\mathcal{V}$  and the update rule on Line \ref{ln:dynamic consensus_xi} in distributed GPR render
$$
    \sum_{j=1}^n\bm{\xi}^{[j]}(t)
=\sum_{j=1}^n\bm{\xi}^{[j]}(t-1)+\sum_{j=1}^n\Delta \bm{r}^{[j]}_{\bm{\xi}}(t)   =\sum_{j=1}^n\bm{r}^{[j]}_{\bm{\xi}}(t)
.$$
Hence we have 
\begin{align*}
    \ubar{\bm{m}}(t)=\min_{j\in\mathcal{V}}\bm{\xi}^{[j]}(t)\leqslant\frac{1}{n}\sum_{j=1}^n \bm{r}^{[j]}_{\bm{\xi}}(t)&\leqslant\max_{j\in\mathcal{V}}\bm{\xi}^{[j]}(t)= \bar{\bm{m}}(t)
\\
\underset{j\in\mathcal{V}}{\max}||\bm{\xi}^{[j]}(t)-\frac{1}{{n}}\sum_{j=1}^n \bm{r}^{[j]}_{\bm{\xi}}(t)||_\infty&\leqslant\|\bar{\bm{m}}(t)-\ubar{\bm{m}}(t)\|_\infty&\\
&=||\bm{\delta}_{\bm{m}}(t)||_\infty. \qquad\qquad\qedsymbol\end{align*}

{\em Proof of Claim \ref{claim: FODAC convergence}:}
Outline: 
Write $t=\tau_0+\tau$ for some $\tau_0,\tau>0$. We first derive a general form of $ ||\bm{\delta}_{\bm{m}}(\tau_0+\tau)||_\infty$. Then we prove the cases when $t<t_0$ and $t\geqslant t_0$ respectively.

First, we give the general form. Applying  inequality (B.1) in  \cite{zhu2010discrete} in vector form, we  have
\begin{align}
    &||\bm{\delta}_{\bm{m}}(\tau_0+\tau)||_\infty
    \leqslant \max\{2(1-\zeta)^{\frac{\tau}{nb-1}-1}||\bm{\delta}_{\bm{m}}(\tau_0)||_\infty,\nonumber\\
    &\qquad\qquad\qquad\qquad 2||\bm{\omega}(\tau_0,\tau_0+\tau)||_\infty\},\label{eq: delta m xi t_0}\\
    &\bm{\omega}(\tau_0,\tau)\triangleq(1-\zeta)^{\ell_\tau-1}\sum_{q=\tau_0}^{\tau_1+\tau_0-1}\bm{\delta}_{ \bm{r}_{\bm{\xi}}}(q)+\cdots\nonumber\\
    &\quad+(1-\zeta)\sum_{q=\tau_{\ell_\tau-2}+\tau_0}^{\tau_{\ell_\tau-1}+\tau_0-1}\bm{\delta}_{ \bm{r}_{\bm{\xi}}}(q)+\sum_{q=\tau_{\ell_\tau-1}+\tau_0}^{\tau_{\ell_\tau}+\tau_0-1}\bm{\delta}_{ \bm{r}_{\bm{\xi}}}(q)\nonumber\\
    &\quad+\sum_{q=\tau_{\ell_\tau}+\tau_0}^{\tau-1}\bm{\delta}_{ \bm{r}_{\bm{\xi}}}(q).\nonumber
\end{align}

Second, we prove the case of $t<t_0$. Let $\tau_0=0$, $\tau=t$ in \eqref{eq: delta m xi t_0}. We first find a uniform upper bound for $\bm{\omega}(0,t)$. 
Recall that  Claim \ref{claim: delta r xi <= bar delta r max} shows that $\bm{\delta}_{\bm{r}_{\bm{\xi}}}(t)\preceq\bar{\Delta} \bm{r}_{\max}(t)$ and Claim \ref{claim: bar delta r max monotonicity} shows that $\bar{\Delta} \bm{r}_{\max}(t)$ is element-wise non-increasing.  It follows that $\bm{\delta}_{\bm{r}_{\bm{\xi}}}(t)\preceq \bar{\Delta} \bm{r}_{\max}(1)$ for all $t\geqslant1$ and hence
\begin{align*}    \|\bm{\omega}(0,t)\|_\infty&\leqslant 
     (nb-1)\|\bar{\Delta} \bm{r}_{\max}(1)\|_\infty(1+\sum_{l=0}^\infty (1-\zeta)^l)\\
     &= ({n}b-1)\|\bar{\Delta} \bm{r}_{\max}(1)\|_\infty(\frac{1}{\zeta}+1).
\end{align*}
Since $0<\zeta\leqslant 1$, we can further write that
$||\bm{\omega}(0,t)||_\infty\leqslant  2(nb-1)\|\bar{\Delta} \bm{r}_{\max}(1)\|_\infty\frac{1}{\zeta}.$
Then \eqref{eq: delta m xi t_0} becomes
\begin{align}\label{eq: delta m xi<=}
     ||\bm{\delta}_{\bm{m}}(t)||_\infty
     &\leqslant \max\{2(1-\zeta)^{\frac{t}{nb-1}-1} \|\bm{\delta}_{\bm{m}}(1)\|_\infty,\nonumber\\
     &\quad
     4(nb-1)\|\bar{\Delta} \bm{r}_{\max}(1)\|_\infty\frac{1}{\zeta}\}.
\end{align}

Note that the time-dependent term on the right hand side of \eqref{eq: delta m xi<=} is exponentially decreasing. Suppose $t_0$ is the smallest integer such that   $$
(1-\zeta)^{\frac{t_0}{{n}b-1}-1}\|\bm{\delta}_{\bm{m}}(1)\|_\infty\leqslant 2(nb-1)\|\bar{\Delta} \bm{r}_{\max}(1)\|_\infty\frac{1}{\zeta},
$$ where $t_0$ can be obtained as  defined.
Hence  we have
$$
  \|\bm{\delta}_{\bm{m}}(t)\|_\infty
  \leqslant 2(1-\zeta)^{\frac{t}{{n}b-1}-1}||\bm{\delta}_{\bm{m}}(1)||_\infty, ~\forall t<t_0.
$$
which proves Claim \ref{claim: FODAC convergence} for $t<t_0$, and for $t\geqslant t_0$, we have  
\begin{align}\label{eq: bound  D m_xi}
    ||\bm{\delta}_{\bm{m}}(t)||_\infty\leqslant4({n}b-1)\|\bar{\Delta} \bm{r}_{\max}(1)\|_{\infty}\frac{1}{\zeta}.
\end{align} 

Finally, we prove the case of $t\geqslant t_0$.
Write $\tau_0=t_0$. Then \eqref{eq: delta m xi t_0} becomes
\begin{align}
    ||\bm{\delta}_{\bm{m}}(t_0+\tau)||_\infty
    &\leqslant \max\{2(1-\zeta)^{\frac{\tau}{nb-1}-1}||\bm{\delta}_{\bm{m}}(t_0)||_\infty,\nonumber\\
    &\quad 2||\bm{\omega}(t_0,t_0+\tau)||_\infty\}.
    \label{eq: delta m xi t_0-}
        \end{align}
Applying analogous algebra as of $\bm{\omega}(0,t)$ gives $$
||\bm{\omega}(t_0,t_0+\tau)||_\infty\leqslant  2({n}b-1)\|\bar{\Delta} \bm{r}_{\max}(t_0)\|_\infty\frac{1}{\zeta}.$$ Using this and \eqref{eq: bound  D m_xi} as the upper bound for $||\bm{\delta}_{\bm{m}}(t_0)||_\infty$, we can rewrite \eqref{eq: delta m xi t_0-} as \begin{align*}
    &||\bm{\delta}_{\bm{m}}(t_0+\tau)||_\infty\\
    &\leqslant \max\{8(1-\zeta)^{\frac{\tau}{n\bar{b}-1}-1}({n}b-1)\frac{\|\bar{\Delta} \bm{r}_{\max}(1)\|_\infty}{\zeta},\nonumber\\
    &\quad 4({n}b-1)\|\bar{\Delta} \bm{r}_{\max}(t_0)\|_\infty\frac{1}{\zeta}\}.
\end{align*}

Similarly, let $t_1$ be the smallest integer such that 
$$
\|\bar{\Delta} \bm{r}_{\max}(t_0)\|_\infty\geqslant2(1-\zeta)^{\frac{t_1}{{n}b-1}-1}\bar{\Delta} \bm{r}_{\max}(1). 
$$ 
Using similar manipulation as $t_0$ renders $t_1$ as defined
    and 
    $$
    ||\bm{\delta}_{\bm{m}}(t)||_\infty\leqslant 4({n}b-1)\|\Delta\bar{\bm{r}}_{\max ,\bm{\xi}}(t_0)\|_\infty\frac{1}{\zeta}, ~\forall t\geqslant t_0+t_1.$$

By similar logic, we have  $t_j$ as defined for all $j\geqslant1$ 
 and 
 $||\bm{\delta}_{\bm{m}}(t)||_\infty\leqslant4({n}b-1)\|\bar{\Delta} \bm{r}_{\max}(t_{l(t)-1})\|_\infty\frac{1}{\zeta}.
$
 $\hfill\qedsymbol$

Corollary \ref{corrollary: distributed variance convergence} shows that the predictive variance from distributed GPR converges to that from aggregated method.
\begin{corollary}\label{corrollary: distributed variance convergence}
Suppose the same conditions as in Proposition \ref{prop:rate of convergence of FODAC} hold. If $\underset{t\to\infty}{\lim}d^{[i]}(t)=0$ for all $i\in\mathcal{V}$, then $\underset{t\to\infty}{\lim}|(\hat{\sigma}^{[i]}_{\bm{z}_*|\mathcal{D}(t)})^2-(\check{\sigma}^{(agg)}_{\bm{z}_*|\mathcal{D}(t)})^2|=0$. $\hfill\blacksquare$
\end{corollary}

\subsubsection{Variance analysis of fused GPR}\label{analysis: var fGPR}

First of all, we show that $\underset{t\to\infty}{\lim}\mathcal{Z}^{[i]}_{agg}(t)$ exists.
\begin{lemma}\label{lmm: lim Z_agg exists}
It holds that $\underset{t\to\infty}{\lim}\mathcal{Z}^{[i]}_{agg}(t)$ exists. 

\textbf{Proof:}
By Corollary \ref{corollary: lim check sigma}, $\check{\sigma}^2_{\bm{z}_*|\mathcal{D}^{[i]}(t)}$ converges. Hence in distributed GPR $\Delta r^{[i]}_{\bm{\lambda},\bm{z}_*}(t)\to 0$. By Line \ref{ln: hat sigma ave} in distributed GPR and Corollary 3.1 in \cite{zhu2010discrete}, $\underset{t\to\infty}{\lim}(\hat{\sigma}_{\bm{z}_{agg}|\mathcal{D}(t)}^{ave,[i]})^2$ exists. 
By Corollary \ref{corrollary: distributed variance convergence} and \eqref{eq:agg_var}, the convergence of $\check{\sigma}^2_{\bm{z}_*|\mathcal{D}^{[i]}(t)}$ also implies the convergence of $(\hat{\sigma}^{[i]}_{\bm{z}_*|\mathcal{D}(t)})^2$.  Hence
by definition of $\mathcal{Z}^{[i]}_{agg}(t)$ in Section \ref{sec:fGPR}, $\underset{t\to\infty}{\lim}\mathcal{Z}^{[i]}_{agg}(t)$ exists. 
$\hfill\blacksquare$
\end{lemma}

Lemma \ref{lmm: sufficient for lim s1<0} below presents two properties of agent $i$ where $\underset{t\to\infty}{\lim}\mathcal{Z}^{[i]}_{agg}(t)\neq\emptyset$.
\begin{lemma}\label{lmm: sufficient for lim s1<0}
Suppose the same conditions for Corollary \ref{corrollary: distributed variance convergence} hold and $d^{[j]}(t)\to0$,  $\forall j\in\mathcal{V}$. If $\underset{t\to\infty}{\lim}\mathcal{Z}^{[i]}_{agg}(t)\neq\emptyset$ for some $i\in\mathcal{V}$, then $\psi^{[i]}< \frac{1}{n}\sum_{j=1}^n\psi^{[j]}\leqslant c \textrm{ and } \mu^{-1}_{\chi}\chi^{[i]}<1 $. 

\textbf{Proof: }
Outline:  We first show that $\psi^{[i]}< \frac{1}{n}\sum_{j=1}^n\psi^{[j]}\leqslant c$. Then we show $\mu^{-1}_{\chi}\chi^{[i]}<1 $.

First, we  show that $\psi^{[i]}<\frac{1}{n}\sum_{j=1}^n\psi^{[j]}\leqslant c$.
Since $\underset{{t\to\infty}}{\lim}\mathcal{Z}^{[i]}_{agg}(t)\neq\emptyset$,  we pick $\bm{z}_*\in\underset{{t\to\infty}}{\lim}\mathcal{Z}^{[i]}_{agg}(t)$.
Note that $\lambda_{\bm{z}_*}^{[i]}(t)$ is tracking the signal $r^{[i]}_{\bm{\lambda},\bm{z}_*}(t)=\check{\sigma}^2_{\bm{z}_*|\mathcal{D}^{[i]}(t)}$ using FODAC algorithm in \cite{zhu2010discrete}.
By Corollary \ref{corollary: lim check sigma} and Corollary 3.1 in \cite{zhu2010discrete}, $\underset{{t\to\infty}}{\lim}\lambda_{\bm{z}_*}^{[i]}(t)=\underset{{t\to\infty}}{\lim}\frac{1}{n}\sum_{j=1}^n\check{\sigma}^2_{\bm{z}_*|\mathcal{D}^{[j]}(t)}$.  By Line \ref{ln: hat sigma ave} in distributed GPR, we have $\underset{{t\to\infty}}{\lim}(\hat{\sigma}_{\bm{z}_{*}|\mathcal{D}(t)}^{ave,[i]})^2=\underset{{t\to\infty}}{\lim}\lambda_{\bm{z}_*}^{[i]}(t)$. Since $\bm{z}_*\in\underset{{t\to\infty}}{\lim}\mathcal{Z}^{[i]}_{agg}(t)$, by Corollay \ref{corollary: lim check sigma} and the definition of $\mathcal{Z}^{[i]}_{agg}(t)$ on Line \ref{ln:Z_agg(t)} in fused GPR,  we have \begin{align*}
    \sigma_f^2-\sigma_f^2\psi^{[i]}&=\sigma_f^2-\frac{\sigma_f^4}{\sigma^2_f+(\sigma^{[i]}_e)^2}=\frac{\sigma_f^2(\sigma^{[i]}_e)^2}{\sigma_f^2+(\sigma^{[i]}_e)^2}\\
    &=\underset{{t\to\infty}}{\lim}\check{\sigma}^2_{\bm{z}_*|\mathcal{D}^{[i]}(t)}\\
    &>\underset{{t\to\infty}}{\lim}(\hat{\sigma}_{\bm{z}_{*}|\mathcal{D}(t)}^{ave,[i]})^2=\lim_{t\to\infty}\frac{1}{n}\sum_{j=1}^n\check{\sigma}^2_{\bm{z}_*|\mathcal{D}^{[j]}(t)}\\
    &=\frac{1}{n}\sum_{j=1}^n(\sigma_f^2-\frac{\sigma_f^4}{\sigma^2_f+(\sigma^{[j]}_e)^2})=\sigma_f^2-\frac{ \sigma_f^2}{n}\sum_{j=1}^n\psi^{[j]},
\end{align*}
which renders $ \psi^{[i]}<\frac{1}{n}\sum_{j=1}^n\psi^{[j]}. 
$

Since both $\psi^{[i]}$ and $\chi^{[i]}$ are monotonically decreasing as $\sigma_e^{[i]}$ increases, $\chi^{[i]}\leqslant\chi^{[j]}$ if and only if $\psi^{[i]}\leqslant\psi^{[j]}$. Without loss of generality, assume that $\chi^{[1]}\leqslant\cdots\leqslant\chi^{[n]}$ and $\psi^{[1]}\leqslant\cdots\leqslant\psi^{[n]}$. Then Chebyshev's sum inequality \cite{besenyei2018picard} gives
\begin{align}\label{ineq: ave psi< c1}
   \frac{1}{n}\sum_{j=1}^n\psi^{[j]}\leqslant(\frac{1}{n}\sum_{j=1}^n\chi^{[j]})^{-1}\big(\frac{1}{n}\sum_{j=1}^n\chi^{[j]}\psi^{[j]}\big)=c.
\end{align}
Hence $\psi^{[i]}< \frac{1}{n}\sum_{j=1}^n\psi^{[j]}\leqslant c$.

Second, we show that $\mu^{-1}_{\chi}\chi^{[i]}<1 $. By definition and Corollary \ref{corollary: lim check sigma}, $\chi^{[i]}=\underset{{t\to\infty}}{\lim}\check{\sigma}^{-2}_{\bm{z}_*|\mathcal{D}^{[i]}(t)}$. By \eqref{eq:agg_var}, $\mu^{-1}_{\chi}=\underset{{t\to\infty}}{\lim}(\check{\sigma}^{(agg)}_{\bm{z}_*|\mathcal{D}(t)})^2$ for all $\bm{z}_*\in\bm{\mathcal{Z}}$. Since $\bm{z}_*\in\underset{{t\to\infty}}{\lim}\mathcal{Z}^{[j]}_{agg}(t)$,  $\underset{{t\to\infty}}{\lim}\check{\sigma}^2_{\bm{z}_*|\mathcal{D}^{[i]}(t)}>\underset{{t\to\infty}}{\lim}(\hat{\sigma}^{[i]}_{\bm{z}_*|\mathcal{D}(t)})^2 =\underset{{t\to\infty}}{\lim}(\check{\sigma}^{(agg)}_{\bm{z}_*|\mathcal{D}(t)})^2$, where the equality follows from Corollary \ref{corrollary: distributed variance convergence}.
Hence 
$$
 \mu^{-1}_{\chi}\chi^{[i]}=\underset{{t\to\infty}}{\lim}(\check{\sigma}^{(agg)}_{\bm{z}_*|\mathcal{D}(t)})^2\check{\sigma}^{-2}_{\bm{z}_*|\mathcal{D}^{[i]}(t)}<1.\qquad\qquad\qquad\blacksquare
$$
\end{lemma}

Now we present of proof of Theorem \ref{thm: variance convergence} Part II.

\textbf{Proof of Theorem \ref{thm: variance convergence} Part II:} 
By Line \ref{ln: return local} in fused GPR, it is obvious that if $\mathcal{Z}^{[i]}_{agg}(t)=\emptyset$, $\gamma^{[i]}_{\sigma,\bm{z}_*}(t)=0$.
Now we consider the case when $\mathcal{Z}^{[i]}_{agg}(t)\neq\emptyset$.

Outline:
The proof is composed of three parts:
expression of $\gamma^{[i]}_{\sigma,\bm{z}_*}(t)$ and its uniform lower bound;  verfication of the selection of $g(\bm{z}_*,t)$;  derivation of the growth factor of $\gamma^{[i]}_{\sigma,\bm{z}_*}(t)$.

First, we show the expression of $\gamma^{[i]}_{\sigma,\bm{z}_*}(t)$ and derive its uniform lower bound.  According to Line \ref{ln:fusion Sigma} in fused GPR, we have 
$
  (\tilde{\sigma}^{[i]}_{\bm{z}_*|\mathcal{D}(t)})^2= \check{\sigma}^2_{\bm{z}_*|\mathcal{D}^{[i]}(t)}-\gamma^{[i]}_{\sigma,\bm{z}_*}(t),
$
where 
\begin{align}
    \gamma^{[i]}_{\sigma,\bm{z}_*}(t)&\triangleq\gamma^{[i]}_{\sigma,\bm{z}_*,1}(t)\gamma^{[i]}_{\sigma,\bm{z}_*,2}(t)\gamma^{[i]}_{\sigma,\bm{z}_*,3}(t)\big(\check{\sigma}^{[i]}_{\min,\bm{z}_*}(t)\big)^2,\nonumber\\
    \big(\check{\sigma}^{[i]}_{\min,\bm{z}_*}(t)\big)^2&\triangleq\min\{ \check{\sigma}^{2}_{\bm{z}^{[i]}_{agg*}(t)|\mathcal{D}^{[i]}(t)},\check{\sigma}^{2}_{\bm{z}_{*}|\mathcal{D}^{[i]}(t)}\},\nonumber\\
    \gamma^{[i]}_{\sigma,\bm{z}_*,1}(t)&\triangleq\frac{k\big(\bm{z}_*,\bm{z}^{[i]}_{agg*}(t)\big)^2\cdot\max\{0,(c-\psi^{[i]})^2\}}{k\big(\bm{z}_*,\bm{z}_*\big)^2},\nonumber\\
    \gamma^{[i]}_{\sigma,\bm{z}_*,2}(t)&\triangleq \frac{\check{\sigma}^2_{\bm{z}^{[i]}_{agg*}(t)|\mathcal{D}^{[i]}(t)}- (\hat{\sigma}^{[i]}_{\bm{z}^{[i]}_{agg *}(t)|\mathcal{D}(t)})^2}{\check{\sigma}^2_{\bm{z}^{[i]}_{agg*}(t)|\mathcal{D}^{[i]}(t)}},\nonumber\\
    \gamma^{[i]}_{\sigma,\bm{z}_*,3}(t)&\triangleq  \frac{\big(\check{\sigma}^{[i]}_{\min,\bm{z}_*}(t)\big)^2}{\check{\sigma}^2_{\bm{z}^{[i]}_{agg*}(t)|\mathcal{D}^{[i]}(t)}}.\label{def: gamma_sigma}
\end{align}
Line \ref{ln:Z_agg(t)} of fused GPR rules that $(\hat{\sigma}^{[i]}_{\bm{z}^{[i]}_{agg *}(t)|\mathcal{D}(t)})^2<\check{\sigma}^2_{\bm{z}^{[i]}_{agg*}(t)|\mathcal{D}^{[i]}(t)}$.
Obviously, $\gamma^{[i]}_{\sigma,\bm{z}_*}(t)\geqslant 0$.

Second, we verify the selection of $g(\bm{z}_*,t)$. 
We verify that the selection of $g(\bm{z}_*,t)$ is valid by showing $(\tilde{\sigma}^{[i]}_{\bm{z}_*|\mathcal{D}(t)})^2> 0$.
We analyze each factor of $\gamma^{[i]}_{\sigma,\bm{z}_*}(t)$ as follows.

Note that $c,\psi^{[i]}\in(0,1)$, hence $0\leqslant\max\{0,(c-\psi^{[i]})^2\}\leqslant1$. The decomposition and monotonicity properties in Assumption \ref{assum:kernel} gives $0\leqslant \frac{k\big(\bm{z}_*,\bm{z}^{[i]}_{agg*}(t)\big)^2\}}{k\big(\bm{z}_*,\bm{z}_*\big)^2}\leqslant1$.   Combining these gives $0\leqslant\gamma^{[i]}_{\sigma,\bm{z}_*,1}\leqslant 1$.

 Line \ref{ln:Z_agg(t)} of fused GPR rules that $(\hat{\sigma}^{[i]}_{\bm{z}^{[i]}_{agg *}(t)|\mathcal{D}(t)})^2<\check{\sigma}^2_{\bm{z}^{[i]}_{agg*}(t)|\mathcal{D}^{[i]}(t)}$. By Claim \ref{claim: xi bound} and Line \ref{ln:estimation_sigma_M} in distributed GPR, $(\hat{\sigma}^{[i]}_{\bm{z}^{[i]}_{agg *}(t)|\mathcal{D}(t)})^2>0$. Therefore $0<\gamma^{[i]}_{\sigma,\bm{z}_*,2}<1$.

By definition, $\big(\check{\sigma}^{[i]}_{\min,\bm{z}_*}(t)\big)^2\leqslant \check{\sigma}^2_{\bm{z}^{[i]}_{agg*}(t)|\mathcal{D}^{[i]}(t)}$, which renders $0<\gamma^{[i]}_{\sigma,\bm{z}_*,3}\leqslant 1$.

The above upper bounds give $ \gamma^{[i]}_{\sigma,\bm{z}_*}(t)< \big(\check{\sigma}^{[i]}_{\min,\bm{z}_*}(t)\big)^2\leqslant\check{\sigma}^2_{\bm{z}_*|\mathcal{D}^{[i]}(t)}$ and $(\tilde{\sigma}^{[i]}_{\bm{z}_*|\mathcal{D}(t)})^2>\check{\sigma}^2_{\bm{z}_*|\mathcal{D}^{[i]}(t)}-\check{\sigma}^2_{\bm{z}_{*}|\mathcal{D}^{[i]}(t)}=0$.

Finally, we derive the growth factor of $\gamma^{[i]}_{\sigma,\bm{z}_*}(t)$. 
According to the definition of $\gamma^{[i]}_{\sigma,\bm{z}_*}(t)$ in \eqref{def: gamma_sigma}, we can derive the growth factor  of $\gamma^{[i]}_{\sigma,\bm{z}_*}(t)$ by analyzing the growth factor of $\gamma^{[i]}_{\sigma,\bm{z}_*,1}(t)$, $\gamma^{[i]}_{\sigma,\bm{z}_*,2}(t)$, $\gamma^{[i]}_{\sigma,\bm{z}_*,3}(t)$ and $\big(\check{\sigma}^{[i]}_{\min,\bm{z}_*}(t)\big)^2$ respectively.

We first consider $\gamma^{[i]}_{\sigma,\bm{z}_*,2}(t)$.
Let $$\hat{c}\triangleq\max\{\frac{4\sigma_f^4}{\zeta}(nb-1),2\sigma_f^4||\bm{\delta}_{\bm{m}}(1)||_\infty\}.$$ The upper bound given in Proposition \ref{prop:rate of convergence of FODAC} can be written as 
\begin{equation}\label{def: epsilon}
h(t)\triangleq
\left\{ \begin{aligned}
&\hat{c}(1-\zeta)^{\frac{t}{{n}b-1}-1},~ t< t_0,\\
 &\hat{c}\|\bar{\Delta} \bm{r}_{\max}(t_{l(t)-1})\|_\infty, ~ t\geqslant t_0.
\end{aligned}\right.
\end{equation}
Then Proposition \ref{prop:rate of convergence of FODAC} gives 
$$ 
(\hat{\sigma}^{[i]}_{\bm{z}^{[i]}_{agg *}(t)|\mathcal{D}(t)})^2\leqslant (\check{\sigma}^{(agg)}_{\bm{z}^{[i]}_{agg*}(t)|\mathcal{D}(t)})^2+h(t).
$$ 
Hence we have lower bound
$$
    \gamma^{[i]}_{\sigma,\bm{z}_*,2}(t)\geqslant\frac{\check{\sigma}^2_{\bm{z}^{[i]}_{agg*}(t)|\mathcal{D}^{[i]}(t)}-(\check{\sigma}^{(agg)}_{\bm{z}^{[i]}_{agg*}(t)|\mathcal{D}(t)})^2-h(t)}{\check{\sigma}^4_{\bm{z}^{[i]}_{agg*}(t)|\mathcal{D}^{[i]}(t)}}.
$$


The upper bound and lower bound of $(\check{\sigma}^{(agg)}_{\bm{z}_*|\mathcal{D}(t)})^2$ is given below, whose proof can be found at the end of the proof.
\begin{claim}\label{claim: check sigma agg}
For each $\bm{z}_*\in\mathcal{Z}_{*}$, the aggregated variance returned from \eqref{eq:agg_var} can be characterized as
$$
\frac{\sigma_f^2(\sigma^{\min}_e)^2}{\sigma_f^2+(\sigma^{\min}_e)^2}\leqslant
(\check{\sigma}^{(agg)}_{\bm{z}_*|\mathcal{D}(t)})^2\leqslant \sigma_f^2-\frac{\frac{1}{n}\sum_{i=1}^n\kappa\big(\rho_{\bm{z}_*}^{\mathcal{Z}^{[i]}(t)}\big)^2}{\sigma_f^2+(\sigma^{\max}_e)^2}. \quad\hfill\qedsymbol
$$ 
\end{claim}

Denote $\phi^{[i]}\triangleq\sigma_f^2+(\sigma^{[i]}_e)^2$. 
Plugging in equality \eqref{eq:  agent-based GPR var rho} for $\check{\sigma}^2_{\bm{z}^{[i]}_{agg*}(t)|\mathcal{D}^{[i]}(t)}$ and the upper bound in Claim \ref{claim: check sigma agg} for $(\check{\sigma}^{(agg)}_{\bm{z}^{[i]}_{agg*}(t)|\mathcal{D}(t)})^2$ in the inequality above gives
\begin{align*}
\gamma^{[i]}_{\sigma,\bm{z}_*,2}(t)\geqslant&\frac{\phi^{[i]}}{\phi^{\max}}\Big(\frac{\frac{\phi^{[i]}}{n}\sum_{j=1}^n\kappa\big(\rho_{\bm{z}^{[i]}_{agg*}(t)}^{\mathcal{Z}^{[j]}(t)}\big)^2}{\big(\sigma_f^2\phi^{[i]}-\kappa(\rho_{\bm{z}^{[i]}_{agg*}(t)}^{\mathcal{Z}^{[i]}(t)})^2\big)^2}\\
&\frac{-\phi^{\max}\kappa\big(\rho_{\bm{z}^{[i]}_{agg*}(t)}^{\mathcal{Z}^{[i]}(t)}\big)^2
    -\phi^{\max}\phi^{[i]}h(t)}{\big(\sigma_f^2\phi^{[i]}-\kappa(\rho_{\bm{z}^{[i]}_{agg*}(t)}^{\mathcal{Z}^{[i]}(t)})^2\big)^2}\Big).
\end{align*}
The boundedness of $\kappa$ in Assumption \ref{assum:kernel} gives $|\sigma_f^2\phi^{[i]}-\kappa(\rho_{\bm{z}^{[i]}_{agg*}(t)}^{\mathcal{Z}^{[i]}(t)})^2|\leqslant \sigma_f^2\phi^{[i]}$. Applying this upper bound to the denominator of the lower bound above gives
\begin{align*}
    \gamma^{[i]}_{\sigma,\bm{z}_*,2}(t)\geqslant&\frac{1}{\phi^{[i]}\phi^{\max}\sigma_f^4} \Big(\frac{\phi^{[i]}}{n}\sum_{j=1}^n\kappa(\rho_{\bm{z}^{[i]}_{agg*}(t)}^{\mathcal{Z}^{[j]}(t)})^2\\
    &-\phi^{\max}\kappa(\rho_{\bm{z}^{[i]}_{agg*}(t)}^{\mathcal{Z}^{[i]}(t)})^2
    -\phi^{\max}\phi^{[i]}h(t)\Big).
\end{align*}

Now we characterize the rest of factors of $\gamma^{[i]}_{\sigma,\bm{z}_*}(t)$.
Theorem \ref{thm: variance convergence} 
By monotonicity and decomposition in Assumption \ref{assum:kernel}, we have $k\big(\bm{z}_*,\bm{z}_*\big)^2=\sigma_f^4$ and $k(\bm{z}_*,\bm{z}^{[i]}_{agg*}(t))^2=\kappa(\rho_{\bm{z}_*}^{\bm{z}^{[i]}_{agg*}(t)})^2$. By Lemma \ref{lmm: sufficient for lim s1<0}, we have $$
c-\psi^{[i]}\geqslant\frac{1}{n}\sum_{j=1}^n\psi^{[j]}-\psi^{[i]}=\sigma_f^2\beta^{[i]}_\psi>0,
$$ where $\beta^{[i]}_\psi\triangleq(\frac{1}{n}\sum_{j=1}^n(\phi^{[j]})^{-1}-(\phi^{[i]})^{-1})$. This gives $\gamma^{[i]}_{\sigma,\bm{z}_*,1}(t)\geqslant\kappa(\rho_{\bm{z}_*}^{\bm{z}^{[i]}_{agg*}(t)})^2\beta^{[i]}_\psi/\sigma_f^2$.

Part I indicates that $\check{\sigma}^2_{\bm{z}^{[i]}_{agg*}(t)|\mathcal{D}^{[i]}(t)}\geqslant \frac{\sigma_f^2(\sigma^{[i]}_e)^2}{\sigma_f^2+(\sigma^{[i]}_e)^2}$. Therefore we have $\big(\check{\sigma}^{[i]}_{\min,\bm{z}_*}(t)\big)^4\geqslant \frac{\sigma_f^4(\sigma^{[i]}_e)^4}{(\phi^{[i]})^2}$. 

Equality \eqref{eq:  agent-based GPR var rho} indicates $\check{\sigma}^2_{\bm{z}^{[i]}_{agg*}(t)|\mathcal{D}^{[i]}(t)}\leqslant \sigma_f^2$. Combining this with the lower bound of $\big(\check{\sigma}^{[i]}_{\min,\bm{z}_*}(t)\big)^4$ above  gives $\gamma^{[i]}_{\sigma,\bm{z}_*,3}(t)\geqslant\frac{\sigma_f^2(\sigma^{[i]}_e)^4}{(\phi^{[i]})^2}$.

Combining the lower bounds of all the factors  gives
\begin{align*}
     \gamma^{[i]}_{\sigma,\bm{z}_*}(t)&\geqslant \kappa(\rho_{\bm{z}_*}^{\bm{z}^{[i]}_{agg*}(t)})^2\Big(\frac{(\sigma^{[i]}_e)^8}{(\phi^{[i]})^3\phi^{\max}} \big(\frac{\phi^{[i]}}{n}\sum_{j=1}^n\kappa(\rho_{\bm{z}^{[i]}_{agg*}(t)}^{\mathcal{Z}^{[j]}(t)})^2\\
    &\quad-\phi^{\max}\kappa(\rho_{\bm{z}^{[i]}_{agg*}(t)}^{\mathcal{Z}^{[i]}(t)})^2
    -\phi^{\max}\phi^{[i]}h(t)\big)\Big)\beta^{[i]}_\psi.
\end{align*}
The definition in \eqref{def: epsilon} and Claim \ref{claim: bar delta r max rate} renders $h(t)\to 0$.
This renders the Big O notion. $\hfill\blacksquare$

{\em Proof of Claim \ref{claim: check sigma agg}: }
Using equality \eqref{eq:  agent-based GPR var rho}, we can characterize $(\check{\sigma}^{(agg)}_{\bm{z}_*|\mathcal{D}(t)})^2$ in \eqref{eq:agg_var} as
$$
   (\check{\sigma}^{(agg)}_{\bm{z}_*|\mathcal{D}(t)})^{-2}= \frac{1}{n}\sum_{i=1}^n(\sigma_f^2-\frac{\kappa\big(\rho_{\bm{z}_*}^{\mathcal{Z}^{[i]}(t)}\big)^2}{\sigma_f^2+(\sigma^{[i]}_e)^2})^{-1}.
$$
Taking the inverse and applying Lemma \ref{lmm: f} by substituting $x_i$ with $(\sigma_f^2-\frac{\kappa\big(\rho_{\bm{z}_*}^{\mathcal{Z}^{[i]}(t)}\big)^2}{\sigma_f^2+(\sigma^{[i]}_e)^2})^{-1}$ to $f_1$ gives
\begin{align*}
    (\check{\sigma}^{(agg)}_{\bm{z}_*|\mathcal{D}(t)})^2&\leqslant \sigma_f^2-\frac{1}{n}\sum_{i=1}^n\frac{\kappa\big(\rho_{\bm{z}_*}^{\mathcal{Z}^{[i]}(t)}\big)^2}{\sigma_f^2+(\sigma^{[i]}_e)^2}\\
    &\leqslant\sigma_f^2-\frac{\frac{1}{n}\sum_{i=1}^n\kappa\big(\rho_{\bm{z}_*}^{\mathcal{Z}^{[i]}(t)}\big)^2}{\sigma_f^2+(\sigma^{\max}_e)^2}.
\end{align*}

The lower bound provided in Part I of Theorem \ref{thm: variance convergence} and equation \eqref{eq:  agent-based GPR var rho} give
$$
   (\check{\sigma}^{(agg)}_{\bm{z}_*|\mathcal{D}(t)})^{-2}
    \leqslant\frac{1}{n}\sum_{j=1}^n(\frac{\sigma_f^2(\sigma^{[j]}_e)^2}{\sigma_f^2+(\sigma^{[j]}_e)^2})^{-1}\leqslant(\frac{\sigma_f^2(\sigma^{\min}_e)^2}{\sigma_f^2+(\sigma^{\min}_e)^2})^{-1},
$$
where the last inequality follows from the fact that $\frac{\sigma_f^2\sigma^2_e}{\sigma_f^2+\sigma^2_e}$ monotonically increases with respect to $\sigma^2_e$, i.e., $\frac{d}{d\sigma^2_e}\frac{\sigma_f^2\sigma^2_e}{\sigma_f^2+\sigma^2_e}=\frac{\sigma_f^4}{(\sigma_f^2+\sigma_e^2)^2}>0$.
Taking the inverse gives the lower bound.
$\hfill\qedsymbol$

\subsection{Proof of Theorem \ref{thm: mean convergence}}\label{sec:analysis mean}
In this section, we present the theoretical results that leads to Theorem \ref{thm: mean convergence}. We first present the error between the predictive mean of  agent-based GPR and the ground truth, which is the result  of Theorem \ref{thm: mean convergence}  Part I, in Section \ref{sec: mean agent-based GPR}.
Secondly, we characterize the predictive mean returned from distributed GPR in Proposition \ref{prop: lim hat mu} in Section \ref{sec: mean distributed GPR}. 
Lastly, we finish the proof for Part II of Theorem \ref{thm: mean convergence} in Section \ref{sec:mean analysis: fGPR}.

\subsubsection{Mean analysis of  agent-based GPR}\label{sec: mean agent-based GPR}

In this section, we provide the proof of Part I of Theorem \ref{thm: mean convergence}.

{\bf Proof of Part I of Theorem \ref{thm: mean convergence}:}
By Assumption \ref{assmp: covariance} and decomposition and monotonicity properties in Assumption \ref{assum:kernel}, Line \ref{ln: agent-based GPR mu} of  agent-based GPR becomes
$
    \check{\mu}_{\bm{z}_*|\mathcal{D}^{[i]}(t)}=\frac{\kappa(\rho_{\bm{z}_*}^{\mathcal{Z}^{[i]}(t)})}{\sigma_f^2+(\sigma^{[i]}_e)^2} \cdot y^{[i]}_{\bm{z}^{[i]}_*(t)}.\label{eq: check mu}
$
It implies that 
\begin{align*}
  \check{\mu}_{\bm{z}_*|\mathcal{D}^{[i]}(t)}-\eta(\bm{z}_*) =&(1-\frac{\kappa(\rho_{\bm{z}_*}^{\mathcal{Z}^{[i]}(t)})}{\sigma_f^2+(\sigma^{[i]}_e)^2})(-\eta(\bm{z}_*))\\
  &    
     +\frac{\kappa(\rho_{\bm{z}_*}^{\mathcal{Z}^{[i]}(t)})}{\sigma_f^2+(\sigma^{[i]}_e)^2}\Big(y^{[i]}_{\bm{z}^{[i]}_*(t)}-\eta(\bm{z}^{[i]}_*(t))\Big)\\
     &+\frac{\kappa(\rho_{\bm{z}_*}^{\mathcal{Z}^{[i]}(t)})}{\sigma_f^2+(\sigma^{[i]}_e)^2}\Big(\eta(\bm{z}^{[i]}_*(t))-\eta(\bm{z}_*)\Big).
\end{align*}
By boundedness of $\kappa$ in Assumption \ref{assum:kernel}, $0<\frac{\kappa(\rho_{\bm{z}_*}^{\mathcal{Z}^{[i]}(t)})}{\sigma_f^2+(\sigma^{[i]}_e)^2}<1$. 
Combining this with triangular inequality gives
\begin{align}\label{eq: local, mu-eta}
  &|\check{\mu}_{\bm{z}_*|\mathcal{D}^{[i]}(t)}-\eta(\bm{z}_*)| <(1-\frac{\kappa(\rho_{\bm{z}_*}^{\mathcal{Z}^{[i]}(t)})}{\sigma_f^2+(\sigma^{[i]}_e)^2})|\eta(\bm{z}_*)|\nonumber\\
    &+|y^{[i]}_{\bm{z}^{[i]}_*(t)}-\eta(\bm{z}^{[i]}_*(t))|+|\eta(\bm{z}^{[i]}_*(t))-\eta(\bm{z}_*)|.
\end{align}

Now we analyze the upper bound of each term on the right hand side of \eqref{eq: local, mu-eta}.
Recall that $\bm{z}^{[i]}_*(t)\in\proj(\bm{z}_*,\mathcal{Z}^{[i]}(t))$. Utilizing the Lipschitz continuity of $\eta$ in Assumption \ref{assmp:lipschitz} gives
$$
    |\eta(\bm{z}^{[i]}_*(t))-\eta(\bm{z}_*)|\leqslant \ell_\eta\rho({\bm{z}_*},\bm{z}^{[i]}_*(t))=\ell_\eta\rho_{\bm{z}_*}^{\mathcal{Z}^{[i]}(t)}.
$$

The observation model \eqref{eq: observation model} gives $y^{[i]}_{\bm{z}^{[i]}_*(t)}\sim\mathcal{N}(\eta(\bm{z}^{[i]}_*(t)),(\sigma^{[i]}_e)^2)$. Therefore by Chebyshev inequality 
(page 151, \cite{papoulis2002probability}), for all $\epsilon>0$, we have $$P\{|y^{[i]}_{\bm{z}^{[i]}_*(t)}-\eta(\bm{z}^{[i]}_*(t))|\geqslant\epsilon\}\leqslant\frac{(\sigma^{[i]}_e)^2}{\epsilon^2}.$$ 
Note that $|\eta(\bm{z}_*)|\leqslant\|\eta\|_{\bm{\mathcal{Z}}}$. Applying these two inequalities to \eqref{eq: local, mu-eta} gives
$$
     |\check{\mu}_{\bm{z}_*|\mathcal{D}^{[i]}(t)}-\eta(\bm{z}_*)|
    \leqslant(1-\frac{\kappa(\rho_{\bm{z}_*}^{\mathcal{Z}^{[i]}(t)})}{\sigma_f^2+(\sigma^{[i]}_e)^2})\|\eta\|_{\bm{\mathcal{Z}}}+\epsilon+\ell_\eta\rho_{\bm{z}_*}^{\mathcal{Z}^{[i]}(t)}
$$
with probability at least $1-\frac{(\sigma^{[i]}_e)^2}{\epsilon^2}\geqslant1-\frac{(\sigma^{\max}_e)^2}{\epsilon^2}$.
The proof is completed by using inequality $d^{[i]}(t)\geqslant\rho_{\bm{z}_*}^{\mathcal{Z}^{[i]}(t)}$ and the monotonicity property of $\kappa$ in Assumption \ref{assum:kernel}. $\hfill\blacksquare$

\begin{remark}\label{remark: lim check mu}
We can write
$
    \check{\mu}_{\bm{z}_*|\mathcal{D}^{[i]}(t)}=\check{r}^{[i]}_{\bm{z}_*}(t)+\check{e}^{[i]}_{\bm{z}_*}(t)$,
where $\check{r}^{[i]}_{\bm{z}_*}(t)\triangleq\frac{\kappa(\rho_{\bm{z}_*}^{\mathcal{Z}^{[i]}(t)})\eta(\bm{z}^{[i]}_*(t))}{\sigma_f^2+(\sigma^{[i]}_e)^2}$ depends on latent function $\eta$ and $\check{e}^{[i]}_{\bm{z}_*}(t)\triangleq \frac{\kappa(\rho_{\bm{z}_*}^{\mathcal{Z}^{[i]}(t)})}{\sigma_f^2+(\sigma^{[i]}_e)^2} \big(y^{[i]}_{\bm{z}^{[i]}_*(t)}-\eta(\bm{z}^{[i]}_*(t))\big)$ depends on measurement noise. 
Denote $\check{r}^{[i]}_{\bm{z}_*}\triangleq\underset{t\to\infty}{\lim} \check{r}^{[i]}_{\bm{z}_*}(t) =\frac{\sigma_f^2\eta(\bm{z}_*)}{\sigma_f^2+(\sigma^{[i]}_e)^2}$ and   $\check{e}^{[i]}_{\bm{z}_*}\triangleq \underset{t\to\infty}{\lim}\check{e}^{[i]}_{\bm{z}_*}(t)=\frac{\sigma_f^2}{\sigma_f^2+(\sigma^{[i]}_e)^2}e_{\bm{z}_*}^{[i]}$, where $e_{\bm{z}_*}^{[i]}\triangleq y^{[i]}_{\bm{z}_*}-\eta(\bm{z}_*) \sim\mathcal{N}(0,(\sigma^{[i]}_e)^2)$  independent over agent $i\in\mathcal{V}$ and input $\bm{z}_*\in\bm{\mathcal{Z}}$.
It is obvious that $\underset{t\to\infty}{\lim}\check{\mu}_{\bm{z}_*|\mathcal{D}^{[i]}(t)}=\check{r}^{[i]}_{\bm{z}_*}+\check{e}^{[i]}_{\bm{z}_*}$. $\hfill\blacksquare$
\end{remark}

\subsubsection{Mean analysis of distributed GPR}\label{sec: mean distributed GPR}
Before presenting the results, we derive the solution to the consensus state $\theta^{[i]}_{\bm{z}_*}(t)$, $i\in\mathcal{V}$, in terms of input signal $\Delta r^{[i]}_{\bm{\theta},\bm{z}_*}(t)$.  We also show the decompositions of $r^{[i]}_{\bm{\theta},\bm{z}_*}(t)$ and $\theta^{[i]}_{\bm{z}_*}(t)$, which separate the two terms into real-valued parts and  stochastic parts.

First, we give the solution to $\theta^{[i]}_{\bm{z}_*}(t)$.
Let vectors  $\bm{\theta}_{\bm{z}_*}(t)\triangleq[\theta^{[1]}_{\bm{z}_*}(t), \cdots, \theta^{[n]}_{\bm{z}_*}(t)]^T$ and $\bm{r}_{\bm{\theta},\bm{z}_*}(t)\triangleq[r^{[1]}_{\bm{\theta},\bm{z}_*}(t),\cdots, r^{[n]}_{\bm{\theta},\bm{z}_*}(t)]^T$. Line \ref{ln:dynamic consensus_x} of distributed GPR across the network $\mathcal{V}$ can be represented by discrete linear time-varying (LTV) system:
$
    \bm{\theta}_{\bm{z}_*}(t)=A(t-1)\bm{\theta}_{\bm{z}_*}(t-1)+\Delta \bm{r}_{\bm{\theta},\bm{z}_*}(t).
$
By page 111 in \cite{chen1998linear}, the solution to this system is: 
\begin{align}\label{eq:LTV}
  \bm{\theta}_{\bm{z}_*}(t)=\Phi(t,0)\bm{\theta}_{\bm{z}_*}(0)+\sum_{l=1}^t\Phi(t,l)\Delta \bm{r}_{\bm{\theta},\bm{z}_*}(l),  
\end{align}
where $\Phi(t,l)\triangleq \prod_{\tau=l}^{t-1}A(\tau)$.

Second, we show the decomposition of $\Delta \bm{r}_{\bm{\theta},\bm{z}_*}(l)$ into a  signal depending on $\eta$ and a zero-mean stochastic process. By definition of $r^{[i]}_{\bm{\theta},\bm{z}_*}(t)$ in Line \ref{ln:dynamics consensus_r} in distributed GPR and Remark \ref{remark: lim check mu}, it holds that 
$$r^{[i]}_{\bm{\theta},\bm{z}_*}(t)=\check{\sigma}^{-2}_{\bm{z}_*|\mathcal{D}^{[i]}(t)}(\check{r}^{[i]}_{\bm{z}_*}(t)+\check{e}^{[i]}_{\bm{z}_*}(t))=\hat{r}^{[i]}_{\bm{z}_*}(t)+\hat{e}^{[i]}_{\bm{z}_*}(t),$$ where $\hat{r}^{[i]}_{\bm{z}_*}(t)\triangleq\check{\sigma}^{-2}_{\bm{z}_*|\mathcal{D}^{[i]}(t)}\check{r}^{[i]}_{\bm{z}_*}(t)$ and $\hat{e}^{[i]}_{\bm{z}_*}(t)\triangleq\check{\sigma}^{-2}_{\bm{z}_*|\mathcal{D}^{[i]}(t)}\check{e}^{[i]}_{\bm{z}_*}(t)$ is a Gaussian random variable with zero mean. 
Hence we have
\begin{align}\label{eq: delta r=r+e}
    \Delta r^{[i]}_{\bm{\theta},\bm{z}_*}(t)=\Delta \hat{r}^{[i]}_{\bm{z}_*}(t)+\Delta \hat{e}_{\bm{z}_*}^{[i]}(t).
\end{align}

Denote $\hat{r}^{[i]}_{\bm{z}_*}\triangleq\underset{t\to\infty}{\lim}\hat{r}^{[i]}_{\bm{z}_*}(t)$.  Corollary \ref{corollary: lim check sigma} and Remark \ref{remark: lim check mu} give $\hat{r}^{[i]}_{\bm{z}_*}= \underset{t\to\infty}{\lim}\check{\sigma}^{-2}_{\bm{z}_*|\mathcal{D}^{[i]}(t)}\check{r}^{[i]}_{\bm{z}_*}(t)=(\frac{\sigma_f^2(\sigma^{[i]}_e)^2}{\sigma_f^2+(\sigma^{[i]}_e)^2})^{-1}\check{r}^{[i]}_{\bm{z}_*}$ 
and 
\begin{align}
   \underset{t\to\infty}{\lim}\hat{e}^{[i]}_{\bm{z}_*}(t)=(\frac{\sigma_f^2(\sigma^{[i]}_e)^2}{\sigma_f^2+(\sigma^{[i]}_e)^2})^{-1}\check{e}^{[i]}_{\bm{z}_*}\label{eq: lim check sigma e}
\end{align}
is zero mean Gaussian.

Third, we show the decomposition of \eqref{eq:LTV}. 
The solution \eqref{eq:LTV}  can be decomposed into a solution to FODAC \cite{zhu2010discrete} with respect to a signal depending on $\eta$ and a solution to FODAC with respect to a zero-mean stochastic process.

Let $\hat{\bm{r}}_{\bm{z}_*}(t)\triangleq[\hat{r}^{[1]}_{\bm{z}_*}(t), \cdots, \hat{r}^{[n]}_{\bm{z}_*}(t)]^T$ and $\hat{\bm{e}}_{\bm{z}_*}(t)\triangleq[\hat{e}^{[1]}_{\bm{z}_*}(t), \cdots, \hat{e}^{[n]}_{\bm{z}_*}(t)]^T$. 
By \eqref{eq: delta r=r+e}, we can write \eqref{eq:LTV} as
\begin{align}\label{eq: decompose theta}
 \bm{\theta}_{\bm{z}_*}(t)&=\bm{\theta}_{\bm{z}_*,\bm{r}}(t)+\bm{\theta}_{\bm{z}_*,\bm{e}}(t),   
\\
  \bm{\theta}_{\bm{z}_*,\bm{r}}(t)&\triangleq\Phi(t,0)\bm{\theta}_{\bm{z}_*}(0)+\sum_{l=1}^t\Phi(t,l)\Delta\hat{\bm{r}}_{\bm{z}_*}(l), \nonumber \\
  \bm{\theta}_{\bm{z}_*,\bm{e}}(t)&\triangleq\sum_{l=1}^t\Phi(t,l)\Delta\hat{\bm{e}}_{\bm{z}_*}(l).\nonumber
\end{align}
Then Proposition \ref{prop: lim hat mu} characterizes  the predictive mean.
 \begin{proposition}\label{prop: lim hat mu}
{\em (Prediction decomposition).}
Suppose Assumptions \ref{assmp:Periodical Strong Connectivity},  \ref{assmp:Balanced Communication}, \ref{assmp:Non-degeneracy} and \ref{assum:kernel} hold.  If $\underset{t\to\infty}{\lim}d^{[j]}(t)=0$ for all $j\in\mathcal{V}$, then  for all $\bm{z}_*\in\mathcal{Z}_{agg}$,  $$
 \hat{\mu}_{\bm{z}_*|\mathcal{D}(t)}^{[i]}
=(\hat{\sigma}^{[i]}_{\bm{z}_*|\mathcal{D}(t)})^2\big({\theta}^{[i]}_{\bm{z}_*,\bm{r}}(t)+{\theta}^{[i]}_{\bm{z}_*,\bm{e}}(t)
\big),$$
where $\underset{t\to\infty}{\lim}{\theta}^{[i]}_{\bm{z}_*,\bm{r}}(t)=\frac{1}{n}\sum_{j=1}^n\hat{r}^{[j]}_{\bm{z}_*}$, ${\theta}^{[j]}_{\bm{z}_*,\bm{e}}(t)$ is a Gaussian random variable with zero mean and
$ \underset{t\to\infty}{\lim}\sum_{j=1}^n\theta^{[j]}_{\bm{z}_*,\bm{e}}(t)=\sum_{j=1}^n(\sigma^{[j]}_e)^{-2}e^{[j]}_{\bm{z}_*}.  $

\textbf{Proof: }
By \eqref{eq: decompose theta} and Line \ref{ln:estimation_mu_M} of distributed GPR, we have
$$
 \hat{\mu}_{\bm{z}_*|\mathcal{D}(t)}^{[i]}
=(\hat{\sigma}^{[i]}_{\bm{z}_*|\mathcal{D}(t)})^2\big({\theta}^{[i]}_{\bm{z}_*,\bm{r}}(t)+{\theta}^{[i]}_{\bm{z}_*,\bm{e}}(t)
\big).$$

First, we show that $\underset{t\to\infty}{\lim}{\theta}^{[i]}_{\bm{z}_*,\bm{r}}(t)=\frac{1}{n}\sum_{j=1}^n\hat{r}^{[j]}_{\bm{z}_*}$. Analogous to $\bm{\theta}_{\bm{z}_*}(t)$, $\bm{\theta}_{\bm{z}_*,\bm{r}}(t)$ is the solution for tracking the average of the signal $\hat{\bm{r}}_{\bm{z}_*}(t)$ using FODAC algorithm \cite{zhu2010discrete}.
Since  $\hat{r}^{[i]}_{\bm{z}_*}=(\frac{\sigma_f^2(\sigma^{[i]}_e)^2}{\sigma_f^2+(\sigma^{[i]}_e)^2})^{-1}\check{r}^{[i]}_{\bm{z}_*}$, $\forall i\in\mathcal{V}$, we have  $\underset{t\to\infty}{\lim}\Delta \hat{r}^{[i]}_{\bm{z}_*}(t)=0$.
Combining this with Corollary 3.1 in \cite{zhu2010discrete} gives $$\underset{t\to\infty}{\lim}\bm{\theta}_{\bm{z}_*,\bm{r}}(t)=\underset{t\to\infty}{\lim}\Big(\frac{1}{n}\sum_{j=1}^n\hat{r}^{[j]}_{\bm{z}_*}(t)\Big)\bm{1}_n=\Big(\frac{1}{n}\sum_{j=1}^n\hat{r}^{[j]}_{\bm{z}_*}\Big)\bm{1}_n.$$

Second, we show that ${\theta}^{[j]}_{\bm{z}_*,\bm{e}}(t)$ is a Gaussian random variable with zero mean. Note that $\bm{\theta}_{\bm{z}_*,\bm{e}}(t)=[{\theta}^{[1]}_{\bm{z}_*,\bm{e}}(t), \cdots, {\theta}^{[n]}_{\bm{z}_*,\bm{e}}(t)]^T$.  
Similar to $\bm{\theta}_{\bm{z}_*}(t)$, $\bm{\theta}_{\bm{z}_*,\bm{e}}(t)$ is the solution for tracking the average of $e^{[i]}_{\bm{z}_*}(t)$ using FODAC: 
\begin{align}\label{eq: FODAC theta 2}
   {\theta}^{[i]}_{\bm{z}_*,\bm{e}}(t)=\sum_{l=1 }^na_{ij}(t-1){\theta}^{[j]}_{\bm{z}_*,\bm{e}}(t-1)+\Delta \hat{e}^{[i]}_{\bm{z}_*}(t),
\end{align}
with initial state $\theta^{[i]}_{\bm{z}_*,\bm{e}}(0)=0$. Note that $$
\Delta \hat{e}_{\bm{z}_*}^{[j]}(t)=\check{\sigma}^{-2}_{\bm{z}_*|\mathcal{D}^{[j]}(t)}\check{e}^{[j]}_{\bm{z}_*}(t)-\check{\sigma}^{-2}_{\bm{z}_*|\mathcal{D}^{[j]}(t-1)}\check{e}^{[j]}_{\bm{z}_*}(t-1).
$$
Recall that $\check{e}^{[j]}_{\bm{z}_*}(t)=\frac{\kappa(\rho_{\bm{z}_*}^{\mathcal{Z}^{[j]}(t)})}{\sigma_f^2+(\sigma^{[j]}_e)^2} e^{[j]}_{\bm{z}^{[j]}_*(t)}$ in Remark \ref{remark: lim check mu} where $e^{[j]}_{\bm{z}}$ is a zero-mean Gaussian random variable independent over $j\in\mathcal{V}$ and $\bm{z}\in\mathcal{V}$. Hence $\check{e}^{[j]}_{\bm{z}_*}(t)$ and $\hat{e}_{\bm{z}_*}^{[j]}(t)$ are both zero-mean Gaussian random variables. 
Therefore, it follows from \eqref{eq: FODAC theta 2} that
${\theta}^{[i]}_{\bm{z}_*,\bm{e}}(t)$ is a Gaussian random variable with zero mean for all $t\geqslant 1$ (Theorem 5.5-1, \cite{hogg1977probability}).

Finally, we show that $ \underset{t\to\infty}{\lim}\sum_{j=1}^n\theta^{[j]}_{\bm{z}_*,\bm{e}}(t)=\sum_{j=1}^n(\sigma^{[j]}_e)^{-2}e^{[j]}_{\bm{z}_*}.$  By Assumption \ref{assmp:Balanced Communication} and initial state $\theta^{[j]}_{\bm{z}_*}(0)=r^{[j]}_{\theta,\bm{z}_*}(0)$, which indicates $\theta^{[j]}_{\bm{z}_*,\bm{e}}(0)=\hat{e}^{[j]}_{\bm{z}_*}(0)=0$ for each $j\in\mathcal{V}$, \eqref{eq: FODAC theta 2} renders
$$
\sum_{j=1}^n\theta^{[j]}_{\bm{z}_*,\bm{e}}(t)
=\sum_{j=1}^n\theta^{[j]}_{\bm{z}_*,\bm{e}}(t-1)+\sum_{j=1}^n\Delta \hat{e}_{\bm{z}_*}^{[j]}(t) =\sum_{j=1}^n\hat{e}_{\bm{z}_*}^{[j]}(t)
,
$$ for all $t\geqslant1$. Therefore $\underset{t\to\infty}{\lim}\sum_{j=1}^n\theta^{[j]}_{\bm{z}_*,\bm{e}}(t)=\underset{t\to\infty}{\lim}\sum_{j=1}^n\hat{e}^{[j]}_{\bm{z}_*}(t)$. 
Combining this with \eqref{eq: lim check sigma e} and the definition of $\check{e}^{[j]}_{\bm{z}_*}$ in Remark \ref{remark: lim check mu} gives
$\label{eq: sum of hat e}
  \underset{t\to\infty}{\lim}\sum_{j=1}^n\theta^{[j]}_{\bm{z}_*,\bm{e}}(t)=\sum_{j=1}^n(\sigma^{[j]}_e)^{-2}e^{[j]}_{\bm{z}_*}.  
$
$\hfill\blacksquare$
\end{proposition}

\subsubsection{Mean analysis of fused GPR}\label{sec:mean analysis: fGPR}
This section provides the analysis of predictive mean returned by fused GPR. Recall that Lemma \ref{lmm: lim Z_agg exists} shows that $\underset{t\to\infty}{\lim}\mathcal{Z}^{[i]}_{agg}(t)$ exists.
Hence, the main results in this section are Proposition \ref{prop: Zagg non empty}, where the case $\underset{t\to\infty}{\lim}\mathcal{Z}^{[i]}_{agg}(t)\neq\emptyset$ is discussed, and Lemma \ref{lemma: sufficient condition for improvement}, where a sufficient condition for $\underset{t\to\infty}{\lim}\mathcal{Z}^{[i]}_{agg}(t)\neq\emptyset$ is presented. Then we discuss the case of $\underset{t\to\infty}{\lim}\mathcal{Z}^{[i]}_{agg}(t)=\emptyset$ to conclude the proof of Theorem \ref{thm: mean convergence}.
We first discuss the case of $\underset{t\to\infty}{\lim}\mathcal{Z}^{[i]}_{agg}(t)\neq\emptyset$.

Remark \ref{remark: lim check mu} and Proposition \ref{prop: lim hat mu} respectively render
\begin{align}
   \check{\mu}_{\bm{z}_*|\mathcal{D}^{[i]}(t)}&=\check{r}^{[i]}_{\bm{z}_*}(t)+\check{e}^{[i]}_{\bm{z}_*}(t),\label{eq: lim check mu, mu=0}\\
\hat{\mu}_{\bm{z}_*|\mathcal{D}(t)}^{[i]}
&=\tilde{r}^{[i]}_{\bm{z}_*}(t)+\tilde{e}^{[i]}_{\bm{z}_*}(t),\label{eq: lim hat mu, mu=0}
 \end{align}
where $\tilde{r}^{[i]}_{\bm{z}_*}(t)\triangleq (\hat{\sigma}^{[i]}_{\bm{z}_*|\mathcal{D}(t)})^2{\theta}^{[i]}_{\bm{z}_*,\bm{r}}(t)$ and $\tilde{e}^{[i]}_{\bm{z}_*}(t)\triangleq (\hat{\sigma}^{[i]}_{\bm{z}_*|\mathcal{D}(t)})^2{\theta}^{[i]}_{\bm{z}_*,\bm{e}}(t)$ is zero-mean, $\forall \bm{z}_*\in\bm{\mathcal{Z}}$. 
Lemma \ref{lmm: check and tilde} summarizes the limiting behaviors of the above variables.

\begin{lemma}\label{lmm: check and tilde}
Suppose the same conditions for Proposition \ref{prop: lim hat mu} hold and $d^{[j]}(t)\to0$, $\forall j\in\mathcal{V}$. It holds that $\forall \bm{z}_*\in\bm{\mathcal{Z}}$,
\begin{align*}
    \check{r}^{[i]}_{\bm{z}_*}&=\psi^{[i]}\eta(\bm{z}_*),~ {\lim}_{t\to\infty} \tilde{r}^{[i]}_{\bm{z}_*}(t)=c\eta(\bm{z}_*) \\
   \check{e}^{[i]}_{\bm{z}_*}&=\psi^{[i]}e^{[i]}_{\bm{z}_*},~ {\lim}_{t\to\infty}\sum_{j=1}^n \tilde{e}_{\bm{z}_*}^{[j]}(t)=\sum_{j=1}^n \big(\mu^{-1}_{\chi}\chi^{[j]}\psi^{[j]}e_{\bm{z}_*}^{[j]}\big).
\end{align*}

{\bf Proof: }
Combining the definition of $\psi^{[i]}$ and Remark  \ref{remark: lim check mu} directly renders $\check{r}^{[i]}_{\bm{z}_*}=\psi^{[i]}\eta(\bm{z}_*)$ and $\check{e}^{[i]}_{\bm{z}_*}=\psi^{[i]}e^{[i]}_{\bm{z}_*}$.

Corollary \ref{corollary: lim check sigma} shows that $\underset{t\to\infty}{\lim}\check{\sigma}^{-2}_{\bm{z}_*|\mathcal{D}^{[i]}(t)}=\chi^{[i]}$. Then Corollary \ref{corrollary: distributed variance convergence} and \eqref{eq:agg_var} render 
$$\underset{t\to\infty}{\lim}(\hat{\sigma}^{[i]}_{\bm{z}_*|\mathcal{D}(t)})^2=\underset{t\to\infty}{\lim}(\check{\sigma}^{(agg)}_{\bm{z}_*|\mathcal{D}(t)})^2=\mu^{-1}_{\chi}.$$ 
Combining this with the definition of $\tilde{r}^{[i]}_{\bm{z}_*}(t)$,  Proposition \ref{prop: lim hat mu}, and the above result about $\check{r}^{[i]}_{\bm{z}_*}$ renders
$\label{eq: lim tilde r}
   \underset{t\to\infty}{\lim} \tilde{r}^{[i]}_{\bm{z}_*}(t)=c\eta(\bm{z}_*). 
$

Combining the definition of $\tilde{e}_{\bm{z}_*}^{[j]}(t)$ with Proposition \ref{prop: lim hat mu} gives $\underset{t\to\infty}{\lim}\sum_{j=1}^n \tilde{e}_{\bm{z}_*}^{[j]}(t)=\sum_{j=1}^n \big(\mu^{-1}_{\chi}\chi^{[j]}\psi^{[j]}e_{\bm{z}_*}^{[j]}\big)$. $\hfill\blacksquare$

\end{lemma}

Next we introduce necessary notations to continue the analysis. Since $\underset{t\to\infty}{\lim}\mathcal{Z}^{[i]}_{agg}(t)\neq\emptyset$ and Corollary \ref{corollary: lim check sigma} hold,   $\bm{z}^{[i]}_{agg*}\in\underset{t\to\infty}{\lim}\textrm{proj}(\bm{z}_*,\mathcal{Z}^{[i]}_{agg}(t))$ and $g(\bm{z}_*)\triangleq \underset{t\to\infty}{\lim}g(\bm{z}_*,t)$ exist. Line \ref{ln:fusion mu} of fused GPR gives
$$
    (\tilde{\mu}^{[i]}_{\bm{z}_*|\mathcal{D}(t)}-\eta(\bm{z}_*))^2
    =(\check{\mu}_{\bm{z}_*|\mathcal{D}^{[i]}(t)}-\eta(\bm{z}_*))^2 + s^{[i]}_{\bm{z}_*}(t),
$$
    where $s^{[i]}_{\bm{z}_*}(t)\triangleq s^{[i]}_{\bm{z}_*, 1}(t)+s^{[i]}_{\bm{z}_*,2}(t)$ with
    \begin{align*}
       s^{[i]}_{\bm{z}_*,1}(t)&\triangleq2(\check{\mu}_{\bm{z}_*|\mathcal{D}^{[i]}(t)}-\eta(\bm{z}_*))v^{[i]}_{\bm{z}_*}(t) \mu'^{[i]}_{\bm{z}^{[i]}_{agg*}(t)|\mathcal{D}(t)},\\
       s^{[i]}_{\bm{z}_*,2}(t)&\triangleq\big(v^{[i]}_{\bm{z}_*}(t) \mu'^{[i]}_{\bm{z}^{[i]}_{agg*}(t)|\mathcal{D}(t)}\big)^2.
    \end{align*}
Let $v^{[i]}_{\bm{z}_*}\triangleq\underset{t\to\infty}{\lim}v^{[i]}_{\bm{z}_*}(t)$, whose existence, according to its definition, is guaranteed by the existences of $\bm{z}^{[i]}_{agg*}$, $g(\bm{z}^{[i]}_{agg*})$, and Corollary \ref{corollary: lim check sigma}. Denote
\begin{align*}
    q^{[i]}_{\bm{z}_*,1}&\triangleq(\psi^{[i]}-1)(c-\psi^{[i]})k(\bm{z}_*,\bm{z}^{[i]}_{agg*}),\\
    q^{[i]}_{\bm{z}_*,2}&\triangleq (c
-\psi^{[i]})^2\sigma_f^2+\underset{t\to\infty}{\limsup}\mathbb{E}[(\tilde{e}_{\bm{z}^{[i]}_{agg*}}^{[i]}(t)-\check{e}_{\bm{z}^{[i]}_{agg*}}^{[i]}(t))^2].
\end{align*}
Lemmas \ref{lmm: lim s1} and \ref{lmm: lim s2} characterize the limit  of $\mathbb{E}[s^{[i]}_{\bm{z}_*,1}(t)]$ and $\mathbb{E}[s^{[i]}_{\bm{z}_*,2}(t)]$ in terms of $q^{[i]}_{\bm{z}_*,1}$ and $q^{[i]}_{\bm{z}_*,2}$, respectively.
\begin{lemma}\label{lmm: lim s1}
Suppose the same conditions in Theorem \ref{thm: mean convergence} Part II hold and $d^{[j]}(t)\to0$,  $\forall j\in\mathcal{V}$. If $\underset{t\to\infty}{\lim}\mathcal{Z}^{[i]}_{agg}(t)\neq\emptyset$ for some $i\in\mathcal{V}$, then 
$\underset{t\to\infty}{\limsup}\mathbb{E}[s^{[i]}_{\bm{z}_*,1}(t)]\leqslant2v^{[i]}_{\bm{z}_*}q^{[i]}_{\bm{z}_*,1}$. 

{\bf Proof: }
Outline:  We first give the expression of $\mathbb{E}[s^{[i]}_{\bm{z}_*,1}(t)]$. Then we analyze the limit of each term in the expression $\mathbb{E}[s^{[i]}_{\bm{z}_*,1}(t)]$. Finally, we plug in the terms and derive the upper bound of $\underset{t\to\infty}{\limsup}\mathbb{E}[s^{[i]}_{\bm{z}_*,1}(t)]$.

First, we give the expression of $\mathbb{E}[s^{[i]}_{\bm{z}_*,1}(t)]$.
Using the definition of $\mu'^{[i]}_{\bm{z}^{[i]}_{agg*}(t)|\mathcal{D}(t)}$ in Line \ref{ln: mu'} of fused GPR and plugging in \eqref{eq: lim check mu, mu=0} and \eqref{eq: lim hat mu, mu=0}, 
we have 
\begin{align}\label{eq: mu'=}
   {\mu'}^{[i]}_{\bm{z}^{[i]}_{agg*}(t)|\mathcal{D}(t)}&=\tilde{r}^{[i]}_{\bm{z}^{[i]}_{agg*}(t)}(t)+\tilde{e}^{[i]}_{\bm{z}^{[i]}_{agg*}(t)}(t)\nonumber\\
   &\quad-\check{r}^{[i]}_{\bm{z}^{[i]}_{agg*}(t)}(t)-\check{e}^{[i]}_{\bm{z}^{[i]}_{agg*}(t)}(t),\\
 s^{[i]}_{\bm{z}_*,1}(t)&= 2
 \big(\check{r}^{[i]}_{\bm{z}_*}(t)
 +\check{e}^{[i]}_{\bm{z}_*}(t)
 -\eta(\bm{z}_*)\big)v^{[i]}_{\bm{z}_*}(t)\nonumber\\
 &\quad\cdot 
 \big(\tilde{r}^{[i]}_{\bm{z}^{[i]}_{agg*}(t)}(t)
 +\tilde{e}^{[i]}_{\bm{z}^{[i]}_{agg*}(t)}(t)
 \nonumber\\
 &\qquad-\check{r}^{[i]}_{\bm{z}^{[i]}_{agg*}(t)}(t)
 -\check{e}^{[i]}_{\bm{z}^{[i]}_{agg*}(t)}(t)\big).\nonumber
\end{align}
Note that $v^{[i]}_{\bm{z}_*}(t)=g^{[i]}_{\bm{z}_*}(t)\check{\sigma}^{-2}_{\bm{z}^{[i]}_{agg*}(t)|\mathcal{D}^{[i]}(t)}$, where the right hand side only depends on  $\bm{z}_*$ and $\bm{z}^{[i]}_{agg*}(t)$ instead of $\eta$ or ${e}^{[i]}_{\bm{z}}$ that is random. This gives
\begin{align*}
\mathbb{E}[s^{[i]}_{\bm{z}_*,1}(t)]&=2v^{[i]}_{\bm{z}_*}(t)\mathbb{E}[\check{r}^{[i]}_{\bm{z}_*}(t)\tilde{r}^{[i]}_{\bm{z}^{[i]}_{agg*}(t)}(t)+\check{r}^{[i]}_{\bm{z}_*}(t)\tilde{e}^{[i]}_{\bm{z}^{[i]}_{agg*}(t)}(t)\\
&\quad-\check{r}^{[i]}_{\bm{z}_*}(t)\check{r}^{[i]}_{\bm{z}^{[i]}_{agg*}(t)}(t)-\check{r}^{[i]}_{\bm{z}_*}(t)\check{e}^{[i]}_{\bm{z}^{[i]}_{agg*}(t)}(t)\\
&+\check{e}^{[i]}_{\bm{z}_*}(t)\tilde{r}^{[i]}_{\bm{z}^{[i]}_{agg*}(t)}(t)
 +\check{e}^{[i]}_{\bm{z}_*}(t)\tilde{e}^{[i]}_{\bm{z}^{[i]}_{agg*}(t)}(t)\\
 &
 -\check{e}^{[i]}_{\bm{z}_*}(t)\check{r}^{[i]}_{\bm{z}^{[i]}_{agg*}(t)}(t)
 -\check{e}^{[i]}_{\bm{z}_*}(t)\check{e}^{[i]}_{\bm{z}^{[i]}_{agg*}(t)}(t)\\
 &
-\eta(\bm{z}_*)\tilde{r}^{[i]}_{\bm{z}^{[i]}_{agg*}(t)}(t)
 -\eta(\bm{z}_*)\tilde{e}^{[i]}_{\bm{z}^{[i]}_{agg*}(t)}(t)\\
 &
 +\eta(\bm{z}_*)\check{r}^{[i]}_{\bm{z}^{[i]}_{agg*}(t)}(t)
+\eta(\bm{z}_*)\check{e}^{[i]}_{\bm{z}^{[i]}_{agg*}(t)}(t)].
\end{align*}
Second, we analyze the limit of each term. The limits of the twelve terms in the expectation are given in the claim below.
\begin{claim}\label{claim: limit of expectations}
It holds that
\begin{align*}
   \lim_{t\to\infty}\mathbb{E}[\check{r}^{[i]}_{\bm{z}_*}(t)\tilde{r}^{[i]}_{\bm{z}^{[i]}_{agg*}(t)}(t)]&=\psi^{[i]}ck(\bm{z}_*,\bm{z}^{[i]}_{agg*}); &\\
   \underset{{t\to\infty}}{\lim}\mathbb{E}[\check{r}^{[i]}_{\bm{z}_*}(t)\tilde{e}^{[i]}_{\bm{z}^{[i]}_{agg*}(t)}]&= 0;&\\
   \underset{{t\to\infty}}{\lim}\mathbb{E}[\check{r}^{[i]}_{\bm{z}_*}(t)\check{r}^{[i]}_{\bm{z}^{[i]}_{agg*}(t)}(t)]&=(\psi^{[i]})^2k(\bm{z}_*,\bm{z}^{[i]}_{agg*});&\\
   \underset{{t\to\infty}}{\lim}\mathbb{E}[\check{r}^{[i]}_{\bm{z}_*}(t)\check{e}^{[i]}_{\bm{z}^{[i]}_{agg*}(t)}(t)]&=0; &\\
   \underset{{t\to\infty}}{\lim}\mathbb{E}[\check{e}^{[i]}_{\bm{z}_*}(t)\tilde{r}^{[i]}_{\bm{z}^{[i]}_{agg*}(t)}(t)]&=0;&\\
   \limsup_{t\to\infty}\mathbb{E}[\check{e}^{[i]}_{\bm{z}_*}(t)\tilde{e}^{[i]}_{\bm{z}^{[i]}_{agg*}(t)}(t)]&\leqslant \frac{\chi^{[i]}(\psi^{[i]})^2(\sigma^{[i]}_e)^2}{\mu_{\chi}}, ~\textrm{if } \bm{z}_{*}=\bm{z}^{[i]}_{agg*},&\\
   \lim_{t\to\infty}\mathbb{E}[\check{e}^{[i]}_{\bm{z}_*}(t)\tilde{e}^{[i]}_{\bm{z}^{[i]}_{agg*}(t)}(t)]&=0, ~\textrm{otherwise; }
   \\
   \underset{{t\to\infty}}{\lim} \mathbb{E}[\check{e}^{[i]}_{\bm{z}_*}(t)\check{r}^{[i]}_{\bm{z}^{[i]}_{agg*}(t)}(t)]&=0;&\\
    \underset{t\to\infty}{\lim}\mathbb{E}[\check{e}^{[i]}_{\bm{z}_*}(t)\check{e}^{[i]}_{\bm{z}^{[i]}_{agg*}(t)}(t)]&=(\psi^{[i]})^2(\sigma^{[i]}_e)^2, ~\textrm{if } \bm{z}_*=\bm{z}^{[i]}_{agg*},&\\ \underset{t\to\infty}{\lim}\mathbb{E}[\check{e}^{[i]}_{\bm{z}_*}(t)\check{e}^{[i]}_{\bm{z}^{[i]}_{agg*}(t)}(t)]&=0, ~\textrm{otherwise; } &\\
   \underset{{t\to\infty}}{\lim}\mathbb{E}[\eta(\bm{z}_*)\tilde{r}^{[i]}_{\bm{z}^{[i]}_{agg*}(t)}(t)]&=ck(\bm{z}_*,\bm{z}^{[i]}_{agg*});&\\
  \underset{{t\to\infty}}{\lim} \mathbb{E}[\eta(\bm{z}_*)\tilde{e}^{[i]}_{\bm{z}^{[i]}_{agg*}(t)}(t)]&=0;&\\
    \underset{{t\to\infty}}{\lim}\mathbb{E}[\eta(\bm{z}_*)\check{r}^{[i]}_{\bm{z}^{[i]}_{agg*}(t)}(t)]&=\psi^{[i]}k(\bm{z}_*,\bm{z}^{[i]}_{agg*});&\\
    \underset{{t\to\infty}}{\lim}\mathbb{E}[\eta(\bm{z}_*)\check{e}^{[i]}_{\bm{z}^{[i]}_{agg*}(t)}(t)]&=0.\quad\qquad\qquad\qquad\qquad\qquad\qedsymbol
\end{align*}

\end{claim}

Finally, we find the upper bound of $\underset{t\to\infty}{\limsup}\mathbb{E}[s^{[i]}_{\bm{z}_*,1}(t)]$. Plugging in the terms in Claim \ref{claim: limit of expectations}  gives 
\begin{align*}
\textrm{when }\bm{z}_{*}=\bm{z}^{[i]}_{agg*},~ &\underset{t\to\infty}{\limsup}\mathbb{E}[s^{[i]}_{\bm{z}_*,1}(t)]\\
 &\leqslant 2v^{[i]}_{\bm{z}_*}\Big((\psi^{[i]}-1)(c-\psi^{[i]})k(\bm{z}_*,\bm{z}^{[i]}_{agg*})\\
 &+(\mu^{-1}_{\chi}\chi^{[i]}-1)(\psi^{[i]})^2(\sigma^{[i]}_e)^2\Big);\\
 \textrm{when } \bm{z}_{*}\neq\bm{z}^{[i]}_{agg*}, ~&
\underset{t\to\infty}{\lim}\mathbb{E}[s^{[i]}_{\bm{z}_*,1}(t)]\\
&=2v^{[i]}_{\bm{z}_*}\Big((\psi^{[i]}-1)(c-\psi^{[i]})k(\bm{z}_*,\bm{z}^{[i]}_{agg*})\Big).
\end{align*}
Invoking Lemma \ref{lmm: sufficient for lim s1<0} gives $\underset{t\to\infty}{\limsup}\mathbb{E}[s^{[i]}_{\bm{z}_*,1}(t)]\leqslant 2v^{[i]}_{\bm{z}_*}\Big((\psi^{[i]}-1)(c-\psi^{[i]})k(\bm{z}_*,\bm{z}^{[i]}_{agg*})\Big)
$, $\forall \bm{z}_*\in\bm{\mathcal{Z}}$.
$\hfill\blacksquare$
\end{lemma}

{\em Proof of Claim \ref{claim: limit of expectations}:}
We analyze the limit of each of the twelve terms in  expectation as follows.

Term 1. 
The solution of the LTV system \eqref{eq: decompose theta} gives
\begin{align*}
    {\theta}^{[i]}_{\bm{z}^{[i]}_{agg*}(t),\bm{r}}(t)=&\sum_{j\in\mathcal{V}}\phi_{ij}(t,0){\theta}^{[j]}_{\bm{z}^{[i]}_{agg*}(t)}(0)+\\
    &\sum_{l=1}^t\sum_{j\in\mathcal{V}}\phi_{ij}(t,l)\Delta\check{\sigma}^{-2}_{\bm{z}^{[i]}_{agg*}(t)|\mathcal{D}^{[j]}(t)}\check{r}^{[j]}_{\bm{z}^{[i]}_{agg*}(t)}(t),
\end{align*} where by Remark  \ref{remark: lim check mu}, $\check{r}^{[j]}_{\bm{z}_*}(t)=\psi^{[i]}_{\bm{z}_*}(t)\eta(\bm{z}^{[i]}_*(t))$, $\psi_{\bm{z}_*}^{[i]}(t)\triangleq\frac{\kappa(\rho_{\bm{z}_*}^{\mathcal{Z}^{[i]}(t)})}{\sigma_f^2+(\sigma^{[i]}_e)^2}$, $\forall \bm{z}_*\in\bm{\mathcal{Z}}$. 
Combining this with the definition of  $\tilde{r}^{[i]}_{\bm{z}^{[i]}_{agg*}(t)}(t)$ gives $$\tilde{r}^{[i]}_{\bm{z}^{[i]}_{agg*}(t)}(t)=c_{\bm{z}^{[i]}_{agg*}(t)}^{[i]}(t)\eta(\bm{z}^{[i]}_{agg*,p}(t)),$$ 
where $\bm{z}^{[i]}_{agg*,p}(t)\in\proj(\bm{z}^{[i]}_{agg*}(t),\mathcal{Z}^{[i]}(t))$,
\begin{align*}
    c^{[i]}_{\bm{z}^{[i]}_{agg*}(t)}(t)\triangleq&(\hat{\sigma}^{[i]}_{\bm{z}^{[i]}_{agg*}(t)|\mathcal{D}(t)})^2\\
    &\cdot\Big(\sum_{j\in\mathcal{V}}\phi_{ij}(t,0){\theta}^{[j]}_{\bm{z}^{[i]}_{agg*}(t)}(0)/\eta(\bm{z}^{[i]}_{agg*,p}(t))+\\
    &\sum_{l=1}^t\sum_{j\in\mathcal{V}}\big(\phi_{ij}(t,l)\Delta\check{\sigma}^{-2}_{\bm{z}^{[i]}_{agg*}(t)|\mathcal{D}^{[j]}(t)}\psi_{\bm{z}^{[i]}_{agg*}(t)}^{[j]}(t)\big)\Big)
\end{align*}
Therefore, we have
\begin{align*}
  &\mathbb{E}[\check{r}^{[i]}_{\bm{z}_*}(t)\tilde{r}^{[i]}_{\bm{z}^{[i]}_{agg*}(t)}(t)]\\
  &=\mathbb{E}[\psi_{\bm{z}_*}^{[i]}(t)\eta(\bm{z}^{[i]}_*(t))c_{\bm{z}^{[i]}_{agg*}(t)}^{[i]}(t)\eta(\bm{z}^{[i]}_{agg*,p}(t))]\\
  &=\psi_{\bm{z}_*}^{[i]}(t)c_{\bm{z}^{[i]}_{agg*}(t)}^{[i]}(t)\mathbb{E}[\eta(\bm{z}^{[i]}_*)\eta(\bm{z}^{[i]}_{agg*,p}(t))]\\
  &=\psi_{\bm{z}_*}^{[i]}(t)c_{\bm{z}^{[i]}_{agg*}(t)}^{[i]}(t)k(\bm{z}^{[i]}_*(t),\bm{z}^{[i]}_{agg*,p}(t)).  
\end{align*}
where the last equality follows from Assumption \ref{assmp: covariance}. Note that Lemma \ref{lmm: check and tilde} indicates $\forall \bm{z}\in\bm{\mathcal{Z}}$, $\underset{t\to\infty}{\lim}\psi_{\bm{z}}^{[i]}(t)=\psi^{[i]}$  and $\underset{t\to\infty}{\lim} c_{\bm{z}}^{[i]}(t)=c$. Hence
\begin{align*}
   &\lim_{t\to\infty}\mathbb{E}[\check{r}^{[i]}_{\bm{z}_*}(t)\tilde{r}^{[i]}_{\bm{z}^{[i]}_{agg*}(t)}(t)]\\
   &=\lim_{t\to\infty}\psi^{[i]}(t)c^{[i]}(t)k(\bm{z}^{[i]}_*(t),\bm{z}^{[i]}_{agg*,p}(t))\\
   &=\psi^{[i]}ck(\bm{z}_*,\bm{z}^{[i]}_{agg*}). 
\end{align*}

Terms 3, 9, 11.
Similar to Term 1, we have
\begin{align*}
    \underset{{t\to\infty}}{\lim}\mathbb{E}[\check{r}^{[i]}_{\bm{z}_*}(t)\check{r}^{[i]}_{\bm{z}^{[i]}_{agg*}(t)}(t)]&=(\psi^{[i]})^2k(\bm{z}_*,\bm{z}^{[i]}_{agg*})\\
    \underset{{t\to\infty}}{\lim}\mathbb{E}[\eta(\bm{z}_*)\tilde{r}^{[i]}_{\bm{z}^{[i]}_{agg*}(t)}(t)]&=ck(\bm{z}_*,\bm{z}^{[i]}_{agg*})\\
    \underset{{t\to\infty}}{\lim}\mathbb{E}[\eta(\bm{z}_*)\check{r}^{[i]}_{\bm{z}^{[i]}_{agg*}(t)}(t)]&=\psi^{[i]}k(\bm{z}_*,\bm{z}^{[i]}_{agg*}).
\end{align*}

Term 2. By definitions, $\check{r}^{[i]}_{\bm{z}_*}(t)$ and $\tilde{e}^{[i]}_{\bm{z}^{[i]}_{agg*}(t)}(t)$  only depend on  $\eta(\bm{z}^{[i]}_*(t))$  and ${e}^{[i]}_{\bm{z}^{[i]}_{agg*}(t)}$, respectively. Since $\tilde{e}^{[i]}_{\bm{z}^{[i]}_{agg*}(t)}$ is zero-mean, we have  $\forall t\geqslant 1$,
$$
\mathbb{E}[\check{r}^{[i]}_{\bm{z}_*}(t)\tilde{e}^{[i]}_{\bm{z}^{[i]}_{agg*}(t)}]=\mathbb{E}_{\eta}[\check{r}^{[i]}_{\bm{z}_*}(t)]\mathbb{E}_{{e}^{[i]}_{\bm{z}^{[i]}_{agg*}(t)}}[\tilde{e}^{[i]}_{\bm{z}^{[i]}_{agg*}(t)}(t)]=0. 
$$

Terms 4, 5, 7, 10, 12. Similar to Term 2,
\begin{align*}
  &\mathbb{E}[\check{r}^{[i]}_{\bm{z}_*}(t)\check{e}^{[i]}_{\bm{z}^{[i]}_{agg*}(t)}(t)]= \mathbb{E}[\check{e}^{[i]}_{\bm{z}_*}(t)\tilde{r}^{[i]}_{\bm{z}^{[i]}_{agg*}(t)}(t)]\\
  &= \mathbb{E}[\check{e}^{[i]}_{\bm{z}_*}(t)\check{r}^{[i]}_{\bm{z}^{[i]}_{agg*}(t)}(t)]= \mathbb{E}[\eta(\bm{z}_*)\tilde{e}^{[i]}_{\bm{z}^{[i]}_{agg*}(t)}(t)]\\
  &= \mathbb{E}[\eta(\bm{z}_*)\check{e}^{[i]}_{\bm{z}^{[i]}_{agg*}(t)}(t)]=0, \quad \forall t\geqslant 1.  
\end{align*}

Terms 6, 8.
Since $e_{\bm{z}_*}^{[i]}$ and $e_{\bm{z}_*}^{[j]}$ are independent zero-mean measurement noises, we have $\mathbb{E}[e_{\bm{z}_*}^{[i]}e_{\bm{z}_*}^{[j]}]=0$. 
Since Remark \ref{remark: lim check mu} states that $\check{e}^{[i]}_{\bm{z}_*}(t)=\psi^{[i]}(t)e_{\bm{z}_*}^{[i]}$, we  have $\check{e}^{[i]}_{\bm{z}_*}(t)$ and $\check{e}^{[j]}_{\bm{z}_*}(t)$ are also zero-mean and independent. Therefore, $$\mathbb{E}[\check{e}^{[i]}_{\bm{z}_*}(t)\check{e}^{[j]}_{\bm{z}_*}(t)]=0,~\forall j\neq i.$$ Since $\hat{e}^{[j]}_{\bm{z}_*}(t)$ is linear in $\check{e}^{[j]}_{\bm{z}_*}(t)$, we have $$
\mathbb{E}[\check{e}^{[i]}_{\bm{z}_*}(t)\hat{e}^{[j]}_{\bm{z}_*}(t)]=0,\forall j\neq i.
$$
Recall that $\tilde{e}^{[i]}_{\bm{z}^{[i]}_{agg*}(t)}(t)\triangleq (\hat{\sigma}^{[i]}_{\bm{z}^{[i]}_{agg*}(t)|\mathcal{D}(t)})^2{\theta}^{[i]}_{\bm{z}^{[i]}_{agg*}(t),\bm{e}}(t)$ and the LTV solution gives $\theta^{[i]}_{\bm{z},\bm{e}}(t)\triangleq\sum_{l=1}^t\sum_{j=1}^n\phi_{ij}(t,l)\Delta\hat{e}^{[j]}_{\bm{z}}(l)$. 
Then 
\begin{align}\label{eq: E check e tilde e}
    &\mathbb{E}[\check{e}^{[i]}_{\bm{z}_*}(t)\tilde{e}^{[i]}_{\bm{z}^{[i]}_{agg*}(t)}(t)]\nonumber\\
    &=(\hat{\sigma}^{[i]}_{\bm{z}^{[i]}_{agg*}(t)|\mathcal{D}(t)})^2\mathbb{E}[\check{e}^{[i]}_{\bm{z}_*}(t)(\sum_{l=1}^t\sum_{j=1}^n\phi_{ij}(t,l)\Delta\hat{e}^{[j]}_{\bm{z}^{[i]}_{agg*}(t)}(l))]\nonumber\\
    &=(\hat{\sigma}^{[i]}_{\bm{z}^{[i]}_{agg*}(t)|\mathcal{D}(t)})^2\mathbb{E}[\check{e}^{[i]}_{\bm{z}_*}(t)(\sum_{l=1}^t\phi_{ii}(t,l)\Delta\hat{e}^{[i]}_{\bm{z}^{[i]}_{agg*}(t)}(l))]\nonumber\\
    &=(\hat{\sigma}^{[i]}_{\bm{z}^{[i]}_{agg*}(t)|\mathcal{D}(t)})^2\phi_{ii}(t,t)\mathbb{E}[\check{e}^{[i]}_{\bm{z}_*}(t)\hat{e}^{[i]}_{\bm{z}^{[i]}_{agg*}(t)}(t)]\nonumber\\
    &=(\hat{\sigma}^{[i]}_{\bm{z}^{[i]}_{agg*}(t)|\mathcal{D}(t)})^2\check{\sigma}^{-2}_{\bm{z}^{[i]}_{agg*}(t)|\mathcal{D}^{[i]}(t)}\phi_{ii}(t,t)\mathbb{E}[\check{e}^{[i]}_{\bm{z}_*}(t)\check{e}^{[i]}_{\bm{z}^{[i]}_{agg*}(t)}(t)]
\end{align}
where the third equality follows from the initial condition $\hat{e}^{[j]}_{\bm{z}^{[i]}_{agg*}(t)}(0)=0$ and $\phi_{ij}(t,l)=\phi_{ij}(t,l')$ for all $0<l,l'\leqslant t$ implied by Assumption \ref{assmp: constant A}.

%

The independence of $e_{\bm{z}}^{[i]}$ over $\bm{z}\in\bm{\mathcal{Z}}$ gives:
\begin{align*}
    \textrm{if } \bm{z}^{[i]}_*(t)=\bm{z}^{[i]}_{agg*}(t),~&\mathbb{E}[\check{e}^{[i]}_{\bm{z}_*}(t)\check{e}^{[i]}_{\bm{z}^{[i]}_{agg*}(t)}(t)]=\mathbb{E}[\check{e}^{[i]}_{\bm{z}_*}(t)\check{e}^{[i]}_{\bm{z}_*}(t)];\\
    \textrm{otherwise, }
&\mathbb{E}[(\check{e}^{[i]}_{\bm{z}_*}(t))^2]=0.
\end{align*}
Notice $\underset{t\to\infty}{\lim}\bm{z}^{[i]}_*(t)=\bm{z}_*$ and $\underset{t\to\infty}{\lim} \bm{z}^{[i]}_{agg*,p}(t)=\bm{z}^{[i]}_{agg*}$. Hence 
\begin{align*}
    \textrm{if } \bm{z}_*=\bm{z}^{[i]}_{agg*}, ~&\underset{t\to\infty}{\lim}\mathbb{E}[\check{e}^{[i]}_{\bm{z}_*}(t)\check{e}^{[i]}_{\bm{z}^{[i]}_{agg*}(t)}(t)]=\mathbb{E}[\check{e}^{[i]}_{\bm{z}_*}\check{e}^{[i]}_{\bm{z}_*}];\\ \textrm{otherwise, } &\underset{t\to\infty}{\lim}\mathbb{E}[\check{e}^{[i]}_{\bm{z}_*}(t)\check{e}^{[i]}_{\bm{z}^{[i]}_{agg*}(t)}(t)]=0.
\end{align*}
The definition of $\check{e}^{[i]}_{\bm{z}_*}$ in Remark \ref{remark: lim check mu} gives $$
\mathbb{E}[\check{e}^{[i]}_{\bm{z}_*}\check{e}^{[i]}_{\bm{z}_*}]=(\psi^{[i]})^2\mathbb{E}[(e_{\bm{z}_*}^{[i]})^2]=(\psi^{[i]})^2(\sigma^{[i]}_e)^2.
$$  
Hence for Term 8, we have
\begin{align}\label{eq: term 8}
    \textrm{if } \bm{z}_*=\bm{z}^{[i]}_{agg*}, ~&\underset{t\to\infty}{\lim}\mathbb{E}[\check{e}^{[i]}_{\bm{z}_*}(t)\check{e}^{[i]}_{\bm{z}^{[i]}_{agg*}(t)}(t)]=(\psi^{[i]})^2(\sigma^{[i]}_e)^2;\nonumber\\ \textrm{otherwise, } &\underset{t\to\infty}{\lim}\mathbb{E}[\check{e}^{[i]}_{\bm{z}_*}(t)\check{e}^{[i]}_{\bm{z}^{[i]}_{agg*}(t)}(t)]=0.
\end{align}

By Corollary \ref{corollary: lim check sigma},  $$\chi^{[i]}=\underset{t\to\infty}{\lim}\check{\sigma}^{-2}_{\bm{z}|\mathcal{D}^{[i]}(t)},\quad \forall \bm{z}\in\bm{\mathcal{Z}},$$
and by Corollary  \ref{corrollary: distributed variance convergence}, $$\mu^{-1}_{\chi}=\underset{t\to\infty}{\lim}(\check{\sigma}^{(agg)}_{\bm{z}|\mathcal{D}(t)})^2=\underset{t\to\infty}{\lim}(\hat{\sigma}^{[i]}_{\bm{z}|\mathcal{D}(t)})^2,\quad \forall \bm{z}\in\bm{\mathcal{Z}}. $$
Note that $\phi_{ii}(t,t)\leqslant 1$ implied by   Assumption \ref{assmp: constant A}.
Combining these with \eqref{eq: E check e tilde e} and \eqref{eq: term 8} gives Term 6
\begin{align*}
    &\limsup_{t\to\infty}\mathbb{E}[\check{e}^{[i]}_{\bm{z}_*}(t)\tilde{e}^{[i]}_{\bm{z}^{[i]}_{agg*}(t)}(t)]\\
    &\leqslant\underset{t\to\infty}{\lim}(\hat{\sigma}^{[i]}_{\bm{z}^{[i]}_{agg*}(t)|\mathcal{D}(t)})^2\check{\sigma}^{-2}_{\bm{z}^{[i]}_{agg*}(t)|\mathcal{D}^{[i]}(t)}\mathbb{E}[\check{e}^{[i]}_{\bm{z}_*}(t)\check{e}^{[i]}_{\bm{z}^{[i]}_{agg*}(t)}(t)]\\
    &\leqslant \mu^{-1}_{\chi}\chi^{[i]}\mathbb{E}[\check{e}^{[i]}_{\bm{z}_*}\check{e}^{[i]}_{\bm{z}_*}]=\mu^{-1}_{\chi}\chi^{[i]}(\psi^{[i]})^2(\sigma^{[i]}_e)^2
\end{align*}
when $\bm{z}_{*}=\bm{z}^{[i]}_{agg*}$; otherwise $\underset{t\to\infty}{\lim}\mathbb{E}[\check{e}^{[i]}_{\bm{z}_*}(t)\tilde{e}^{[i]}_{\bm{z}^{[i]}_{agg*}(t)}(t)]=0$.
$\hfill\qedsymbol$

Lemma \ref{lmm: lim s2} shows the limiting behavior of $\mathbb{E}[s^{[i]}_{\bm{z}_*,2}(t)]$.
\begin{lemma}\label{lmm: lim s2}
Suppose the same conditions in Theorem \ref{thm: mean convergence} part II hold and $d^{[j]}(t)\to0$,  $\forall j\in\mathcal{V}$. If $\underset{t\to\infty}{\lim}\mathcal{Z}^{[i]}_{agg}(t)\neq\emptyset$ for some $i\in\mathcal{V}$, then $\underset{t\to\infty}{\limsup}\mathbb{E}[s^{[i]}_{\bm{z}_*,2}(t)]=(v^{[i]}_{\bm{z}_*})^2q^{[i]}_{\bm{z}_*,2}$.

{\bf Proof: } The proof is done by combining \eqref{eq: mu'=} with the definition of $s^{[i]}_{\bm{z}_*,2}(t)$ and applying similar term-by-term analysis as in Lemma \ref{lmm: lim s1}. $\hfill\blacksquare$
\end{lemma}



Proposition \ref{prop: Zagg non empty} shows the limiting behavior of $\mathbb{E}[s^{[i]}_{\bm{z}_*}(t)]$ when $\displaystyle\underset{t\to\infty}{\lim}\mathcal{Z}^{[i]}_{agg}(t)\neq\emptyset$.
\begin{proposition}\label{prop: Zagg non empty}
Suppose the same conditions in Theorem \ref{thm: mean convergence} Part II hold  and $d^{[j]}(t)\to0$, $\forall j\in\mathcal{V}$. If $\underset{{t\to\infty}}{\lim}\mathcal{Z}^{[i]}_{agg}(t)\neq\emptyset$ for some $i\in\mathcal{V}$, then
$\underset{t\to\infty}{\limsup}\mathbb{E}[s^{[i]}_{\bm{z}_*}(t)]\leqslant-\mathcal{O}\Big(k(\bm{z}_*,\bm{z}^{[i]}_{agg*})\Big)<0$. 

\textbf{Proof:}
Denote $
b^{[i]}_{\bm{z}_*}\triangleq-2q^{[i]}_{\bm{z}_*,1}/q^{[i]}_{\bm{z}_*,2}
$. Then Lemma \ref{lmm: lim s1} and \ref{lmm: lim s2} imply
\begin{align}\label{eq: limsup s<=}
\limsup_{t\to\infty}\mathbb{E}[s^{[i]}_{\bm{z}_*}(t)]\leqslant v^{[i]}_{\bm{z}_*}q^{[i]}_{\bm{z}_*,2}(-b^{[i]}_{\bm{z}_*}+v^{[i]}_{\bm{z}_*}).    
\end{align}


We first show that $0<v^{[i]}_{\bm{z}_*}<b^{[i]}_{\bm{z}_*}$.
The definition of $v^{[i]}_{\bm{z}_*}(t)$ on Line \ref{ln: fusion v} in fused GPR gives
\begin{align}\label{eq: v* lim}
   v^{[i]}_{\bm{z}_*}=\underset{t\to\infty}{\lim}\{g^{[i]}_{\bm{z}_*}(t)\check{\sigma}^{-2}_{\bm{z}^{[i]}_{agg*}|\mathcal{D}^{[i]}(t)}\}. 
\end{align}
Corollary \ref{corollary: lim check sigma} renders $\underset{t\to\infty}{\lim}\check{\sigma}^{-2}_{\bm{z}^{[i]}_{agg*}|\mathcal{D}^{[i]}(t)}>0$ and
$
   \underset{t\to\infty}{\lim}g^{[i]}_{\bm{z}_*}(t)>0; 
$
hence $v^{[i]}_{\bm{z}_*}>0$. 
This also indicates that 
\begin{align}\label{eq: v = lim/lim}
    v^{[i]}_{\bm{z}_*}&=\frac{\underset{t\to\infty}{\lim}g^{[i]}_{\bm{z}_*}(t)}{\underset{t\to\infty}{\lim}\check{\sigma}^2_{\bm{z}^{[i]}_{agg*}(t)|\mathcal{D}^{[i]}(t)}}.
\end{align}
 By boundedness in Assumption \ref{assum:kernel} and  Corollary \ref{corollary: lim check sigma}, Line \ref{ln: g} in fused GPR renders 
\begin{align}\label{eq: lim g}
    \underset{t\to\infty}{\lim}g^{[i]}_{\bm{z}_*}(t)&=(1-\psi^{[i]})(c-\psi^{[i]})k(\bm{z}_*,\bm{z}^{[i]}_{agg*})\sigma_f^{-2}\\
    &=-q^{[i]}_{\bm{z}_*,1}/\sigma_f^{2}\nonumber.
\end{align}
The following claim characterizes the lower bound of $\underset{t\to\infty}{\lim}\check{\sigma}^2_{\bm{z}^{[i]}_{agg*}(t)|\mathcal{D}^{[i]}(t)}$ in terms of $q^{[i]}_{\bm{z}_*,2}$.
\begin{claim}\label{claim: check sigma> q2}
It holds that $\underset{t\to\infty}{\lim}\check{\sigma}^2_{\bm{z}^{[i]}_{agg*}(t)|\mathcal{D}^{[i]}(t)}>\frac{q^{[i]}_{\bm{z}_*,2}}{2(c\sigma_f^2
-\psi^{[i]})}>0.$ $\hfill\qedsymbol$
\end{claim}

 Combining \eqref{eq: lim g} and Claim \ref{claim: check sigma> q2}  with \eqref{eq: v = lim/lim} gives
\begin{align*}
  v^{[i]}_{\bm{z}_*}&=\frac{\underset{t\to\infty}{\lim}g^{[i]}_{\bm{z}_*}(t)}{\underset{t\to\infty}{\lim}\check{\sigma}^2_{\bm{z}^{[i]}_{agg*}(t)|\mathcal{D}^{[i]}(t)}}=\frac{-q^{[i]}_{\bm{z}_*,1}/\sigma_f^{2}}{\underset{t\to\infty}{\lim}\check{\sigma}^2_{\bm{z}^{[i]}_{agg*}(t)|\mathcal{D}^{[i]}(t)}}\\
    &<\frac{-q^{[i]}_{\bm{z}_*,1}/\sigma_f^{2}}{q^{[i]}_{\bm{z}_*,2}/2(c\sigma_f^2
-\psi^{[i]})}<\frac{-q^{[i]}_{\bm{z}_*,1}/\sigma_f^{2}}{q^{[i]}_{\bm{z}_*,2}/2\sigma_f^2}=b^{[i]}_{\bm{z}_*},
\end{align*}
noticing that $0<c<1$.


Notice that Lemma \ref{lmm: sufficient for lim s1<0} implies $q^{[i]}_{\bm{z}_*,2}>0$. Since $0<v^{[i]}_{\bm{z}_*}
    < b^{[i]}_{\bm{z}_*}$,  \eqref{eq: limsup s<=} implies  $\underset{t\to\infty}{\limsup}\mathbb{E}[s^{[i]}_{\bm{z}_*}(t)]<0$.  
    Combining  \eqref{eq: v* lim} and \eqref{eq: lim g} renders $v^{[i]}_{\bm{z}_*}=\mathcal{O}\Big(k(\bm{z}_*,\bm{z}^{[i]}_{agg*})\Big)$.
    Combining this with \eqref{eq: limsup s<=} renders $\underset{t\to\infty}{\limsup}\mathbb{E}[s^{[i]}_{\bm{z}_*}(t)]\leqslant-\mathcal{O}\Big(k(\bm{z}_*,\bm{z}^{[i]}_{agg*})\Big)$. 
$\hfill\blacksquare$

\end{proposition}

{\em Proof of Claim \ref{claim: check sigma> q2}:}
Outline: Based on the definition of $q^{[i]}_{\bm{z}_*,2}$, the proof is broken down into two parts: deriving the upper bound of $(c
-\psi^{[i]})\sigma_f^2$ and the upper bound of $\underset{t\to\infty}{\limsup}\mathbb{E}[\big( \tilde{e}_{\bm{z}^{[i]}_{agg*}(t)}^{[i]}(t)-\check{e}_{\bm{z}^{[i]}_{agg*}(t)}^{[i]}(t)\big)^2]$.

First, we derive the upper bound of $(c
-\psi^{[i]})\sigma_f^2$. Since $\psi^{[j]}<1$, $\forall j\in\mathcal{V}$ and
 $c$ is a convex combination of $\psi^{[j]}$, we have $c<1$ and 
 \begin{align}\label{eq: c-psi<check sigma}
    &(c
-\psi^{[i]})\sigma_f^2< (1
-\psi^{[i]})\sigma_f^2\nonumber\\
&=\sigma_f^2-\frac{\sigma_f^4}{\sigma_f^2+(\sigma^{[i]}_e)^2}=\underset{t\to\infty}{\lim}\check{\sigma}^2_{\bm{z}^{[i]}_{agg*}(t)|\mathcal{D}^{[i]}(t)},
 \end{align}
where the equality follows from  Corollary \ref{corollary: lim check sigma}.

Second, we derive the upper bound of $\allowbreak \underset{t\to\infty}{\limsup} \mathbb{E}[\big( \tilde{e}_{\bm{z}^{[i]}_{agg*}(t)}^{[i]}(t)-\check{e}_{\bm{z}^{[i]}_{agg*}(t)}^{[i]}(t)\big)^2]$.  
Consider the following properties regarding the covariances involving $\tilde{e}_{\bm{z}_*}^{[i]}(t)$ and $\check{e}_{\bm{z}_*}^{[j]}(t)$, where the proofs are at the end. 
\begin{claim}\label{claim: cov tilde check}
It holds that  $\cov\big(\tilde{e}_{\bm{z}_*}^{[j]}(t),\check{e}_{\bm{z}_*}^{[i]}(t)\big)= \mathbb{E}[\tilde{e}_{\bm{z}_*}^{[j]}(t)\check{e}_{\bm{z}_*}^{[i]}(t)]\geqslant 0$,  $\forall t\geqslant 1$, $\bm{z}_*\in\bm{\mathcal{Z}}$, $i, j\in\mathcal{V}$. $\hfill\qedsymbol$
\end{claim}
\begin{claim}\label{claim: cov tilde + check}
It holds that $\cov(\tilde{e}_{\bm{z}_*}^{[i]}(t)+\check{e}_{\bm{z}_*}^{[i]}(t),\tilde{e}_{\bm{z}_*}^{[j]}(t)+\check{e}_{\bm{z}_*}^{[j]}(t))\geqslant 0$, $\forall t\geqslant 1$, $\bm{z}_*\in\bm{\mathcal{Z}}$, $i, j\in\mathcal{V}$. $\hfill\qedsymbol$
\end{claim}


Since $\tilde{e}_{\bm{z}^{[i]}_{agg*}(t)}^{[j]}(t)$ and $\check{e}_{\bm{z}^{[i]}_{agg*}(t)}^{[j]}(t)$ are zero-mean, we have $$
\mathbb{E}[\big( \tilde{e}_{\bm{z}^{[i]}_{agg*}(t)}^{[j]}(t)-\check{e}_{\bm{z}^{[i]}_{agg*}(t)}^{[j]}(t)\big)^2]=\var\big( \tilde{e}_{\bm{z}^{[i]}_{agg*}(t)}^{[j]}(t)-\check{e}_{\bm{z}^{[i]}_{agg*}(t)}^{[j]}(t)\big).
$$
By Claim \ref{claim: cov tilde check}, we have 
\begin{align*}
    &\var\big( \tilde{e}_{\bm{z}^{[i]}_{agg*}(t)}^{[j]}(t)-\check{e}_{\bm{z}^{[i]}_{agg*}(t)}^{[j]}(t)\big)\\
    &=\var\big( \tilde{e}_{\bm{z}^{[i]}_{agg*}(t)}^{[j]}(t)\big)-2\cov\big(\tilde{e}_{\bm{z}^{[i]}_{agg*}(t)}^{[j]}(t),\check{e}_{\bm{z}^{[i]}_{agg*}(t)}^{[j]}(t)\big)\\
    &\quad+\var\big(\check{e}_{\bm{z}^{[i]}_{agg*}(t)}^{[j]}(t)\big)\\
    &\leqslant\var\big( \tilde{e}_{\bm{z}^{[i]}_{agg*}(t)}^{[j]}(t)+\check{e}_{\bm{z}^{[i]}_{agg*}(t)}^{[j]}(t)\big)\\
    &\leqslant \sum_{j=1}^n~\var\big( \tilde{e}_{\bm{z}^{[i]}_{agg*}(t)}^{[j]}(t)+\check{e}_{\bm{z}^{[i]}_{agg*}(t)}^{[j]}(t)\big).
\end{align*}
By Claim \ref{claim: cov tilde + check}, we have 
\begin{align*}
&\sum_{j=1}^n~\var\big( \tilde{e}_{\bm{z}^{[i]}_{agg*}(t)}^{[j]}(t)+\check{e}_{\bm{z}^{[i]}_{agg*}(t)}^{[j]}(t)\big)\\
&\leqslant \sum_{j=1}^n~\var\big( \tilde{e}_{\bm{z}^{[i]}_{agg*}(t)}^{[j]}(t)+\check{e}_{\bm{z}^{[i]}_{agg*}(t)}^{[j]}(t)\big)\\
&\quad+\sum_{j=1}^n\sum_{l\neq j}\cov\big(\tilde{e}_{\bm{z}^{[i]}_{agg*}(t)}^{[j]}(t)+\check{e}_{\bm{z}^{[i]}_{agg*}(t)}^{[j]}(t),\\
&\qquad\tilde{e}_{\bm{z}^{[i]}_{agg*}(t)}^{[l]}(t)+\check{e}_{\bm{z}^{[i]}_{agg*}(t)}^{[l]}(t)\big) \\
&=
\var(\sum_{j=1}^n \big( \tilde{e}_{\bm{z}^{[i]}_{agg*}(t)}^{[j]}(t)+\check{e}_{\bm{z}^{[i]}_{agg*}(t)}^{[j]}(t)\big)).    
\end{align*}

The above three statements render
\begin{align}\label{eq: E < var sum}
&\mathbb{E}[\big( \tilde{e}_{\bm{z}^{[i]}_{agg*}(t)}^{[j]}(t)-\check{e}_{\bm{z}^{[i]}_{agg*}(t)}^{[j]}(t)\big)^2]\nonumber
\\
&\leqslant 
\var(\sum_{j=1}^n \big( \tilde{e}_{\bm{z}^{[i]}_{agg*}(t)}^{[j]}(t)+\check{e}_{\bm{z}^{[i]}_{agg*}(t)}^{[j]}(t)\big)).
\end{align}

By Lemma \ref{lmm: check and tilde}, we have $\underset{t\to\infty}{\lim}\sum_{j=1}^n\big( \tilde{e}_{\bm{z}^{[i]}_{agg*}(t)}^{[j]}(t)+\check{e}_{\bm{z}^{[i]}_{agg*}(t)}^{[j]}(t)\big)=\sum_{j=1}^n\Big( \mu^{-1}_{\chi}\chi^{[j]}\psi^{[j]}+\psi^{[j]}\Big)e_{\bm{z}^{[i]}_{agg*}}^{[j]}$. 
Taking limit on both sides of \eqref{eq: E < var sum} renders
\begin{align}\label{eq: limsup E< psi sigma max}
    &\limsup_{t\to\infty}\mathbb{E}[\big( \tilde{e}_{\bm{z}^{[i]}_{agg*}(t)}^{[j]}(t)-\check{e}_{\bm{z}^{[i]}_{agg*}(t)}^{[j]}(t)\big)^2]\nonumber\\
    &\leqslant \underset{t\to\infty}{\lim}\var(\sum_{j=1}^n \big( \tilde{e}_{\bm{z}^{[i]}_{agg*}(t)}^{[j]}(t)+\check{e}_{\bm{z}^{[i]}_{agg*}(t)}^{[j]}(t)\big))\nonumber\\
&=\var(\sum_{j=1}^n\Big( \mu^{-1}_{\chi}\chi^{[j]}\psi^{[j]}+\psi^{[j]}\Big)e_{\bm{z}^{[i]}_{agg*}}^{[j]})\nonumber\\
&=\sum_{j=1}^n\Big(( \mu^{-1}_{\chi}\chi^{[j]}+1)^2(\psi^{[j]})^2(\sigma_e^{[j]})^2\Big)\nonumber\\
&\leqslant \sum_{j=1}^n\Big(( \mu^{-1}_{\chi}\chi^{[j]}+1)^2\Big)(\psi^{\max})^2(\sigma_e^{\max})^2.
\end{align}

Note that $\sum_{j=1}^n( \mu^{-1}_{\chi}\chi^{[j]}+1)^2=\frac{\sigma^2_{\chi}}{\mu^2_{\chi}}$ based on the definitions in Section \ref{sec: hyperparameter}.
Lemma \ref{lmm: sufficient for lim s1<0} indicates
$c-\psi^{[i]}>0$. Since $\sigma_f^2\geqslant1$ in Section \ref{sec: hyperparameter}, we have $c\sigma_f^2-\psi^{[i]}>0$. By definition of $\epsilon_+$ in Section \ref{sec: hyperparameter}, we further have $c\sigma_f^2-\psi^{[i]}\geqslant \epsilon_+$. 
Combining this with \eqref{eq: limsup E< psi sigma max} gives
\begin{align*}
      &\limsup_{t\to\infty}\mathbb{E}[\big( \tilde{e}_{\bm{z}^{[i]}_{agg*}(t)}^{[i]}(t)-\check{e}_{\bm{z}^{[i]}_{agg*}(t)}^{[i]}(t)\big)^2]\\
    &\leqslant\frac{\sigma^2_{\chi}(\psi^{\max})^2(\sigma_e^{\max})^2(c\sigma_f^2-\psi^{[i]})}{\mu^2_{\chi}\epsilon_+}\\
    &\leqslant(\psi^{\max})^2(\sigma_e^{\min})^2(c\sigma_f^2-\psi^{[i]}),  
\end{align*}
where the last inequality follows from \eqref{eq: sigma_f condition}. Notice that $\psi^{\max}=\frac{\sigma_f^2}{\sigma_f^2+(\sigma_e^{[\min]})^2}<1$ and hence
\begin{align*}
    (\psi^{\max})^2(\sigma_e^{\min})^2&<\frac{\sigma_f^2(\sigma_e^{[\min]})^2}{\sigma_f^2+(\sigma_e^{[\min]})^2}=\sigma_f^2-\frac{\sigma_f^4}{\sigma_f^2+(\sigma_e^{[\min]})^2}\\
    &\leqslant \sigma_f^2-\frac{\sigma_f^4}{\sigma_f^2+(\sigma^{[i]}_e)^2}=\underset{t\to\infty}{\lim}\check{\sigma}^2_{\bm{z}^{[i]}_{agg*}(t)|\mathcal{D}^{[i]}(t)}
\end{align*}
where the last equality follows from Corollary \ref{corollary: lim check sigma}. Therefore,
\begin{align}\label{eq: limsup E<check sigma}
    &\underset{t\to\infty}{\limsup}\mathbb{E}[\big( \tilde{e}_{\bm{z}^{[i]}_{agg*}(t)}^{[i]}(t)-\check{e}_{\bm{z}^{[i]}_{agg*}(t)}^{[i]}(t)\big)^2]\nonumber
    \\    &
    \leqslant\underset{t\to\infty}{\lim}\check{\sigma}^2_{\bm{z}^{[i]}_{agg*}(t)|\mathcal{D}^{[i]}(t)}(c\sigma_f^2-\psi^{[i]}).
\end{align}

Finally, we find the lower bound of $\underset{t\to\infty}{\lim}\check{\sigma}^2_{\bm{z}^{[i]}_{agg*}(t)|\mathcal{D}^{[i]}(t)}.$
Since $\sigma_f^2\geqslant 1$, $c\sigma_f^2>c$. Combining this with \eqref{eq: c-psi<check sigma} and \eqref{eq: limsup E<check sigma} renders
\begin{align*}
   q^{[i]}_{\bm{z}_*,2}&=(c
-\psi^{[i]})^2\sigma_f^2+\underset{t\to\infty}{\limsup}\mathbb{E}[(\tilde{e}_{\bm{z}^{[i]}_{agg*}}^{[i]}(t)-\check{e}_{\bm{z}^{[i]}_{agg*}}^{[i]}(t))^2]\\
&<2(c\sigma_f^2
-\psi^{[i]})\underset{t\to\infty}{\lim}\check{\sigma}^2_{\bm{z}^{[i]}_{agg*}(t)|\mathcal{D}^{[i]}(t)},
\end{align*}
and obviously $ q^{[i]}_{\bm{z}_*,2}>0$.
$\hfill\qedsymbol$

{\em Proof of Claim \ref{claim: cov tilde check}:}
Since $\tilde{e}_{\bm{z}_*}^{[j]}(t)$ and $\check{e}_{\bm{z}_*}^{[i]}(t)$ are zero-mean, it follows that $\cov(\tilde{e}_{\bm{z}_*}^{[j]}(t),\check{e}_{\bm{z}_*}^{[i]}(t))= \mathbb{E}[\tilde{e}_{\bm{z}_*}^{[j]}(t)\check{e}_{\bm{z}_*}^{[i]}(t)]$.

Recall that $\tilde{e}^{[i]}_{\bm{z}_*}(t)\triangleq (\hat{\sigma}^{[i]}_{\bm{z}_*|\mathcal{D}(t)})^2{\theta}^{[i]}_{\bm{z}_*,\bm{e}}(t)$ and the LTV solution \eqref{eq: decompose theta} gives $$\theta^{[i]}_{\bm{z}_*,\bm{e}}(t)=\sum_{l=1}^t\sum_{j=1}^n\phi_{ij}(t,l)\Delta\hat{e}^{[j]}_{\bm{z}_*}(l).$$  
By  Assumption \ref{assmp: constant A}, $\phi_{ij}(t,l)=\phi_{ij}(t,l')$ for all $0<l,l'\leqslant t$. Therefore, $$\theta^{[i]}_{\bm{z}_*,\bm{e}}(t)=\sum_{l=1}^t\sum_{j=1}^n\phi_{ij}(t,t)\Delta\hat{e}^{[j]}_{\bm{z}_*}(l).$$ Because of the initial condition $\hat{e}^{[j]}_{\bm{z}_*}(0)=0$, we have
$\theta^{[i]}_{\bm{z}_*,\bm{e}}(t)=\sum_{j=1}^n\phi_{ij}(t,t)\hat{e}^{[j]}_{\bm{z}_*}(t)$.
Then 
\begin{align*}
\mathbb{E}[\check{e}^{[i]}_{\bm{z}_*}(t)\tilde{e}^{[j]}_{\bm{z}_*}(t)]=(\hat{\sigma}^{[i]}_{\bm{z}_*|\mathcal{D}(t)})^2\mathbb{E}[\check{e}^{[i]}_{\bm{z}_*}(t)(\sum_{j=1}^n\phi_{ij}(t,t)\hat{e}^{[j]}_{\bm{z}_*}(t))].   
\end{align*}

Since $e_{\bm{z}_*}^{[i]}$ and $e_{\bm{z}_*}^{[j]}$ are zero-mean and independent if $i\neq j$, we have $\mathbb{E}[e_{\bm{z}_*}^{[i]}e_{\bm{z}_*}^{[j]}]=0$. 
Since Remark \ref{remark: lim check mu} indicates that $\check{e}^{[i]}_{\bm{z}_*}(t)=\check{\sigma}^{-2}_{\bm{z}_*|\mathcal{D}^{[i]}(t)}e_{\bm{z}_*}^{[i]}$, we  have $\check{e}^{[i]}_{\bm{z}_*}(t)$ and $\check{e}^{[j]}_{\bm{z}_*}(t)$ are also zero-mean and independent. Therefore, $\mathbb{E}[\check{e}^{[i]}_{\bm{z}_*}(t)\check{e}^{[j]}_{\bm{z}_*}(t)]=0$, $\forall j\neq i$. Since $\hat{e}^{[i]}_{\bm{z}_*}(t)$ is linear in $\check{e}^{[i]}_{\bm{z}_*}(t)$, we have $\mathbb{E}[\check{e}^{[i]}_{\bm{z}_*}(t)\hat{e}^{[j]}_{\bm{z}_*}(t)]=0$, $j\neq i$.
Hence, we further have
$$
    \mathbb{E}[\check{e}^{[i]}_{\bm{z}_*}(t)\tilde{e}^{[j]}_{\bm{z}_*}(t)]
   =(\hat{\sigma}^{[i]}_{\bm{z}_*|\mathcal{D}(t)})^2\check{\sigma}^{-2}_{\bm{z}_*|\mathcal{D}^{[i]}(t)}\phi_{ii}(t,t)\mathbb{E}[(\check{e}^{[i]}_{\bm{z}_*}(t))^2].
$$
Since $\phi_{ii}(t,t)\geqslant0$ implied by Assumption \ref{assmp:Non-degeneracy}, we have $\mathbb{E}[\check{e}^{[i]}_{\bm{z}_*}(t)\tilde{e}^{[j]}_{\bm{z}_*}(t)]\geqslant 0$. $\hfill\qedsymbol$

{\em Proof of Claim \ref{claim: cov tilde + check}:} 
Recall that $\tilde{e}_{\bm{z}_*}^{[p]}(t)$ and $\check{e}_{\bm{z}_*}^{[p]}(t)$ are zero mean for all $p\in\mathcal{V}$, hence 
\begin{align*}
    &\cov(\tilde{e}_{\bm{z}_*}^{[i]}(t)+\check{e}_{\bm{z}_*}^{[i]}(t),\tilde{e}_{\bm{z}_*}^{[j]}(t)+\check{e}_{\bm{z}_*}^{[j]}(t))\\
    &=\mathbb{E}[\big(\tilde{e}_{\bm{z}_*}^{[i]}(t)+\check{e}_{\bm{z}_*}^{[i]}(t)\big)\big(\tilde{e}_{\bm{z}_*}^{[j]}(t)+\check{e}_{\bm{z}_*}^{[j]}(t)\big)].
\end{align*}
Recall that $\hat{e}^{[p]}_{\bm{z}_*}(l)=\check{\sigma}^{-2}_{\bm{z}_*|\mathcal{D}^{[p]}(l)}\check{e}^{[p]}_{\bm{z}_*}(l)$, $\check{e}^{[p]}_{\bm{z}_*}(l)=\frac{\kappa(\rho^{\mathcal{Z}^{[p]}(l)}_{\bm{z}_*})}{\sigma_f^2+(\sigma^{[i]}_e)^2}e^{[p]}_{\bm{z}_*^{[p]}(l)}$.
 By independence of $e^{[p]}_{\bm{z}}$ over $p\in\mathcal{V}$, we have $\mathbb{E}[e^{[p]}_{\bm{z}}e^{[p']}_{\bm{z}}]=0$ and hence $\mathbb{E}[\check{e}^{[p]}_{\bm{z}}\check{e}^{[p']}_{\bm{z}}]=0$ if $p\neq p'$, and then
 \begin{align}\label{ineq: cov >= E}
     &\cov(\tilde{e}_{\bm{z}_*}^{[i]}(t)+\check{e}_{\bm{z}_*}^{[i]}(t),\tilde{e}_{\bm{z}_*}^{[j]}(t)+\check{e}_{\bm{z}_*}^{[j]}(t))\nonumber\\
     &\geqslant\mathbb{E}[\tilde{e}_{\bm{z}_*}^{[i]}(t)\tilde{e}_{\bm{z}_*}^{[j]}(t)]+\mathbb{E}[\tilde{e}_{\bm{z}_*}^{[i]}(t)\check{e}_{\bm{z}_*}^{[j]}(t)]+\mathbb{E}[\tilde{e}_{\bm{z}_*}^{[j]}(t)\check{e}_{\bm{z}_*}^{[i]}(t)]\nonumber\\
     &\geqslant \mathbb{E}[\tilde{e}_{\bm{z}_*}^{[i]}(t)\tilde{e}_{\bm{z}_*}^{[j]}(t)], 
 \end{align}
 where the last inequality follows from Claim \ref{claim: cov tilde check}. Obviously, $\mathbb{E}[\tilde{e}_{\bm{z}_*}^{[i]}(t)\tilde{e}_{\bm{z}_*}^{[j]}(t)]\geqslant 0$ if $i=j$. Next, we consider $i\neq j$.

By definition of $\tilde{e}_{\bm{z}_*}^{[i]}(t)$ and the LTV solution \eqref{eq: decompose theta} of ${\theta}^{[i]}_{\bm{z}_*,\bm{e}}(t)$, we can write
$$
    \tilde{e}_{\bm{z}_*}^{[i]}(t)=(\hat{\sigma}^{[i]}_{\bm{z}^{[i]}_{agg*}(t)|\mathcal{D}(t)})^2\sum_{l=1}^t\sum_{p=1}^n\phi_{ip}(t,l)\Delta\hat{e}^{[p]}_{\bm{z}_*}(l).
$$
Assumption 
\ref{assmp: constant A} implies ${\phi}_{ip}(t,l)={\phi}_{ip}(t,l')$  $\forall l,l'\in[1,t]$, $t\geqslant1$. 
Therefore, we can denote $\tilde{\phi}_{ip}(t)\triangleq(\hat{\sigma}^{[i]}_{\bm{z}^{[i]}_{agg*}(t)|\mathcal{D}(t)})^2{\phi}_{ip}(t,l)$,  $\forall l\in[1,t]$. Due to the initial condition $\hat{e}^{[j]}_{\bm{z}_*}(0)=0$, we can further write
$\tilde{e}_{\bm{z}_*}^{[i]}(t)=\sum_{p=1}^n\tilde{\phi}_{ip}(t)\hat{e}^{[p]}_{\bm{z}_*}(t)$.
It gives
$$
    \mathbb{E}[\tilde{e}_{\bm{z}_*}^{[i]}(t)\tilde{e}_{\bm{z}_*}^{[j]}(t)]=
    \mathbb{E}[\sum_{p=1}^n\big(\tilde{\phi}_{ip}(t)\hat{e}^{[p]}_{\bm{z}_*}(t)\big)\sum_{p'=1}^n\big(\tilde{\phi}_{jp'}(t)\hat{e}^{[p']}_{\bm{z}_*}(t)\big)],$$
Since $\mathbb{E}[e^{[p]}_{\bm{z}}e^{[p']}_{\bm{z}}]=0$ if $p\neq p'$ (the independence of $e^{[p]}_{\bm{z}}$ over $p\in\mathcal{V}$) and $\hat{e}^{[p]}_{\bm{z}}(t)$ is linear in $e^{[p]}_{\bm{z}}$, we have $\mathbb{E}[\hat{e}^{[p]}_{\bm{z}}(t)\hat{e}^{[p']}_{\bm{z}}(t)]=0$ if $p\neq p'$. This gives
$$\mathbb{E}[\tilde{e}_{\bm{z}_*}^{[i]}(t)\tilde{e}_{\bm{z}_*}^{[j]}(t)]
    =\mathbb{E}[\sum_{p=1}^n\tilde{\phi}_{ip}(t)\tilde{\phi}_{jp}(t)\big(\hat{e}^{[p]}_{\bm{z}_*}(t)\big)^2].
$$
Assumption \ref{assmp:Non-degeneracy} implies
$\phi_{ip}(t,l)\geqslant 0$, $\forall l\in[1,t],~i\in\mathcal{V}$. Hence $\tilde{\phi}_{ip}(t)\tilde{\phi}_{jp}(t)\geqslant 0$. Therefore, $\mathbb{E}[\tilde{e}_{\bm{z}_*}^{[i]}(t)\tilde{e}_{\bm{z}_*}^{[j]}(t)]\geqslant0$. Combining this with \eqref{ineq: cov >= E}  finishes the proof. $\hfill\qedsymbol$

Lemma \ref{lemma: sufficient condition for improvement} shows a sufficient condition for $\underset{{t\to\infty}}{\lim}\mathcal{Z}^{[i]}_{agg}(t)\neq\emptyset$.
\begin{lemma}\label{lemma: sufficient condition for improvement}
Suppose the same conditions for Corollary \ref{corrollary: distributed variance convergence} hold  and $d^{[j]}(t)\to0$ for all $j\in\mathcal{V}$. If $(\sigma_e^{[i]})^2>\frac{1}{n}\sum_{j=1}^n(\sigma_e^{[j]})^2$ for some $i\in\mathcal{V}$, then  $\underset{t\to\infty}{\lim}\mathcal{Z}^{[i]}_{agg}(t)=\mathcal{Z}_{agg}$.

\textbf{Proof: }
Since function $f_2$ in Lemma \ref{lmm: f} is strictly increasing, we have
\begin{align}\label{eq: f> 1/n f}
    f_2((\sigma_e^{[i]})^2)> f_2(\frac{1}{n}\sum_{j=1}^n(\sigma_e^{[j]})^2)\geqslant\frac{1}{n}\sum_{j=1}^nf_2((\sigma_e^{[j]})^2).
\end{align}
By Corollary \ref{corollary: lim check sigma},
$f_2((\sigma^{[i]}_e)^2)=\underset{{t\to\infty}}{\lim}\check{\sigma}^2_{\bm{z}_{agg}|\mathcal{D}^{[i]}(t)}$ for any $\bm{z}_{agg}\in\mathcal{Z}_{agg}$, and by Corollary 3.1 in \cite{zhu2010discrete},
\begin{align*}
  \lim_{t\to\infty}(\hat{\sigma}^{ave,[i]}_{\bm{z}_{agg}|\mathcal{D}(t)})^2
&=\underset{{t\to\infty}}{\lim}\lambda_{\bm{z}_{agg}}^{[i]}(t)=\underset{{t\to\infty}}{\lim}\frac{1}{n}\sum_{j=1}^n\check{\sigma}^2_{\bm{z}_{agg}|\mathcal{D}^{[j]}(t)}\\
&=\frac{1}{n}\sum_{j=1}^nf_2((\sigma^{[j]}_e)^2) 
\end{align*}
Combining these two statements with \eqref{eq: f> 1/n f} gives
\begin{align}\label{ineq: jensen}
   \underset{t\to\infty}{\lim}\check{\sigma}^2_{\bm{z}_{agg}|\mathcal{D}^{[i]}(t)}>
   \underset{t\to\infty}{\lim}(\hat{\sigma}_{\bm{z}_{agg}|\mathcal{D}(t)}^{ave,[i]})^2.
\end{align}

Taking the inverse of \eqref{eq: f> 1/n f} gives 
\begin{align*}
    \big(f_2((\sigma_e^{[i]})^2)\big)^{-1}    &<\big(\frac{1}{n}\sum_{j=1}^nf_2((\sigma_e^{[j]})^2)\big)^{-1}\\
    &\leqslant \frac{1}{n}\sum_{j=1}^n\big(f_2((\sigma_e^{[j]})^2)\big)^{-1},
\end{align*}
where the last inequality follows from Lemma \ref{lmm: f}. This gives
\begin{align*}
   \lim_{t\to\infty}\check{\sigma}^{-2}_{\bm{z}_{agg}|\mathcal{D}^{[i]}(t)}&<\frac{1}{n}\sum_{j=1}^n\underset{{t\to\infty}}{\lim}\check{\sigma}^{-2}_{\bm{z}_{agg}|\mathcal{D}^{[j]}(t)}\\
   &=\lim_{t\to\infty}(\check{\sigma}^{(agg)}_{\bm{z}_{agg}|\mathcal{D}(t)})^{-2} 
\end{align*}
where the equality follows from \eqref{eq:agg_var}. This is equivalent to 
$$
    \underset{t\to\infty}{\lim}\check{\sigma}^{2}_{\bm{z}_{agg}|\mathcal{D}^{[i]}(t)}>\underset{t\to\infty}{\lim}(\check{\sigma}^{(agg)}_{\bm{z}_{agg}|\mathcal{D}(t)})^{2}=\underset{t\to\infty}{\lim}(\hat{\sigma}^{[i]}_{\bm{z}_{agg}|\mathcal{D}(t)})^2,
$$
where the equality follows from Corollary \ref{corrollary: distributed variance convergence}.
Given \eqref{ineq: jensen} and the inequality above, since $\bm{z}_{agg}\in\mathcal{Z}_{agg}$ is arbitrary, we have $\underset{t\to\infty}{\lim}\mathcal{Z}^{[i]}_{agg}(t)=\mathcal{Z}_{agg}$. $\hfill\blacksquare$

\end{lemma}

We now proceed to finish the proof of Part II of Theorem \ref{thm: mean convergence}.

{\bf Proof of Theorem \ref{thm: mean convergence} Part II:
} By Line \ref{ln: empty Z_agg}-\ref{ln: return local} in fused GPR, it is obvious that if $\underset{t\to\infty}{\lim}\mathcal{Z}^{[i]}_{agg}(t)=\emptyset$, then $\underset{t\to\infty}{\lim}(\tilde{\mu}^{[i]}_{\bm{z}_*|\mathcal{D}(t)}-\eta(\bm{z}_*))^2=\underset{t\to\infty}{\lim}(\check{\mu}_{\bm{z}_*|\mathcal{D}^{[i]}(t)}-\eta(\bm{z}_*))^2$.
Proposition \ref{prop: Zagg non empty} presents the case of $\underset{t\to\infty}{\lim}\mathcal{Z}^{[i]}_{agg}(t)\neq\emptyset$. Lemma \ref{lemma: sufficient condition for improvement} corresponds to the sufficient condition for  Proposition \ref{prop: Zagg non empty}, which is the sufficient condition for $\underset{t\to\infty}{\lim}\mathcal{Z}^{[i]}_{agg}(t)\neq\emptyset$.
 $\hfill\blacksquare$

\section{Simulation}\label{sec:simulation}

In this section, we conduct Monte Carlo simulation to evaluate the developed algorithm. For the algorithms introduced below, we use (NN) to denote the version of the algorithm related to Nearest-neighbor GPR and (full) to denote the version related to full GPR. We compare LiDGPR (NN), i.e., Algorithm \ref{alg:distributed_GPR}, with five benchmarks:
(i) agent-based GPR (NN), i.e., Nearest-neighbor GPR (Algorithm \ref{alg:GPR});  
(ii) agent-based GPR (full), i.e.,  Algorithm \ref{alg:GPR} is replaced by \eqref{eq:full GPR} and hence $\check{\bm{\mu}}_{\mathcal{Z}_{*}|\mathcal{D}^{[i]}(t)}=\bm{\mu}_{\mathcal{Z}_{*}|\mathcal{D}^{[i]}(t)}, \check{\bm{\sigma}}^2_{\mathcal{Z}_{*}|\mathcal{D}^{[i]}(t)}=[\Sigma_{\bm{z}_*|\mathcal{D}^{[i]}(t)}]_{\bm{z}_*\in\mathcal{Z}_*}$;  
(iii) LiDGPR (full), i.e., Algorithm \ref{alg:distributed_GPR} with  Algorithm \ref{alg:GPR} replaced by agent-based GPR (full); 
(iv) centralized Nearest-neighbor GPR (cNN-GPR, the centralized counterpart of LiDGPR (NN)), i.e., Nearest-neighbor GPR using all the data collected by all the agents; 
(v) centralized full GPR, i.e., \eqref{eq:full GPR} using all the data collected by all the agents.
The simulations are run in Python, Linux Ubuntu 18.04 on an Intel Xeon(R) Silver 
4112 CPU, 2.60 GHz with 32 GB of RAM.

\begin{figure}[t]
      \centering
      \vspace{-0.4cm}
      \begin{subfigure}[b]{0.24\textwidth}
      \includegraphics[width=0.9\textwidth,height=0.63\textwidth]{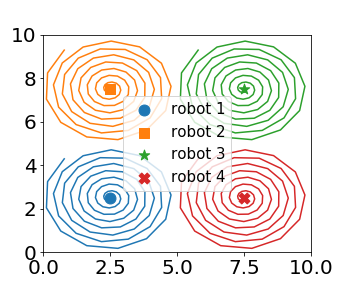}
       \caption{The trajectories of the robots} 
      \label{fig:trajectory}
      \end{subfigure}
        \hfil
    \begin{subfigure}[b]{0.24\textwidth}
      \centering
        \includegraphics[width=1.\textwidth]{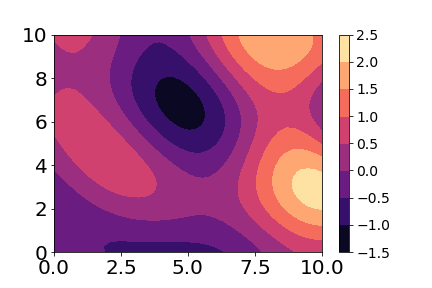}
        \caption{Ground truth of $\eta$}
        \label{fig: truth eta}
      
    \end{subfigure}
    \caption{Robot trajectories and ground truth of $\eta$}
    \label{fig:agent trajectory and ground truth}
\end{figure}

Consider the scenario where four mobile robots are wandering in $\bm{\mathcal{Z}}\triangleq [1,10]\times[1,10]$ and learning  spatial  signals, such as temperature or wind fields. Specifically, the robots are learning  10 different signals in the form  $\eta(\bm{z})=\beta\sum_{m=1}^{10}\alpha_m\sin(w_{m,1}z_1+w_{m,2}z_2)$, where   $\alpha_m\sim\mathcal{N}(0,0.01)$, $w_{m,1}\sim\mathcal{N}(0,1)$, $w_{m,2}\sim\mathcal{N}(0,1)$, $\beta$ is chosen such that $SNR\triangleq \frac{\int\eta(\bm{z})^2d\bm{z}}{\sigma_e^2}=2$. A realization of $\eta$ is shown in Figure \ref{fig: truth eta}. For each signal, the robots repeat the trajectories for 10 times, and the observations along each trajectory are subject to a different noise, where the variances of the observation noises follow $(\sigma_e^{[i]})^2=\sigma_e^2\sim\mathcal{U}(0,0.25)$ for all $i\in\mathcal{V}$.  Notice that there are totally 100 simulations.

The communication graph of the robots is characterized by adjacency matrix
$
    A(t)= \frac{1-(-1)^t}{2}\begin{bmatrix}0.5&0&0.5&0\\0&0.5&0&0.5\\0.5&0&0.5&0\\0&0.5&0&0.5
    \end{bmatrix}+\frac{1+(-1)^t}{2}\begin{bmatrix}0.5&0.25&0&0.25\\0.25&0.5&0.25&0\\0&0.25&0.5&0.25\\0.25&0&0.25&0.5\end{bmatrix},
$
which satisfies Assumption \ref{assmp:Periodical Strong Connectivity}, \ref{assmp:Balanced Communication} and \ref{assmp:Non-degeneracy}.
As shown in Figure \ref{fig:trajectory}, the robots have spiral trajectories generated by dynamics  $\begin{bmatrix}
z^{[i]}_1(t)\\z^{[i]}_2(t)
\end{bmatrix}=\begin{bmatrix}
z^{[i]}_1(t-1)\\z^{[i]}_2(t-1)
\end{bmatrix}+0.05t\begin{bmatrix}
\sin(0.5t)\\\cos(0.5t)
\end{bmatrix}$, where the initial states of the robots are $(2.5,2.5)$, $(2.5,7.5)$, $(7.5,7.5)$ and $(7.5,2.5)$ respectively.   Each robot $i$ collects training data along its trajectory, i.e., $\big(\bm{z}^{[i]}(t),\eta(\bm{z}^{[i]}(t))+e^{[i]}(t)\big)$, $e^{[i]}(t)\sim\mathcal{N}(0,(\sigma^{[i]}_e)^2)$, $t\geqslant1$. The set $\mathcal{Z}_*$ of test points are uniformly separated over $\bm{\mathcal{Z}}$, and $|\mathcal{Z}_*|=1600$.  We use 25\% of the test points for the set $\mathcal{Z}_{agg}$, i.e., $|\mathcal{Z}_{agg}|=400$. The points in $\mathcal{Z}_{agg}$ are uniformly separated.  The kernel is $k(\bm{z},\bm{z}')=\sigma_f^2\exp(-2\|\bm{z}-\bm{z}'\|^2)$, where $\sigma_f^2$ is chosen following the procedure under Remark \ref{remark: GPR kernel}. The resulting  $\sigma_f^2$  ranges from $1.1$ to $5.8$ for each experiment in the Monte Carlo simulation.
The prior mean is $\mu(\bm{z})=0$ for all $\bm{z}\in\mathcal{Z}$.

\begin{figure}[thpb]
      \centering
      \begin{subfigure}[t]{0.24\textwidth}
      \includegraphics[width=1.\textwidth]{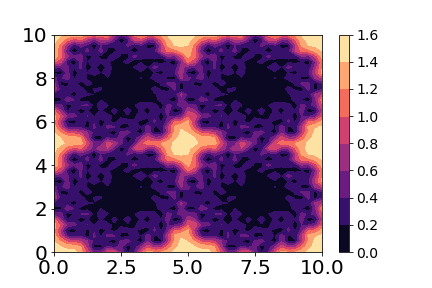}
       \caption{Predictive variance} 
      
      \end{subfigure}
        \hfil
    \begin{subfigure}[t]{0.24\textwidth}
      \centering
        \includegraphics[width=1.\textwidth]{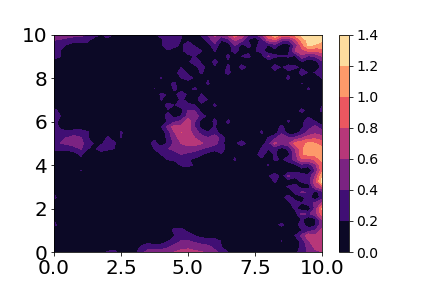}
        \caption{Predictive error}
      
    \end{subfigure}
    \caption{Predictive variance and  error of cNN-GPR}
    \label{fig:all data}
\end{figure}
 The performances of the robots are similar, and we present the figures for robot 1 due to space limitation.
Let the predictive  error at $\bm{z}_*\in\mathcal{Z}_*$ be the distance between predictive mean and the ground truth of $\eta$ at $\bm{z}_*$, where the distance adopts 2-norm. For example, the predictive  error at $\bm{z}_*$ of  agent-based GPR (NN) is $(\check{\mu}_{\bm{z}_*|\mathcal{D}^{[i]}(t)}-\eta(\bm{z}_*))^2$. When robots' trajectories and $\eta$ are those in Figure \ref{fig:agent trajectory and ground truth}, Figure \ref{fig:all data} shows the predictive variance and predictive error over $\mathcal{Z}_*$ of cNN-GPR. We can see that the predictive variances and errors are smaller near the trajectories of the robots.

\begin{figure}[thpb]
    \centering
    \begin{subfigure}[t]{0.24\textwidth}
      \centering
        \includegraphics[width=1.\textwidth]{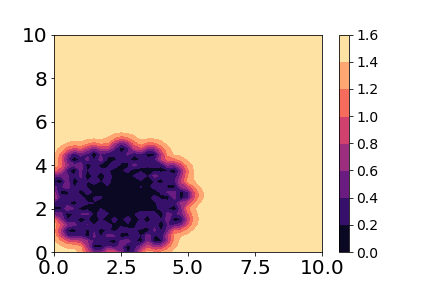}
        \caption{Variance of  agent-based GPR (NN)}
      
    \end{subfigure}
    \hfil
    \begin{subfigure}[t]{0.24\textwidth}
      \centering
        \includegraphics[width=1.\textwidth]{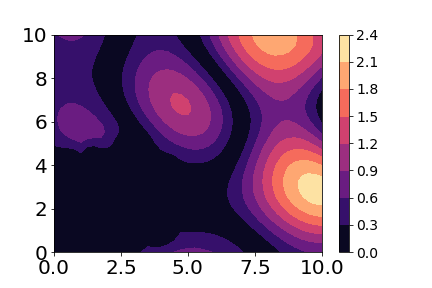}
        \caption{Predictive error of agent-based GPR (NN)}

    \end{subfigure}
    \hfil
    \begin{subfigure}[t]{0.24\textwidth}
      \centering
        \includegraphics[width=1.\textwidth]{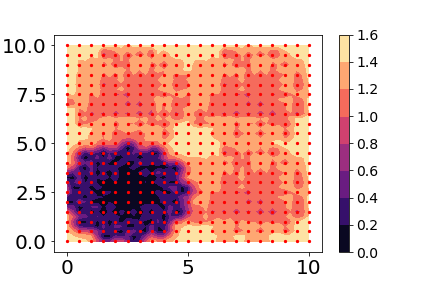}
        \caption{Variance of  LiDGPR (NN)}
      \label{fig: fused var}
    \end{subfigure}
    \hfil
    \begin{subfigure}[t]{0.24\textwidth}
      \centering
        \includegraphics[width=1.\textwidth]{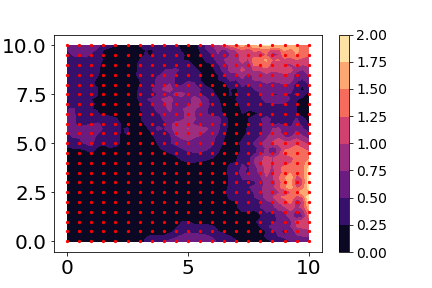}
        \caption{Predictive error of  LiDGPR (NN)}
        \label{fig: fused error}
    \end{subfigure}
    \caption{Comparison of  agent-based GPR (NN) and  LiDGPR (NN)}
\label{fig:local vs  LiDGPR}
\end{figure}

Figure \ref{fig:local vs  LiDGPR} shows the predictive variances and predictive errors of  agent-based GPR (NN) and   LiDGPR (NN) over $\mathcal{Z}_*$ of robot 1. 
We can see that by only communicating a portion of the testing sets,  LiDGPR (NN) improves the learning performances over agent-based GPR (NN) with reduced predictive variances and errors. The red dots   in Figures \ref{fig: fused var} and \ref{fig: fused error} are the points of $\mathcal{Z}_{agg}$, and the ``holes'' indicate that the improvements take place around the trajectories (training data) of the other robots, which corresponds to the term $\kappa(\rho^{\mathcal{Z}^{[j]}(t)}_{\bm{z}^{[i]}_{agg*}(t)})^2$  in Part II of Theorem \ref{thm: variance convergence}. In addition, the improvements reduce as the test points are moving away from $\mathcal{Z}_{agg}$, which corresponds to the terms $\kappa(\rho^{\bm{z}^{[i]}_{agg*}(t)}_{\bm{z}_{*}})^2$  in Part II of Theorem \ref{thm: variance convergence} and $\kappa(\rho_{\bm{z}_*}^{\mathcal{Z}_{agg}})$  in Part II of  Theorem \ref{thm: mean convergence} respectively.

\begin{figure}[thpb]
 \begin{subfigure}[t]{0.49\textwidth}
      \centering
        \includegraphics[width=1.\textwidth]{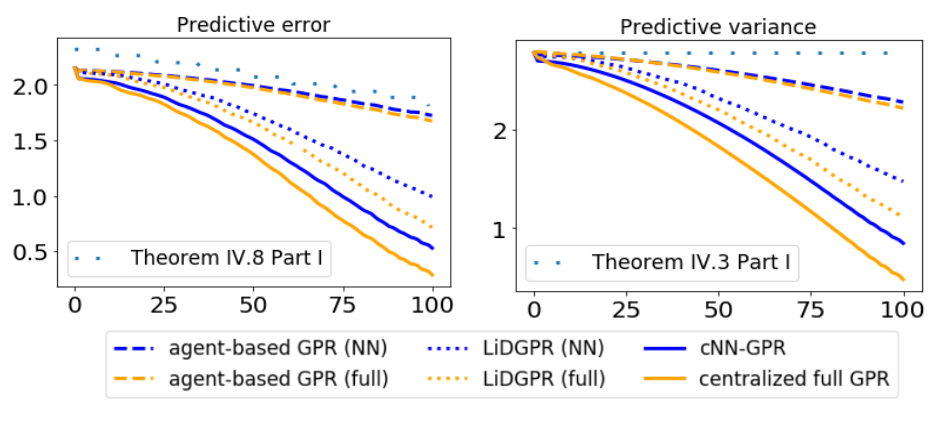}
        \caption{Comparison in predictive errors and variances (The upper bound in Theorem \ref{thm: mean convergence} Part 1 is scaled by 0.023)}
        \label{fig: average predictive variane}
    \end{subfigure}
    \hfil
    \begin{subfigure}[t]{0.23\textwidth}
      \centering
        \includegraphics[width=1\textwidth]{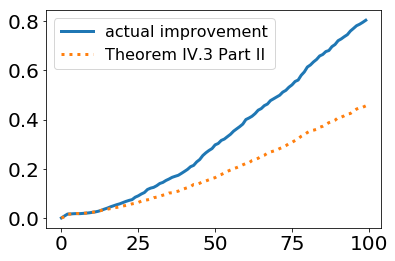}
        \caption{$\check{\sigma}^2_{\bm{z}_*|\mathcal{D}^{[i]}(t)}-(\tilde{\sigma}^{[i]}_{\bm{z}_*|\mathcal{D}(t)})^2$ vs. Theorem \ref{thm: variance convergence} Part II (scaled by 10)}
         \label{fig: var improve}
    \end{subfigure}
    \hfil
    \begin{subfigure}[t]{0.23\textwidth}
      \centering
        \includegraphics[width=1\textwidth]{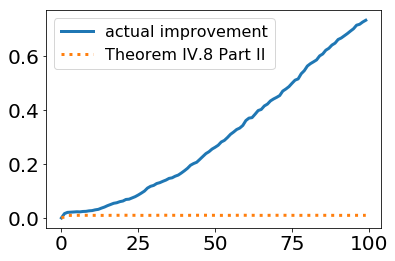}
        \caption{$\check{\mu}_{\bm{z}_*|\mathcal{D}^{[i]}(t)}-(\tilde{\sigma}^{[i]}_{\bm{z}_*|\mathcal{D}(t)})^2$ vs. Theorem \ref{thm: mean convergence} Part II (scaled by 10)}
      \label{fig: error improve}
    \end{subfigure}
    \hfil
    \caption{Average performance of robot 1 versus iteration number}
    \label{fig:temporalerror}
\end{figure}

 Figures \ref{fig: average predictive variane} compares the average predictive errors and variances of  LiDGPR (NN) with the five benchmarks.
The $x$-axis is the iteration number, corresponding to the size of training data. 
The predictive variance and error at each iteration are represented by the corresponding averages over $\mathcal{Z}_*$.

Note that the complexities in computation and memory are  respectively $\mathcal{O}(nt)$ and $\mathcal{O}(nt)$  for cNN-GPR, and $\mathcal{O}((nt)^3)$ and $\mathcal{O}((nt)^2)$ for centralized full GPR.   Notice that the differences in predictive variances and errors between cNN-GPR and centralized full GPR are small, while the diminishing rates are comparable. This shows that cNN-GPR has small performance loss compared to the benefit in reducing the complexities in computation and memory.

Comparing the curves of LiDGPR (NN) with agent-based GPR (NN) and agent-based GPR (full), we can see that LiDGPR (NN) not only compensates the information loss of using agent-based GPR (NN) to approximate agent-based GPR (full), but also gains extra information from the other robots. 

Figures \ref{fig: average predictive variane} plots the theoretic error bounds in Part I of Theorems \ref{thm: variance convergence} and \ref{thm: mean convergence} over the whole Monte Carlo simulation. By multiplying by a constant, we scale down the bound by factor $0.023$ in Part I of Theorem \ref{thm: mean convergence} for better visual comparison. The orders of rates of the bounds remain the same regardless of the scaling.
Comparisons between the theoretic improvement and the actual improvement of LiDGPR (NN) over agent-based GPR (NN) are shown in Figures \ref{fig: var improve} and \ref{fig: error improve}. Since the theoretic bounds are not tight, to make a meaningful comparison, we scale up the bounds by factor $10$ in Part II of Theorem IV.3  and IV.8. 

  The wall clock time for prediction using LiDGPR (NN) versus $t$, the number of local data points, after linear least-square fitting, has a slope 2.13e-6 second per test point per data point and a bias 1.19e-3 second per test point. Recall that Section \ref{sec: discussion} indicates that agent-based GPR (NN) has complexity $\mathcal{O}(t)$ and agent-based GPR (full) has complexity $\mathcal{O}(t^3)$.  The growth rate of the computation times (milliseconds) of LiDGPR (NN)  and LiDGPR (full) in the simulation are respectively $33.2t+200$ and $0.256t^3-0.1t^2-0.512t+27.6$. Over the simulation, the average is 1000 test-point predictions/second, or 1 kHz, with standard deviation 153 predictions/second.

\section{Conclusion}\label{sec:conclusion}
We propose the algorithm LiDGPR which allows a group of agents to collaboratively learn a common static latent  function through streaming data. The algorithm is cognizant of agents' limited resources in communication, computation and memory. We analyze the transient and steady-state behaviors of the algorithm and quantify the improvement brought by inter-agent communication. Simulations are conducted to confirm the theoretical findings.  Possible future works include analysis with different implementations of agent-based GPR and the consideration of dynamic latent functions.






\bibliographystyle{IEEEtran}
\bibliography{Biblio-dataset}
\vspace{-4em}
\begin{IEEEbiography}[{\includegraphics[width=1in,height=1.25in,clip,keepaspectratio]{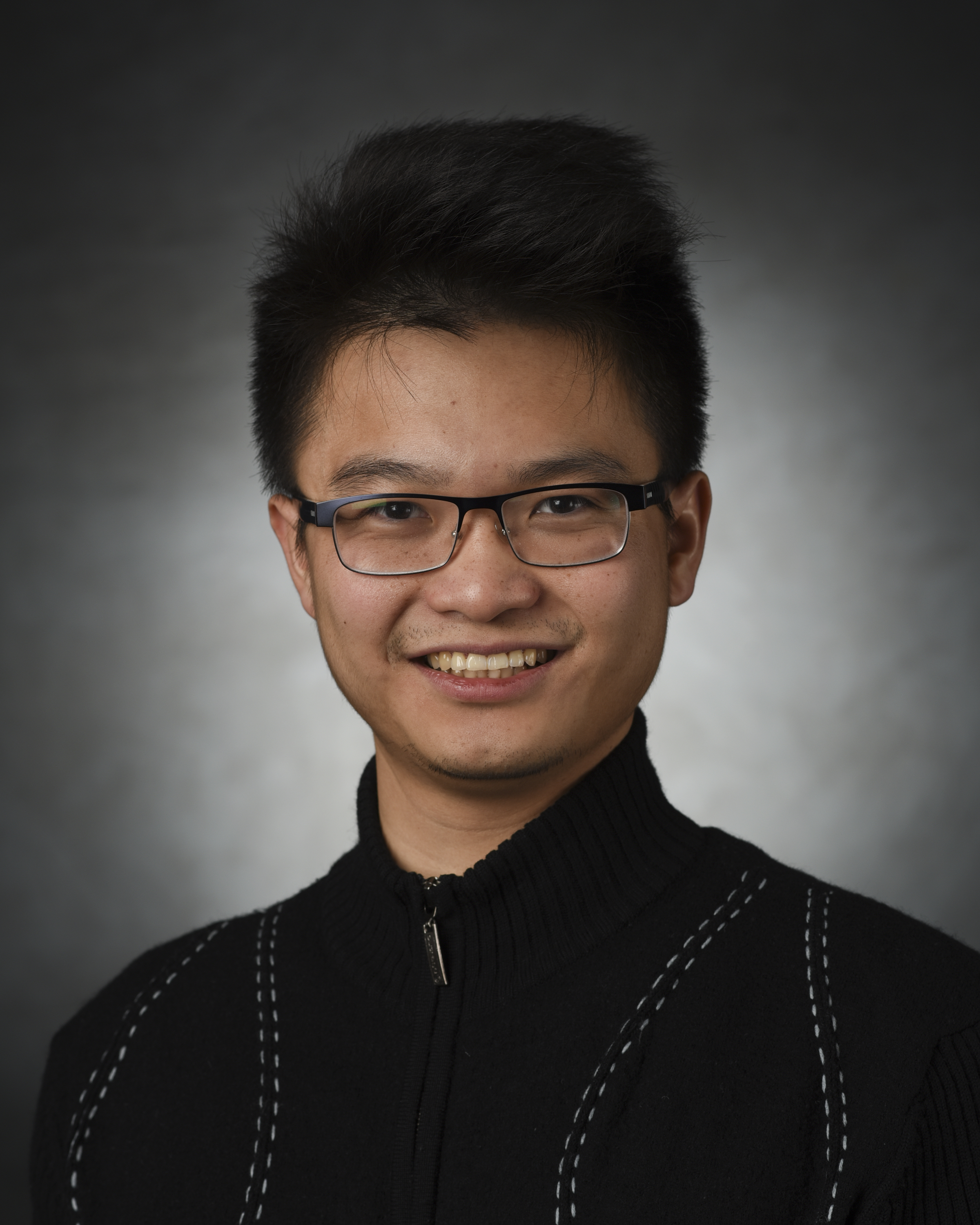}}]{Zhenyuan Yuan}
is  a  Ph.D.  candidate  in  the  School of  Electrical  Engineering  and  Computer  Science  at the  Pennsylvania  State  University.  He  received   B.S. in Electrical Engineering and B.S. in Mathematics from the  Pennsylvania  State  University  in  2018. His research interests lie in machine learning and motion planning with applications in robotic networks. He is a recipient of the Rudolf Kalman Best Paper Award of  the ASME Journal of Dynamic Systems Measurement and Control in 2019 and
the Penn State Alumni Association Scholarship for Penn State Alumni in the Graduate School in 2021.
\end{IEEEbiography}
\vspace{-4em}
\begin{IEEEbiography}[{\includegraphics[width=1in,height=1.25in,clip,keepaspectratio]{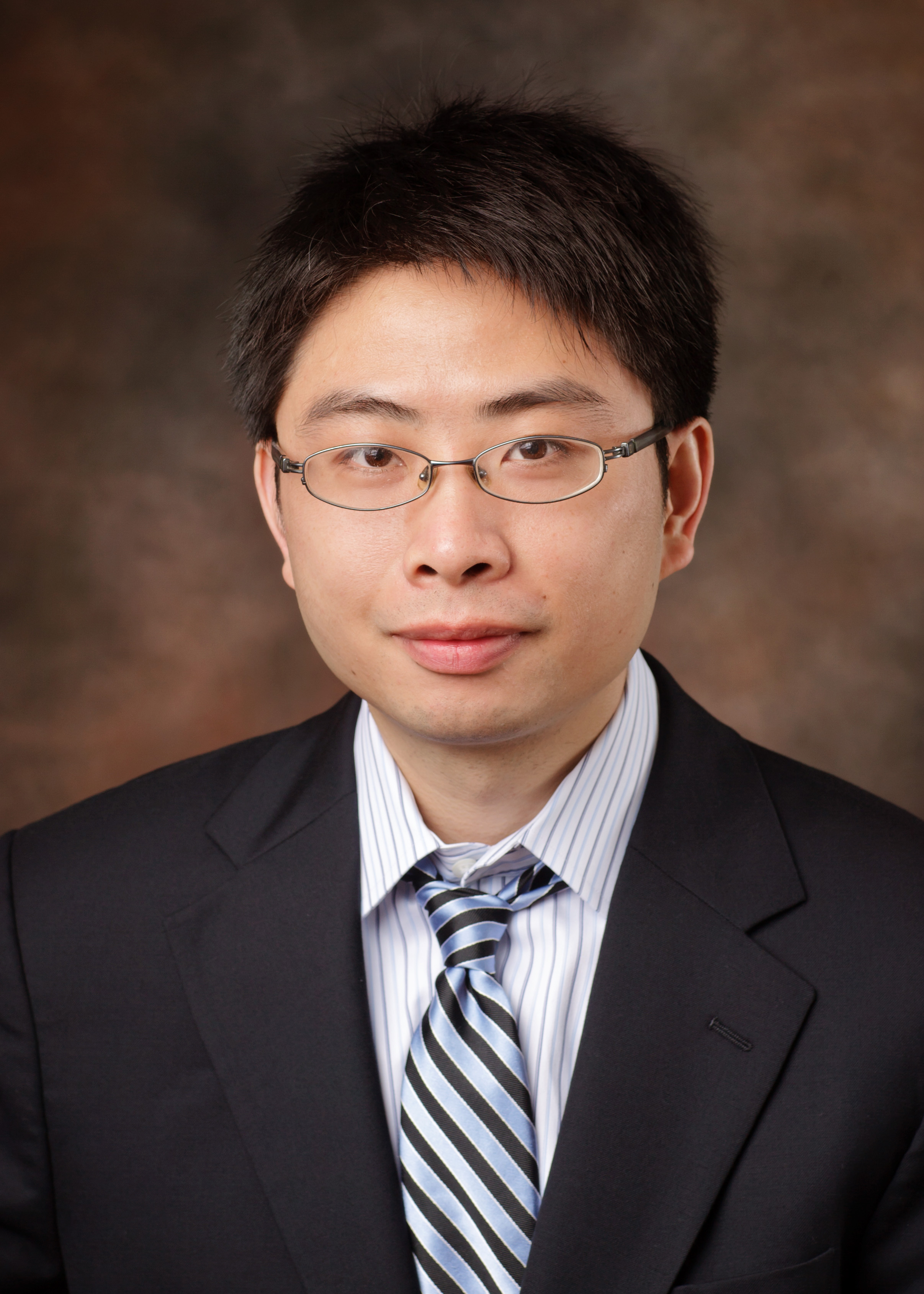}}]{Minghui Zhu}
is an Associate Professor in the School of Electrical Engineering and Computer Science at the Pennsylvania State University. Prior to joining Penn State in 2013, he was a postdoctoral associate in the Laboratory for Information and Decision Systems at the Massachusetts Institute of Technology. He received Ph.D. in Engineering Science (Mechanical Engineering) from the University of California, San Diego in 2011. His research interests lie in distributed control and decision-making of multi-agent networks with applications in robotic networks, security and the smart grid. He is the co-author of the book "Distributed optimization-based control of multi-agent networks in complex environments" (Springer, 2015). He is a recipient of the Dorothy Quiggle Career Development Professorship in Engineering at Penn State in 2013, the award of Outstanding Reviewer of Automatica in 2013 and 2014, and the National Science Foundation CAREER award in 2019. He is an associate editor of the Conference Editorial Board of the IEEE Control Systems Society and IET Cyber-systems and Robotics.
\end{IEEEbiography}

\appendices
\section{Hyperparameter optimization of $\sigma^{[i]}_f$}
Notice that inequality \eqref{eq: sigma_f condition} provides a sufficient condition for Theorem \ref{thm: mean convergence} and Theorem \ref{thm: variance convergence}. To further optimize the performance of the algorithm, we provide two approaches for optimizing $\sigma^{[i]}_f$. One is an offline approach that uses data collected during initilization, and the other is an online approach that updates the hyperparameter once new data are collected. Note that the selection of $\sigma^{[i]}_f$ is not  independent of the observations, as the noise variance $\sigma_e^{[i]}$ in \eqref{eq: sigma_f condition} reveals the noise level of the observations. We can also explicitly take data fitting into account. As discussed below \eqref{eq: sigma_f condition}, inequality \eqref{eq: sigma_f condition} can be satisfied when $\sigma^{[i]}_f$ is sufficiently large. Hence, the hyperparameter $\sigma^{[i]}_f$ obtained a priori can be considered as a lower bound.

For the offline approach, we follow the validation set approach in the provided reference
\cite{romeres2016online}. Specifically, after obtaining $\sigma_e^{[j]}$ from all $j\in\mathcal{V}$ using the Floodset algorithm mentioned in the paper, 
each agent $i$ can execute the following procedure to determine $\sigma^{[i]}_f$ in the initialization in a distributed manner:
\begin{enumerate}
	\item Compute the lower bound $\ubar{\sigma}^{[i]}_f$ satisfying \eqref{eq: sigma_f condition} a priori.
	\item Select a length $L$ and a resolution $\epsilon$.
	\item Construct a set  $\Sigma^{[i]}\triangleq\{\ubar{\sigma}^{[i]}_f, \ubar{\sigma}^{[i]}_f+\epsilon, \cdots, \ubar{\sigma}^{[i]}_f+L\epsilon\}$ that contains  feasible $\sigma^{[i]}_f$.
	\item Collect $n_{init}$ data points. Denote the resulting data set as $\mathcal{D}^{[i]}_{init}\triangleq\{\bm{z}^{[i]}(\tau),y^{[i]}(\tau)\}_{\tau=1}^{n_{init}}$.
	\item Select $\sigma^{[i]}_f\in\Sigma^{[i]}$ such that the mean-square error  over the locally collected dataset is minimized: $\sigma^{[i]}_f=\arg\min_{\sigma\in\Sigma^{[i]}}\frac{1}{n_{init}}\sum_{\tau=1}^{n_{init}}(\check{\mu}_{\bm{z}^{[i]}(\tau)|\mathcal{D}^{[i]}_{init}}(\sigma)-y^{[i]}(\tau))^2 $. Notice that the predictive mean $\check{\mu}_{\bm{z}|\mathcal{D}^{[i]}}$ from the agent-based GPR inherently depends on $\sigma^{[i]}_f$.
	\item Execute maximum consensus to unify $\sigma^{[i]}_f$ for all $i\in\mathcal{V}$.
\end{enumerate}
Since the hyperparameter $\sigma^{[i]}_f$ is one-dimensional, the above grid-based optimization method should be considerably effective. Notice that steps 1)-5) are done locally, and step 6) is done in a distributed manner and  terminates in $n(b-1)$ iterations.

For the online approach, we can incorporate \eqref{eq: sigma_f condition} with the recursive algorithm in \cite{huber2014recursive} to update $\sigma^{[i]}_f$. After computing $\ubar{\sigma}_f^{[i]}$ satisfying \eqref{eq: sigma_f condition} a priori in the initialization and initializing $\sigma^{[i]}_{f,0}=\ubar{\sigma}_f^{[i]}$, the following procedure can be performed at each iteration $t$ by each agent $i$ after data collection (Line 6 in Algorithm 1) and before the agent-based GPR (Line 7 in Algorithm 1):
\begin{enumerate}
	\item Compute $\sigma^{[i]}_{f,t}$ using  $(\bm{z}^{[i]}(t),y^{[i]}(t))$ and $\sigma^{[i]}_{f,t-1}$ via the recursive  algorithm in \cite{huber2014recursive}.
	\item Choose $\sigma^{[i]}_{f}(t)=\max\{\sigma^{[i]}_{f,t}, \ubar{\sigma}_f^{[i]}\}$.
	\item Execute maximum consensus to unify $\sigma^{[i]}_f(t)$ for all $i\in\mathcal{V}$.
\end{enumerate}
Notice that the online approach requires executing consensus at every iteration $t$. Therefore, it can increase the computation complexity of the whole algorithm. Since the maximum consensus terminates in $n(b-1)$ rounds, this approach favors the scenarios when the agent number $n$ is small. Since it actually uses all the data collected to estimate $\sigma^{[i]}_f$, it is most likely to achieve a better performance. On the other hand, the offline approach is a complement of the online approach and can trade off performance for computation complexity. It favors a larger $n$ and costs less  computation resources in total, since $\sigma^{[i]}_f$ is computed once during the initialization. Since the offline approach computes $\sigma^{[i]}_f$ only using a fixed amount of data, it may not perform as well as the online approach.

\section{Estimation of noise variance $\sigma_e^{[i]}$}
This paper assumes that $\sigma_e^{[i]}$ is known a priori. We do note that the noise variance $\sigma_e^{[i]}$ is also a hyperparameter which usually needs to be learned from data in real-world applications 
\cite{mackay1999comparison}\cite{bachoc2013cross}.  On the other hand,
there are also cases where the noise variance $\sigma_e^{[i]}$ can be obtained a priori. For example, in robotic applications, $\sigma_e^{[i]}$ can be the system/measurement noise variance of onbroad sensors and can be obtained from  factory  specification sheets or learned offline through historic measurements \cite{wang2018safe}\cite{deisenroth2009gaussian}. 

In case  the noise variance $\sigma_e^{[i]}$ is unknown a priori, as a hyperparameter,  it can be directly estimated using existing methods (e.g., cross validation and maximum likelihood estimation) as those in \cite{mackay1999comparison}\cite{bachoc2013cross} using a given set of data collected during initialization, similar to \cite{romeres2016online}. 
Parameter
$\sigma_e^{[i]}$ can also be learned online recursively using existing methods \cite{huber2014recursive}. Since this problem  can be solved completely using existing methods, we assume it is  given for the sake of brevity in the exposition of our contribution and algorithm.

\section{Selection of number of data points for prediction}
In a benchmark of the simulation, the agent-based GPR (full) executes the classic full GPR, which uses all the local data. The simulation results in Figure \ref{fig:temporalerror} demonstrate that the improvement of the LiDGPR (full) over the agent-based GPR (full) is similar to that of LiDGPR (NN) over the agent-based GPR (NN).  Full GPR and NN GPR represent two extreme cases and are both studied in the simulation. Empirically, the agent-based GPR can use any number of data points. On the other hand, the current proof only  works for NN GPR which only uses one data point. 
Extending the analysis to more than one data point is not trivial, since it involves additional analysis of the inverse of the kernel Gram matrix.
We leave it as future work. The end of simulation Section VI highlights that the framework
empirically works for an arbitrary number of data points and one data point is used in the agent-based
GPR for the sake of computation complexity. 
The trade-off in performance is evident in Fig. \ref{fig: average predictive variane}. 

Next we provide a formal criterion to choose the number of data points for the agent-based GPR. The selection of the number of data points used in the agent-based GPR, or more generally, the selection of the number of data points (or inducing points) in sparse variational approximation of GPR, mostly depends on the desired computation complexity of the algorithm (Section 8, \cite{williams2006gaussian}).  Paper \cite{burt2019rates} derives an upper bound to characterize the performance loss in terms of the number of inducing points for  sparse variational approximation of GPR. Theorem 3 in \cite{burt2019rates} shows that the following holds with probability at least $1-\delta$,
\begin{align}\label{ineq: KL}
	KL(Q||\hat{P})\leq \frac{(M+1)N\sum_{m=M+1}^\infty\lambda_m+2N\sigma^2_f\epsilon}{\delta \sigma^2_e}(1+\frac{\|\mathbf{y}\|_2^2}{\sigma^2_e}),
\end{align}
where $\hat{P}$ is the posterior Gaussian process, $Q$ is the variational approximation distribution, $KL(Q||\hat{P})\triangleq \int_{-\infty}^{\infty}Q(x)\log\Big(\frac{Q(x)}{\hat{P}(x)}\Big)dx$ represents the Kullback-Leibler divergence between distributions $Q$ and $\hat{P}$,  $M$ is the number of inducing points, $N$ is the total number of data points,  $\epsilon$ represents the quality of the initialization,   $\mathbf{y}$ is the vector aggregating all the $N$ outputs, and  $\lambda_m$ is the $m$-th largest eigenvalue of the selected kernel $k$ when expressed in terms of the eigenvalues and eigenfunctions (Mercer's theorem, page 96, \cite{williams2006gaussian}).
Empirically, using the computer specified in the current paper,  we find out that the relation between the computation time $t$  (milliseconds) and the number of data points $M$ used in the agent-based GPR is given by  $t=0.256M^3-0.1M^2-0.512M+27.6$.
Therefore, selecting the number of data points $M$ can be formulated as the following optimization problem
\begin{align}
	\min_{M\in\{1,\cdots,N\}}&\quad(M+1)\sum_{m=M+1}^\infty\lambda_m\nonumber\\
	\textrm{subject to } &0.256M^3-0.1M^2-0.512M+27.6\leq h,\label{constraint}
\end{align}
where $h\in\mathbb{R}$ is a constant specifying the constraint in computation time and the objective function is derived from the upper bound in \eqref{ineq: KL}. 
Since $M$ is the only decision variable, the above optimization problem can be solved  brute-force by enumerating all possible values of $M$. This method can return the exact solution with computation cost  $\mathcal{O}(N)$. Note that problem \eqref{constraint} requires the eigenvalue $\lambda_m$.
Closed-form expressions of the eigenvalues of the most widely used squared exponential kernel can be found in \cite{zhu1997gaussian}. Approximations in Big O notation of the eigenvalues of the Matern kernel can be found in \cite{ritter1995multivariate}. Numerical approximations of the  eigenvalues of a general kernel can be found in Section 4.3.2 in \cite{williams2006gaussian}.

Alternatively, a multi-objective optimization problem can be formulated to balance between performance loss and computation complexity as
\begin{align}\label{soft}
	\min_{M\in\{1,\cdots,N\}}&\quad\omega \Big((M+1)\sum_{m=M+1}^\infty\lambda_m\Big)\nonumber\\
	&+ (1-\omega)\Big(0.256M^3-0.1M^2-0.512M\Big),
\end{align}
where $\omega\in[0,1]$ is a weight to be tuned based on a user's preference over performance and computation complexity. This formulation guarantees the solution being on the Pareto frontier of the two objectives of minimizing $(M+1)\sum_{m=M+1}^\infty\lambda_m$ and $ 0.256M^3-0.1M^2-0.512M$ \cite{emmerich2018tutorial}. Similarly, the above optimization problem can be solved brute-force by evaluating all the values of $M$.

Notice that this paper focuses on online learning, where data arrive sequentially. Therefore, the parameter $N$ in the above formulations monotonically increases with respect to time. In this case, we can solve the problem \eqref{constraint} or \eqref{soft} offline and store the solutions of $M$ with respect to $N$ in  a lookup table for online use. Notice that when hard constraints, such as \eqref{constraint}, are imposed, or when the maximum $N$ is known a priori, the lookup table is finite.
As  the analysis of this paper only works for NN GPR which uses only one data point, we do not include the above discussion in the paper in order to avoid confusion. 

\end{document}